%% file: main.tex
\documentclass[10pt,journal]{IEEEtran}

\usepackage[font=small,labelfont=bf,
   justification=justified,
   justification=justified,
   format=plain]{caption}
   
\usepackage[sort, numbers]{natbib} 

\usepackage{relsize}

\usepackage[english]{babel}
\usepackage{setspace}
\usepackage{amsmath}
\usepackage[colorlinks=true, allcolors=blue]{hyperref}
\usepackage{url} 
\usepackage{amssymb}
\usepackage{color, graphics, graphicx}
\usepackage{colortbl}
\usepackage[dvipsnames]{xcolor}
\usepackage{multirow}
\usepackage{helvet}
\usepackage{lscape}
\usepackage{bbm}
\usepackage{dsfont}
\usepackage{amsthm}
\usepackage{enumitem}
\usepackage{afterpage} 
\usepackage{longtable} 
\usepackage{booktabs} 
\usepackage{bbm}
\usepackage{flafter}
\usepackage{rotating}
\usepackage{tocbibind}
\usepackage{algorithm}
\usepackage{algpseudocode}
\usepackage[labelfont=bf, font=footnotesize]{caption}
\usepackage{subcaption}
\usepackage{textgreek}
\usepackage{bm}
\usepackage{amssymb}
\usepackage{mathrsfs}
\usepackage{xargs}
\usepackage{booktabs} 
\usepackage{xcolor,colortbl}
\usepackage{authblk}
\usepackage{datetime}
\usepackage{comment}
\usepackage{helvet}
\usepackage{pdflscape}
\usepackage{mathtools}
\usepackage{array,tabularx}
\usepackage{url} 
\usepackage{changepage} 
\usepackage{float}
\usepackage{bm}

\newcommand*{\B}[1]{\ifmmode\bm{#1}\else\textbf{#1}\fi}

\newenvironment{conditions*}
  {\par\vspace{\abovedisplayskip}\noindent
   \tabularx{\columnwidth}{>{$}l<{$} @{\ : } >{\raggedright\arraybackslash}X}}
  {\endtabularx\par\vspace{\belowdisplayskip}}

\begin{document}
\bstctlcite{IEEEexample:BSTcontrol}

\title{\LARGE OPTIMUS$^*$\thanks{$^*$Outcome Prediction Targetedly Integrating Multimodality with Unavailable Signals.}: Predicting Multivariate Outcomes
in Alzheimer’s Disease Using Multi-modal Data amidst Missing Values
}

\author{\large Christelle Schneuwly Diaz,
Duy-Thanh Vu,
Julien Bodelet,
Duy-Cat Can, \break
Guillaume Blanc,
Haiting Jiang,
Lin Yao,
Guiseppe Pantaleo,
ADNI$^\dag$\thanks{$^\dag$The data used in this article were from the Alzheimer’s Disease Neuroimaging Initiative (ADNI) database (adni.loni.usc.edu). As such, the investigators within the ADNI contributed to the design and implementation of ADNI and/or provided data but did not participate in the analysis or writing of this paper. A complete list of ADNI investigators can be found at: \url{http://adni.loni.usc.edu/wp-content/uploads/how_to_apply/ADNI_Acknowledgement_List.pdf}.},
Oliver Y. Chén

\thanks{
\textit{(Corresponding author: C. Schneuwly Diaz and O. Y. Ch\'en.)} } 
\thanks{C. Schneuwly Diaz is with the Platform of Bioinformatics (PBI), University Lausanne Hospital (CHUV) and Faculty of Biology and Medicine (FBM), University of Lausanne (UNIL), Lausanne, 1012, Switzerland. (email: christelle.schneuwly@chuv.ch).} 
\thanks{D.T. Vũ, J. Bodelet, D.C. Can, and G. Blanc are with PBI, CHUV, and FBM, UNIL, Lausanne, 1012, Switzerland.}
\thanks{H. Jiang and L. Yao are with the Frontier Science Center for Brain Science and Brain-machine Integration, Zhejiang University, Zhejiang; Zhejiang University School of Medicine, Zhejiang, 310058, China.} 
\thanks{O.Y. Chén is with PBI, CHUV and FBM, UNIL, Lausanne, 1012, Switzerland. (email: olivery.chen@chuv.ch).}

}

\markboth{ } 
{Shell \MakeLowercase{C.S. Diaz \textit{et al.}}: OPTIMUS}

\maketitle

\begin{abstract}
\textit{Objective}: Alzheimer's disease, a neurodegenerative disorder, is associated with neural, genetic, and proteomic factors while affecting multiple cognitive and behavioral faculties. Traditional AD prediction largely focuses on univariate disease outcomes, such as disease stages and severity. Multimodal data encode broader disease information than a single modality and may, therefore, improve disease prediction; but they often contain missing values. Recent ``deeper’’ machine learning approaches show promise in improving prediction accuracy, yet the biological relevance of these models needs to be further charted.
\textit{Methods}: Integrating missing data analysis, predictive modeling, multimodal data analysis, and explainable AI, we propose OPTIMUS, a predictive, modular, and explainable machine-learning framework, to unveil the \textit{many-to-many} predictive pathways between multimodal input data and multivariate disease outcomes amidst missing values.
\textit{Results}: OPTIMUS first applies modality-specific imputation to uncover data from each modality while optimizing overall prediction accuracy. It then maps multimodal biomarkers to multivariate outcomes using machine-learning and extracts biomarkers respectively predictive of each outcome. Finally, OPTIMUS incorporates XAI to explain the identified multimodal biomarkers.
\textit{Conclusion}: Using data from 346 cognitively normal subjects, 608 persons with mild cognitive impairment, and 251 AD patients, OPTIMUS identifies neural and transcriptomic signatures that jointly but differentially predict multivariate outcomes related to executive function, language, memory, and visuospatial function.
\textit{Significance}: Our work demonstrates the potential of building a predictive and biologically explainable machine-learning framework to uncover multimodal biomarkers that capture disease profiles across varying cognitive landscapes. The results improve our understanding of the complex \textit{many-to-many} pathways in AD.
\end{abstract}


\section{Introduction}
\label{Sec:Introduction}

\input{Sections/1-Introduction}

\section{Methods}
\label{Sec:Methods}
\input{Sections/2-Methods}

\section{Experiments and Results} \label{Sec:Results}

\input{Sections/3-Results}

\section{Discussion}
\label{Sec:Discussion}
\input{Sections/4-Discussion}

\section{Conclusion}
\label{Sec:Conclusion}
\input{Sections/5-Conclusion}

\section*{Acknowledgments}

C.S.D. and O.Y.C. conceptualised the study. C.S.D. developed the methods, wrote the codes, performed data processing and analysis, and wrote and maintained the software package. D.T.V. prepared and pre-processed all the data. J.B., D.C.C, G.B., H.J., and L.Y. provided methodological support. G.P. provided clinical support. C.S.D. and O.Y.C. wrote the manuscript, with comments from all other authors.

Data in this paper are from the Alzheimer's Disease Neuroimaging Initiative (ADNI) (National Institutes of Health Grant U01 AG024904) and DOD ADNI (Department of Defense award number W81XWH-12-2-0012). See Supplementary Materials for further information about ADNI.

\bibliographystyle{IEEEtran}
\bibliography{OPTIMUS_paper}

\clearpage  

\renewcommand{\thetable}{S\arabic{table}} 
\setcounter{table}{0}
\renewcommand{\thefigure}{S\arabic{figure}}
\setcounter{figure}{0} 


\onecolumn  
\renewcommand{\appendixname}{}  
\appendix

\begin{flushleft}
    {\Large Supporting Information for \par}
    \vspace{0.3cm}
    {\Large OPTIMUS: Predicting Multivariate Outcomes in Alzheimer’s Disease Using Multi-modal Data amidst Missing Values \par}
    \vspace{0.5cm}

   Christelle Schneuwly Diaz$^{1,2*}$, Duy Thanh Vũ$^{1,2}$, Julien Bodelet$^{1,2}$, Duy-Cat Can$^{1,2}$, \\
    Guillaume Blanc$^{1}$, Haiting Jiang$^{3,4}$, Lin Yao$^{3,4}$, Giuseppe Pantaleo$^{5}$, Oliver Y. Chén$^{1,2*}$
    \vspace{0.5cm}

    {\small
    $^1$Platform of Bioinformatics (PBI), University Lausanne Hospital (CHUV), Lausanne, Switzerland.\\
    $^2$Faculty of Biology and Medicine (FBM), University of Lausanne (UNIL), Lausanne, Switzerland.\\
    $^3$Frontier Science Center for Brain Science and Brain-machine Integration, Zhejiang University, Zhejiang, China.\\
    $^3$Zhejiang University School of Medicine, Zhejiang, China.\\
    $^5$Service of Immunology and Allergy, CHUV, Lausanne, Switzerland.\\
    $^*$Corresponding authors: \texttt{christelle.schneuwly@chuv.ch}, \texttt{olivery.chen@chuv.ch}
    }
    \vspace{0.5cm}
    
\end{flushleft}

\label{Sec:Supplementary}
\input{Sections/6-Supplementary}

\end{document}

%% file: Sections/1-Introduction.tex
Alzheimer's Disease (AD) is a neurodegenerative disorder affecting millions worldwide, with projections for future growth. The prevalence of AD is high: the global mean prevalence (weighted by population size) for (A$\beta$-positive) AD dementia is 4.2\%~(M) and 5.2\%~(F) for those between 60 and 74 years old, 9.1\%~(M) and 12.2\%~(F) for those between 75 and 84 years old, and 26.5\%~(M) and 38.4\%~(F) for those older than 85 years old~\citep{gustavssonGlobalEstimatesNumber2023}.

The quality of life for AD patients decreases as patients start to lose important cognitive, behavioral, and social abilities to cope with basic daily activities~\citep{
morrisClinicalDementiaRating1997}. Following diagnosis, an AD patient's typical life expectancy ranges from three to nine years~\citep{querfurthAlzheimersDisease2010}. Because AD and related dementia (ADRD) are directly associated with disability and, eventually, death, it brings both health and finantial burdens to manage the disease. The global value of a statistical life (VSL)-based economic burden for ADRD is at \$2.8 trillion in 2019 and is expected to reach \$16.9 trillion by 2050~\citep{nandiGlobalRegionalProjections2022}.

Effective biomarker discovery and early outcome prediction are essential for improving AD management. Because of AD's neural, genetic and proteomic associations, this requires a multimodal approach that integrates these sources to capture cognitive and behavioral manifestations of the disease. Central to this pursuit is mapping the complex relationships linking these diverse data modalities to various clinical outcomes. As multimodal data are often high-dimensional and the outcomes multivariate, we define biomarker discovery, outcome prediction, and pathway identification, in concert, as a \textit{many-to-many} problem for AD. 

In general, the many-to-many problem for AD consists of four components. First, one needs to find features from different modalities that are associated with each of the outcomes.
Second, one needs to find out how much contribution each selected biomarker makes to predicting every outcome, that is, to find the weighted pathways from the inputs to the outcomes. Third, one needs to predict multivariate outcomes using the identified modality-specific biomarkers and the many-to-many pathways. Finally, it is important to find biological insights of the identified biomarkers in light of clinical relevance. To do so, one needs to, for example, map brain imaging features to their neuroanatomical locations or link molecular markers to established disease pathways.

Chief to address the four components in many-to-many problem is to use machine learning (ML) methods. Indeed, several methods exist to handle these components individually. For example, to find modality-specific biomarkers, one can use variable selection tools, such as mass-multivariate methods, regularization, and projection pursuit~\citep{chenRolesStatisticsHuman2019a}.  
To find multimodal biomarkers, one can use deep learning~\citep{
qiuMultimodalDeepLearning2022}. Multivariate disease outcome prediction, although in its infancy, has seen promises using methods handling multivariate response variables, such as partial least squares~\citep{chenResidualPartialLeast2024b}. 
Addressing the components in many-to-many problem altogether, however, has been as-of-yet little explored, as it requires a comprehensive framework linking these four components. 

In parallel, as an integral part of the many-to-many problem, multimodal data often contain missing values. Although there exist missing data imputation techniques~\citep{leMultimodalMissingData2025}, due in part to the underlying heterogeneous, modality-specific distribution and in part to the disparate amount of missing data in different modalities, it is unclear which imputation method is most suitable for addressing missingness in multimodal data considering disease prediction and biological explanation. If not properly selected, some methods may yield more biased estimates, potentially resulting in misleading biomarkers and less accurate predictions.  
Finally, although ML continues to improve AD prediction, one tangling problem 
is their lack of explanation: how to translate the model and features to clinical insights~\citep{barredoarrietaExplainableArtificialIntelligence2020a}?

\begin{figure*}[!ht]
    \centering
\includegraphics[width=1\textwidth]{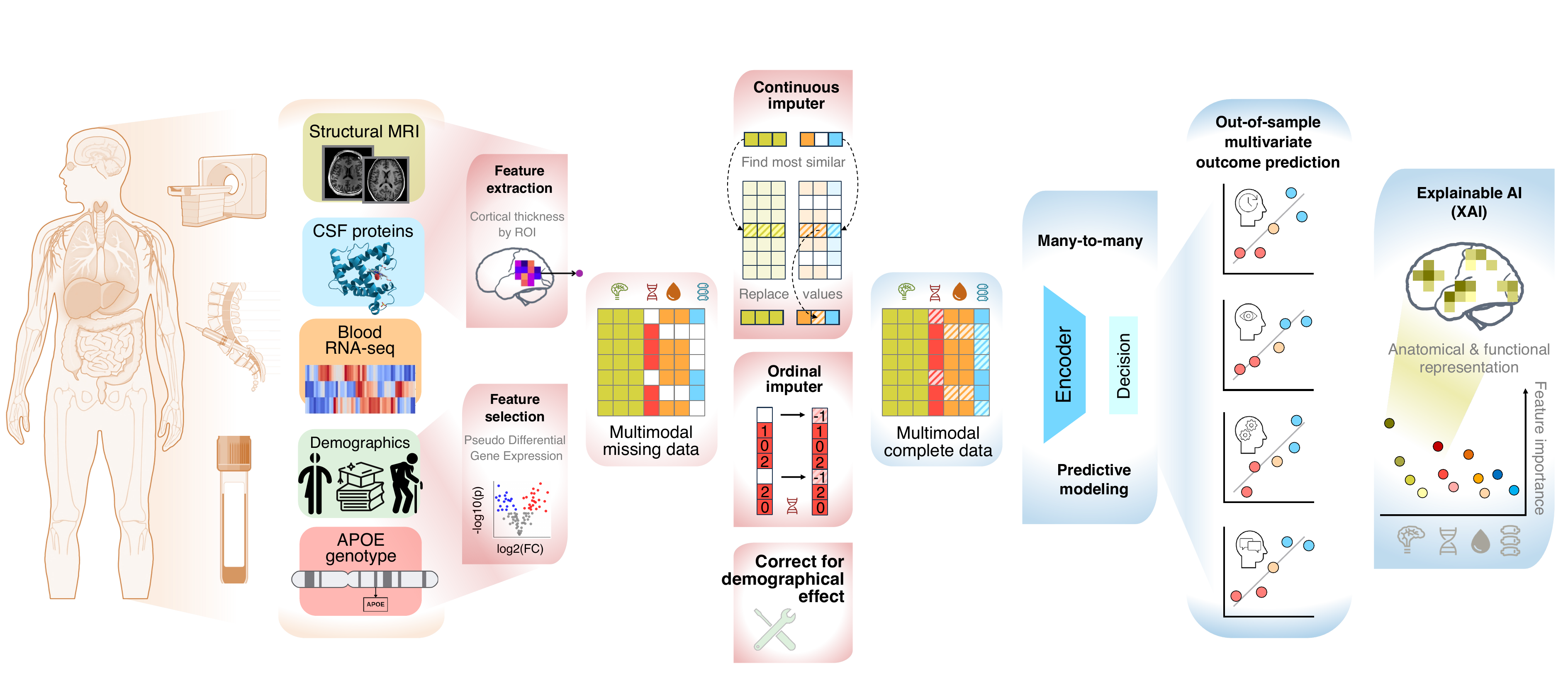} 
    \caption{\textbf{The schematic representation of the OPTIMUS architecture.} From left to right. (1) Neuroimaging (structural MRI), genetic (APOE genotype), Blood transcriptomics (RNA sequencing or RNA-seq), proteomic (CSF data including phosphorylated tau (p-tau), total tau (t-tau), and amyloid beta (A$\beta$)), and demographic information undergo pre-processing and feature extraction, and enter the OPTIMUS model. (2) OPTIMUS performs modality-specific missing data imputation. It generates imputed data for each modality whose modality-specific distributions resemble those of the observed data. (3) OPTIMUS performs many-to-many prediction. It predicts multivariate outcomes using multimodal data. (4) Explaining selected biomarkers via explainable AI (XAI). OPTIMUS numerically quantifies the explainability of the selected features using permutation importance scores, and, by plotting the weights of the top features back to the anatomical space, it assesses their clinical and pathological relevance.
    }
    \label{fig:pipeline}
\end{figure*}

Here, reconciliating the increasing health and economic challenges caused by AD and the need to improve the diagnosis, management, and potential treatment of the disease, we propose a comprehensive ML framework, called OPTIMUS, to study the many-to-many problem to discover potential multimodal AD biomarkers and predict multivariate disease outcomes. Specifically, to jointly address both technical and biological questions, OPTIMUS fuses missing data imputations, predictive modeling, multimodal data analysis, and explainable AI (XAI) (see Fig.~\ref{fig:pipeline}).

\begin{figure}[!ht]
    \centering
    \includegraphics[width=\linewidth]{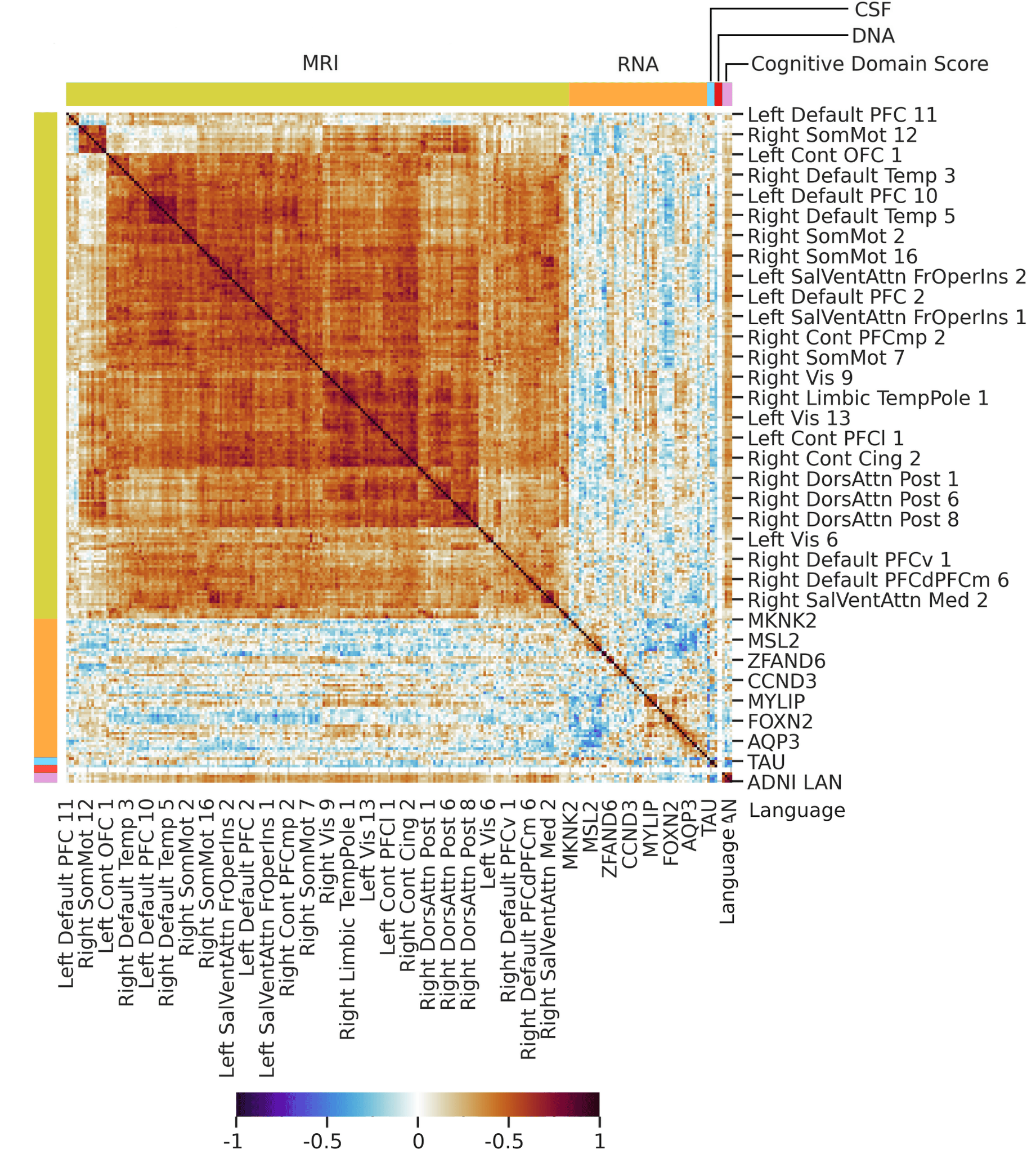}
    \caption[Intra- and inter-modality feature similarity.]{\textbf{Intra- and inter-modality feature similarity and their relationship to the multivariate outcomes.} Multimodal data show strong intra-modality similarity but weaker inter-modality similarity. This suggests that features from different modalities potentially offer complemental information. Concurrently, each modality contains features strongly associated with the outcomes; but for each outcome, the most relevant features vary within- and across-modality. This suggests there are potential multimodal biomarkers differentially predictive of the outcomes. Each entry of the matrix represents Pearson correlation between paired features, or that between a feature and an outcome. The colors of the rows and columns correspond to data modalities: MRI (cortical thickness) in yellow, RNA (blood transcriptomics) in orange, CSF in blue DNA in red, and cognitive domain scores in pink. Variables are: 200 brain regions of interest (MRI), 54 gene counts (RNA), 3 protein quantifications (CSF), 3 allele counts (DNA), and 4 cognitive domain scores. A hierarchical clustering was performed modality-wise. A further sub-clustering of the MRI data based on functional brain regions is in Fig.~\ref{fig:mri_correlation}.
     }
    \label{fig:features_correlation}
\end{figure}

\section{The Many-to-many Problem in AD}
\label{sec:problem_statement}

\subsubsection{Multimodal AD Biomarkers}
\label{subsubsec:intro_multimodal}

Whereas the definitive cause for most AD cases is largely unknown~\citep{querfurthAlzheimersDisease2010}, the disease has several biological associates. As such, multimodal data analysis in AD has shown increasing promises in AD prediction~\citep{elazabAlzheimersDiseaseDiagnosis2024b}:
each modality provides distinct pathological information and complements other modalities in explaining the total disease variability (see Fig.~\ref{fig:features_correlation}); their integration, therefore, enhances prediction accuracy compared to uni-modal data~\citep{thanhTensorKernelLearning2024}.

As a brain disease, most AD-related cognitive and behavioral malfunctions can be, in one way or another, traced back to the brain. Brain imaging techniques provide a direct look into the brain areas affected by the disease. Here, we focus on structural magnetic resonance imaging (sMRI) data because they measure brain morphometry and unveil structural biomarkers underpinning cognitive impairment, such as gray and white matter atrophy~\citep{vemuriRoleStructuralMRI2010}, and are relatively easy and more frequently available in practical examinations.

There is also a genetic basis for AD. While AD is recognized as a polygenic disease~\citep{harrisonPolygenicScoresPrecision2020}, APOE$\epsilon$4 allele remains the strongest single genetic predictor, and is  the most well-established genetic risk factor for late-onset AD~\cite{bellenguezGeneticsAlzheimersDisease2020a}. In this study, we incorporated both blood gene expression data and APOE genotype to investigate biological mechanisms that may influence AD progression and symptom variability. We do so to give attention to both the role of the peripheral immune system and to capture broader systemic factors that could contribute to AD heterogeneity. 

The CSF exchanges with the extracellular space of the brain, facilitating the transport of proteins and metabolites that may be closely linked to AD pathology~\citep{shawCerebrospinalFluidBiomarker2009a}. Here, we consider A$\beta$, phosphorylated tau (p-tau), and total tau (t-tau) in CSF data because they are well-established AD biomarkers~\citep{shawCerebrospinalFluidBiomarker2009a} and because we aim to examine to what extend they contribute to or complement other data modality in AD prediction. 

\subsubsection{Multivariate AD Outcomes}
\label{subsubsec:intro_multivariate}
AD disrupts multiple cognitive and behavioral functions, including memory, language, orientation, judgment, and problem-solving~\citep{morrisClinicalDementiaRating1997}. 
Neuropsychological tests, however, often provide an aggregated total score, summing sub-scores related to memory, language, etc., to evaluate disease status and severity. A total score is useful for describing the general disease profile, but it overlooks the nuanced effects of AD on individual cognitive subdomains. In practice, impairments in executive function, language, memory, and visuospatial function differ between individuals, leading to different symptoms, even between people with similar total disease scores. 
Recent studies using uni-modal imaging data~\citep{chenResidualPartialLeast2024b} suggest the potential of predicting multivariate disease outcomes. Extending these approaches to multimodal data for multivariate outcome prediction, however, remains an open challenge. Additionally, it is unclear which subset of biomarkers from each modality are associated with specific cognitive domains and which machine learning methods best identify and quantify the pathways linking the former to the latter.

\subsubsection{Missing Data in Multimodal AD Studies}
\label{subsubsec:intro_missing}

Missing data are ubiquitous in real-world data analysis and particularly in multimodal studies, where the integration of diverse data types increases the likelihood of incomplete datasets.~\citep{leMultimodalMissingData2025}. 

While one may use imputation strategies, such as linear interpolation~\cite{abdelazizAlzheimersDiseaseDiagnosis2021}, multiple imputation~\cite{chenPersonalizedLongitudinalAssessment2023}, and deep learning~\cite{thungMultistageDiagnosisAlzheimers2017}, to tackle missing data~\citep{ritterMultimodalPredictionConversion2015}, the primary challenge for these methods in handling multimodal data lies in replicating the underlying heterogeneous feature distributions for each modality using available data while maintaining robust predictive accuracy. This is further complicated when predicting multivariate AD outcomes are considered. In Section \ref{Sec:Results} and the Supplementary Materials, we conduct a thorough investigation and comparison of these methods.

\subsubsection{Machine Learning in AD Research}
\label{subsubsec:intro_ml}

ML methods, particularly supervised ML methods, in AD research generally serve two purposes: to predict disease outcomes, and to find biomarkers that are predictive of these outcomes. Methodologically, these methods can be classified into two classes: shallow models, such as support vector machines~\citep{ sharmaAlzheimersPatientsDetection2021}, random forests~\citep{velazquezRandomForestModel2021}, and $k$-nearest neighbors~\citep{zhangEnhancedHarrisHawks2023}, and deep learning methods, such as convolutional neural network~\citep{dakdarehDiagnosisAlzheimersDisease2024a}, recurrent neural network, Transformers~\citep{huVGGTSwinformerTransformerbasedDeep2023}, and recently, large language models~\citep{wangAugmentedRiskPrediction2024}.

Deep learning has become increasingly prominent in AD prediction using multimodal data~\citep{
qiuMultimodalDeepLearning2022}. Despite advances, they are facing challenges, especially in handling missing data in multimodal analyses and in interpreting the identified biomarkers. Relatedly, optimizing the balance between predictive accuracy and biological interpretability remains an open question. In Section \ref{Sec:Results}, we address these challenges and propose potential solutions.

\subsubsection{Explainable AI in AD Biomarker Discovery and Disease Prediction}
\label{subsubsec:intro_explainable}

Interpretability is crucial for the real-world application of ML methods in AD biomarker discovery and disease prediction. A useful way to interpret ML-derived biomarkers is to quantify the contribution each feature makes to predicting the outcome, such as disease group and disease severity, using XAI techniques~\citep{barredoarrietaExplainableArtificialIntelligence2020a}. For AD studies, in addition to quantifying feature weights, it is important to project features back to the anatomical (brain) space to explain which brain regions are involved in the prediction. Yet, given AD's complexity, with multimodal feature associates and multivariate disease outcomes, it remains unclear whether, and if so, to what extend XAI can address the many-to-many problem. Can XAI identify biomarkers uniquely associated with specific outcomes, or those shared across domains? Are the predictive neuroimaging markers randomly distributed or meaningfully located in anatomically and functionally relevant brain regions? In Section \ref{Sec:Results} and Fig.~\ref{fig:tabnet_target}, we investigate the explainability of ML-derived multimodal biomarkers in predicting multivariate outcomes.

%% file: Sections/2-Methods.tex
\normalsize

\begin{figure*}[htbp]
    \centering
    \includegraphics[width=0.94\linewidth]{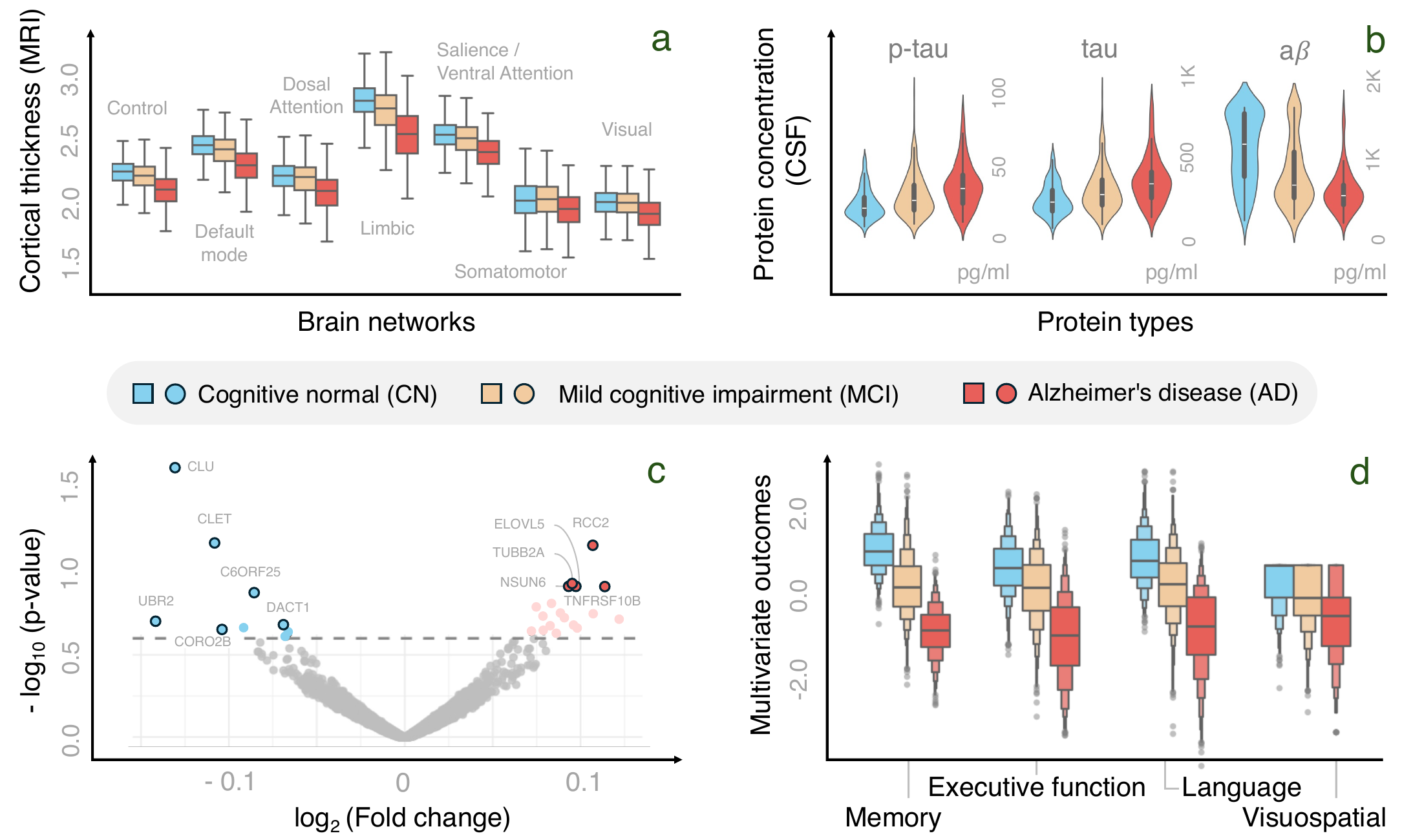}
    \caption[Inter-group feature and outcome differences.]{\textbf{Inter-group feature and outcome differences.}
    (a) Cortical thickness by functional brain network across CN, MCI and AD groups. The Cortical thickness generally decreases as AD progresses but differential across brain regions. (b) CSF biomarkers between CN, MCI and AD group. As AD advances, p-tau and tau increases and A$\beta$ decreases. (c) Genes differentially expressed in CN and AD. CN-MCI and between MCI-AD differentially expressed genes are in Fig.~\ref{fig:results_dge} (d) Multivariate cognitive scores across CN, MCI and AD groups. As AD progresses, scores related to memory, executive function, visuospatial, and language become worsen.}
\label{fig:data_description}
\end{figure*}

\subsection{The Dataset}
\label{subsec:meth_data}

We use data from the Alzheimer's Disease Neuroimaging
Initiative (ADNI)~\citep{petersenAlzheimersDiseaseNeuroimaging2010}. The data used in our study are from 1,205 participants, including 346 cognitively normal (CN) subjects, 608 subjects with mildly cognitive impairment (MCI), and 251 AD patients, with a total of 2,974 samples (see Table ~\ref{tab:demographics} and Fig.~\ref{fig:exploratory_analysis} for detailed demographic information). 

Trained clinicians classified participants as CN, MCI, or AD based on subject scores from the Clinical Dementia Rating (CDR) global and memory scales, Mini-Mental State Examination (MMSE), and the Wechsler Logical Memory II sub-scale. These classifications are primarily based on neuropsychological assessments and are further refined during clinical diagnostic evaluations conducted at the screening visit. The study excluded individuals whose dementia symptoms were likely due to other neurodegenerative disorders. Additional information about ADNI data and protocols are available on the ADNI website (\url{adni.loni.usc.edu}). 

To study the many-to-many problem, we consider four data modalities: RNA-seq, CSF, APOE genotype, and brain cortical thickness derived from structural MRI (see details in Section ~\ref{subsec:math_multimodal_data}), and four disease outcomes related to executive function, language, memory, and visuospatial function.

\subsection{Multimodal Features and Multivariate Outcomes} \label{subsec:math_multimodal_data}

\paragraph{Structural MRI}
We use preprocessed MRI images acquired with 1.5T and 3T scanners from ADNI. T1-MRI data were processed using the CAT12 toolbox~\citep{chenResidualPartialLeast2024b}, including inhomogeneity correction, voxel-based morphometry for spatial registration and segmentation into gray matter and white matter, and surface-based morphometry for cortical thickness estimation. Cortical thickness was calculated using high-dimensional spatial registration with a 200-parcel atlas~\citep{schaeferLocalGlobalParcellationHuman2018a}. The final data contain features from 200 regions of interests across 7 functional networks (Fig.~\ref{fig:data_combined_network_mri}).

\paragraph{Blood transcriptomics and APOE genotype}

We employed blood bulk RNA-seq data, obtained from Affymetrix expression microarray technology and normalized using Robust Multi-Array Average 
\citep{saykinGeneticStudiesQuantitative2015b}. Probes corresponding to the same genes were merged by retaining the maximum count value for each gene (see Supporting Information and  Fig.~\ref{fig:rnaseq_volcanoes}).
We then selected genes with the highest $\log_2$ Fold Changes as candidate features.
Given that the APOE genotype is the primary genetic risk factor for AD, we incorporated it into the analysis. APOE genotyping was performed using DNA extracted from blood samples~\cite{saykinGeneticStudiesQuantitative2015b}.

\paragraph{Cerebrospinal fluid}
We include A$\beta$, phophorylated tau (p-tau) and total tau (t-tau). The sample collection and processing of CSF data are available in ADNI procedure manuals~\citep{shawCerebrospinalFluidBiomarker2009a}.

\paragraph{Neuropsychological tests}

We consider four scores related to memory, executive function, language, and visuospatial function~\citep{craneCognitiveAssessmentsADNI2021a, choiDevelopmentValidationLanguage2020b}. The scores were obtained from trained clinicians who selected and merged tasks from several neuropsychological tests related to each domain and combined them using bi-factor models.

\subsection{The OPTIMUS Pipeline}
\label{subsec:meth_pipeline}

The OPTIMUS is a comprehensive and modular ML framework for predicting multivariate outcomes using multimodal data with missing values. It consists of three main components: missing data analysis, many-to-many predictive modeling, and XAI. 

Consider modality-specific biomarkers from structural MRI, CSF protein quantification, blood transcriptomics, and APOE genotype data. Define $\bm{X}$ as the multimodal feature data with $M = 4$ modalities: three continuous-valued ($\bm{X}^{(m)} \in \mathbb{R}^{N \times p_m}$ for $m \in \{1, 2, 3\}$) and one categorical (APOE genotype, $\bm{X}^{(4)} \in \{0, 1, 2\}^{N \times p_4}$). Define $\bm{Y} \in \mathbb{R}^{N \times 4}$ as four cognitive scores related to executive function, language, memory, and visuospatial ability. 
Define the set of observed indices as $\mathcal{O} = \{(i, j, m) \mid x_{ij}^{(m)} \text{ is observed}\}$, with sample index $i$, feature index $j$, and modality $m$.

\paragraph{Imputer}

Consider an imputer, $g_\mathcal{I}$, which outputs a completed dataset $\bm{\hat{X}}= g_\mathcal{I}(\bm{X}, \mathcal{O})$:
\begin{equation*}
    \hat{x}^{(m)}_{ij} = 
    \begin{cases}
    x_{ij}^{(m)}, & \text{if } (i, j, m) \in \mathcal{O} 
    \\
    g_{ij}^{(m)}(\bm{X};\mathcal{O}), & \text{otherwise.} 
    \end{cases}
\end{equation*}

In order to select the optimal imputation method, we evaluated different imputation methods using the Kullback-Leibler (KL) divergence between the modality-specific distributions of the imputed samples and those of the observed data. We chose the imputer $g_{\mathcal{I}}^*$ that produced the lowest KL divergence to best preserving the underlying data distribution. See Supporting Information for more details.

\paragraph{Predictor}
Given the complete data $\bm{\hat{X}}^*:= g_\mathcal{I}^*(\bm{X}, \mathcal{O})$ and confounders $\bm{Z} \in \mathbb{R}^{N \times 3}$ (age, gender, and education), OPTIMUS's optimal predictor $f^*$ tries to satisfy:
\begin{equation*}
    \underset{ f  }{\mathrm{argmin}} 
    \left \|
    \bm{Y} - 
    f\left(
  \bm{\epsilon}_{\hat{\bm{X^*}}\vert \bm{Z}}
    \right) - \bm{Z}\hat{\bm{\beta}}
    \right \|_F,
\end{equation*}
Such an $f^*$ is TabNet~\citep{arikTabNetAttentiveInterpretable2020} a transformer-based neural network architecture (see Supplementary Information), 
$\epsilon_{\hat{\bm{X}}^*\vert \bm{Z}} = \bm{\hat{X}}^* - \bm{Z} \hat{\bm{\gamma}}$;
and $\hat{\bm{\gamma}}$ and $\hat{\bm{\beta}}$ are OLS estimators for $\hat{\bm{X}}^* = \bm{Z} \bm{\gamma} + \epsilon_{\hat{\bm{Y}}\vert \bm{Z}}$ and $\bm{Y} = \bm{Z} \bm{\beta} + \epsilon_{\bm{Y}\vert \bm{Z}}$, respectively, in residual learning.

The final predictive multivariate outcome is then:
\begin{equation*}
\hat{\bm{Y}} = f^* \left ( \bm{\epsilon}_{\hat{\bm{X}}^* \mid \bm{Z}} \right) + \bm{Z}\hat{\bm{\beta}}.
\end{equation*}

\paragraph{Explainer}
OPTIMUS evaluates biomarkers' interpretability by examining both features importance and biological relevance. For feature importance, OPTIMUS calculates, for example, a feature importance score, $I_j$, for feature $j$. Here, we use the permutation score (but see Section \ref{Sec:Discussion} for a discussion and the Supporting Information for alternative scores):
\begin{equation*}
\label{eq:permutation_test}
    I_j =  \mathbb{E}_{\pi} 
    \left \{ 
    r \left (\bm{Y}, \hat{\bm{Y}} (\hat{\bm{X}}^*) \right) - r \left(\bm{Y}, \hat{\bm{Y}} (\hat{\bm{X}}^*_{\pi(j)}) \right) 
    \right \},
\end{equation*}

where $r$ denotes Pearson's correlation coefficient between the observed outcome $\bm{Y}$ and its prediction $\hat{\bm{Y}} (\hat{\bm{X}}^*)$ (resp. $\hat{\bm{Y}} (\hat{\bm{X}}^*_{\pi(j)}$) using $\hat{\bm{X}}^*$ (resp. $\hat{\bm{X}}^*_{\pi(j)}$);
and~$\pi$ is a random permutation where $\hat{\bm{X}}^*_{\pi(j)}$ represents a dataset with feature $j$ shuffled across rows while others intact. 

Subsequently, OPTIMUS projects the weights of feature with anatomical references back to anatomical/spatial space; for features without specific references, OPTIMUS outputs their weights in association with each prediction to investigate their biological underpinning.

Technically, the OPTIMUS imputer, predictor, and explainer each solves a task-specific (sub)-optimization problem, and together, aim to solve a global optimization problem accounting for predictability and explainability. Specifically, OPTIMUS conducts model comparisons and choose the best model for each suboptimisation problem; it then iteratively evaluates the combinations of methods that optimize predictability and explainability and output the optimal combinations of methods.

%% file: Sections/3-Results.tex
Here, we conduct model comparisons for methods used, respectively, for multimodal missing data imputation, many-to-many predictive modeling, and XAI. Specifically, we assess different imputers' ability to retrieve underlying modality-specific distributions using visual, statistical, and predictive assessments. For many-to-many predictive modeling, we evaluate the prediction accuracy of different predictive methods by quantifying the differences between the observed and predicted multivariate outcomes. To explain the selected biomarkers, we apply XAI techniques to top-performing models and evaluate neural features' anatomical/functional specialization and other top features' biological implication. 

In brief, our results suggest that a combination of 1-nearest neighbors (imputer), TabNet (predictor; a Transformer-based neural network~~\citep{arikTabNetAttentiveInterpretable2020}), and permutation importance (explainer) optimize prediction accuracy and biomarker explainability. Tables ~\ref{tab:imputation_corr} and ~\ref{tab:imputation_mae} show missing data analysis result using different imputation methods. Tables ~\ref{tab:prediction_test_train_corr} to \ref{tab:prediction_loona_mae} exhibit many-to-many prediction results using different predictive methods. Fig.~\ref{fig:tabnet_target} presents the anatomical and functional locations of the top neural features and the top genetic signatures.

\subsection{Multimodal Missing Data Imputation and Comparisons} 
\label{subsection:missing_data}

\begin{figure}[h]
    \centering
    \includegraphics[width=\linewidth]{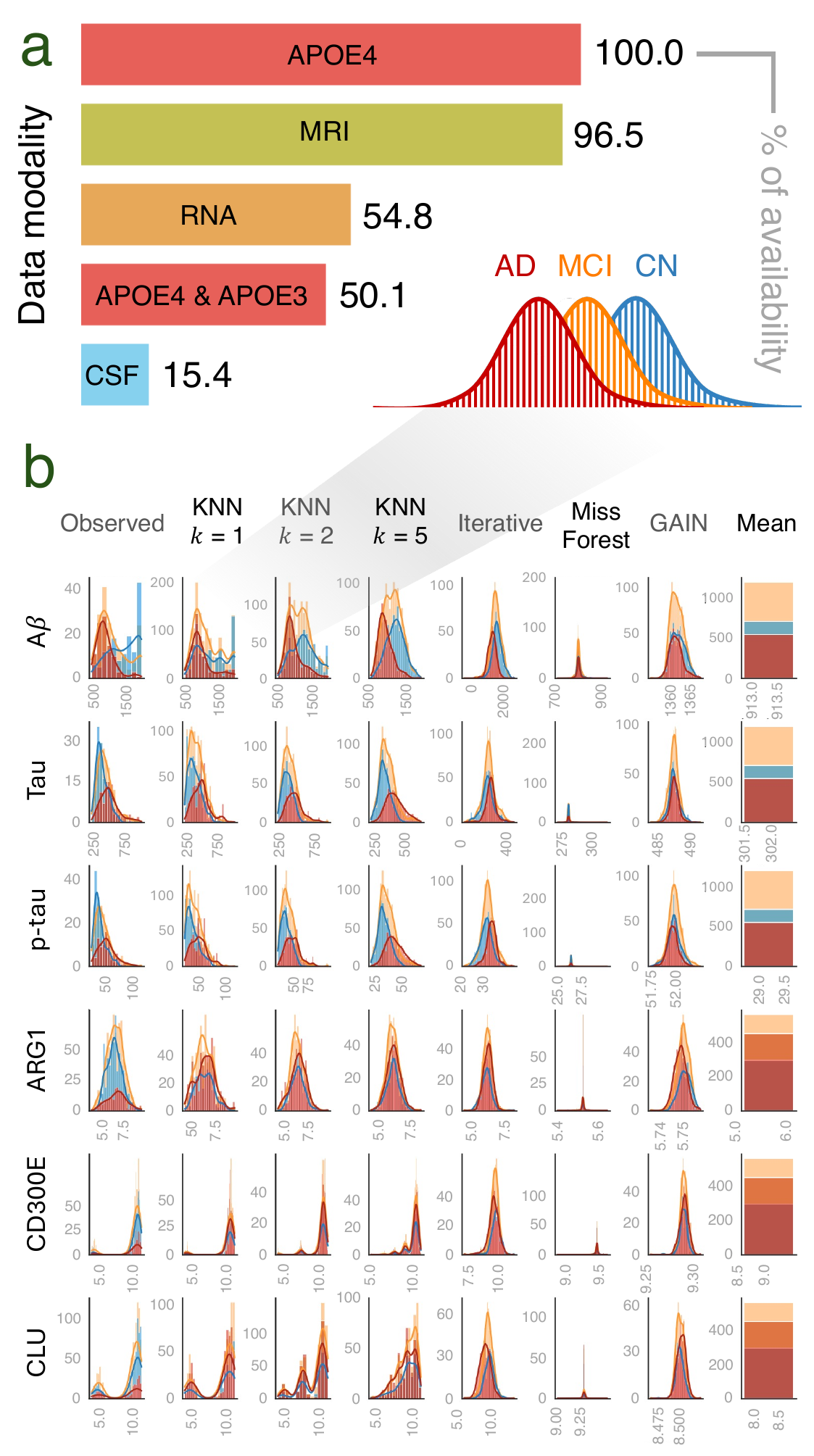}
    \caption[Multimodal missing data analysis.]{\textbf{Multimodal missing data analysis.} {(a) Percentage of available data across modalities}. 
    (b) Modality-specific missing data imputation. First row contains histograms and kernel density estimation (KDE) curves for A$\beta$, phosphorylated tau (p-tau), and t-tau from CSF, and AGR1, CD300E and CLU gene counts from blood RNA-seq expression data. The first column corresponds to observed data; each subsequent column represent distributions from a specific imputer (see Table ~\ref{tab:imputation}), ranked based on the KL-distance between the observed and imputed distributions. Within each subplot, the x-axis represents the feature values and the y-axis shows frequency; the color codes for the distributions are: red~= AD, orange~= MCI, and blue~= CN.}
\label{fig:data_imputation}
\end{figure}

We considered various algorithms for multimodal missing data imputation, including mean imputation, $k$-nearest neighbors (KNN), multiple imputation by chained equations~(MICE), MissForest, and generative adversarial imputation networks (GAIN)~\citep{stekhovenMissForestNonParametric2012a} (see technical details in Table~\ref{tab:imputation}).

We conducted three types of assessments - visual, statistical, and predictive. Visually, we used histograms, ECDFs, and KDEs to compare the distributions of imputed data with their observed counterparts (Figs.~\ref{fig:data_imputation}, ~\ref{fig:imputation_heatmaps} and~\ref{fig:imputation_edf}). The KNN imputer ($k=1$) performed best: the data distributions showed minimal divergence from the observed distributions (Fig.~\ref{fig:imputation_heatmaps}).
Statistical assessments using KL divergence further confirmed this. In predictive analysis, data from KNN imputation achieved the highest accuracy (both in terms of correlation and MAE) in multivariate outcome prediction. For ordinal features (e.g., APOE genotype), the imputation method had little impact on the prediction performance (Tables~\ref{tab:imputation_corr} and~\ref{tab:imputation_mae}). For these data, we used a constant imputer with $-1$ representing missingness.

\subsection{Many-to-many Predictive Models and Comparisons} 
\label{subsec:model_comparison_predictive}
We tested several prominent ML methods for predicting multivariate AD outcomes using multimodal data, corrected for age, gender, and education. In our data, 12 subjects have complete data and 1,193 subjects have data containing missing values. Thus, we evaluated the models using three cross-validation scenarios. (1) A standard train-test split. We trained models on data from 1,193 subjects containing imputed values and tested them on data from 12 subjects whose data are fully observed. (2) A leave-one-complete (subject)-out (LOCO-CV). We trained on 1,193 plus 11 subjects and test on fully observed data from each of the 12 subjects, where each of the 12 subjects was tested once. (3)~Leave-one-missing-out (LOMO-CV).
Scenarios (1) and (2) considered fully observable data in the testing set; scenario (3) considers realistic cases where testing data also have missing values. Here, we randomly selected 20 subjects with missing data. We trained an imputer $g_\mathcal{I}^*$ on multimodal imputed data $\tilde{\bm{X}}^{\text{training}}$ from 1,185 training subjects and a predictor $f^*$ using $\tilde{\bm{X}}^{\text{training}}$ and $\tilde{\bm{Y}}^{\text{training}}$. For the test data, we first obtain the imputed multimodal imputed data using $g_\mathcal{I}^*$ learned from the training data (to avoid leakage), and then predict the outcomes from one testing subject using $f^*$, without further training (to also avoid leakage); each of the 20 testing subjects was tested once. See Fig.~\ref{fig:crossvalidation} for cross-validation settings.

For the train-test split, TabNet with a KNN ordinal imputer achieved the best performance, with a correlation coefficient $r=0.716$ and a mean MAE of $0.543$, followed by XGBoost with a constant ordinal imputer ($r=0.662$, $\text{MAE} = 0.602$) and Random Forest ($r=0.625$, $\text{MAE}=0.639$) (Tablew ~\ref{tab:prediction_test_train_corr} and ~\ref{tab:prediction_test_train_mae}). In the LOCO-CV setting, TabNet again outperformed the other models, achieving $r=0.744$ and $\text{MAE} = 0.445$, compared to XGBoost's ($r=0.553$, $\text{MAE} = 0.656$) and Random Forest ($r=0.669$, $\text{MAE} = 0.595$) (Tables ~\ref{tab:prediction_loco_main}, ~\ref{tab:prediction_loonona_corr} and ~\ref{tab:prediction_loonona_mae}). Finally, in the LOMO-CV scenario -- where imputation biases were more pronounced -- TabNet with a constant imputer performed best ($r=0.492$, $\text{MAE} = 0.445$), while XGBoost and Random Forest achieved $r=0.452$ with $\text{MAE} = 0.509$ and $r=0.457$ with $\text{MAE} = 0.514$, respectively (Tables ~\ref{tab:prediction_loona_corr} and ~\ref{tab:prediction_loona_mae}).

Taken together, our results suggest that TabNet with KNN imputation was the most effective for performing many-to-many prediction. Nevertheless, the biological relevance of the selected biomarkers remain to examined, which we address with XAI in the following section. 

\begin{table*}[!ht]
  \fontsize{8pt}{10pt}\selectfont
  \centering
\caption[Pearson correlation scores of regression models comparison in leave-one-complete-out cross-validation]{
    \textbf{Model performance in many-to-many prediction.} Values represent the Pearson correlation score between multivariate observed and predicted scores related to executive function, language, memory, and visuospatial function. Imputations for ordinal values are done using methods from the first column; imputations for continuous values are done using 1-Nearest Neighbor. Leave-one-out cross-validation was carried out on 20 randomly selected complete test samples. Mean and standard deviation (Std) were calculated row-wise across four cognitive scores. Confounders were adjusted before final prediction.
    } 
    \begin{tabular}{|p{4cm}p{2.5cm}||p{1.1cm}p{1.1cm}p{1.1cm}p{1.1cm}||p{0.6cm}p{0.6cm}|}
        \hline
\textbf{Model}
 & 
\textbf{Modality}
& 
\textbf{Executive Function}
& 
\textbf{Language}
& 
\textbf{Memory}
& 
\textbf{Visuo-spatial} & \textbf{Mean} & \textbf{Std} \\ \hline
        \hline
        Linear Regression & Best Unimodal & 0.138 & 0.346 & 0.604 & -0.304 & 0.196 & 0.384 \\ \hline
        Linear Regression & Multimodal  & 0.241 & 0.433 & 0.630 & -0.308 & 0.249 & 0.404 \\ \hline
        Multi-task Elastic-Net & Best Unimodal & -0.486 & -0.190 & 0.462 & 0.488 & 0.068 & 0.485 \\ \hline
        Multi-task Elastic-Net & Multimodal  & -0.017 & 0.307 & 0.143 & 0.134 & 0.142 & 0.132 \\ \hline
        Multi-task Lasso & Best Unimodal & -0.486 & -0.190 & 0.462 & 0.488 & 0.068 & 0.485 \\ \hline
        Multi-task Lasso & Multimodal  & -0.017 & 0.307 & 0.143 & 0.133 & 0.142 & 0.132 \\ \hline
        PLS Regression - 2 Components & Best Unimodal & 0.380 & 0.669 & 0.643 & -0.059 & 0.408 & 0.338 \\ \hline
        PLS Regression - 2 Components & Multimodal  & 0.379 & 0.671 & 0.622 & -0.030 & 0.411 & 0.321 \\ \hline
        PLS Regression - 4 Components & Best Unimodal & 0.342 & 0.686 & 0.609 & -0.041 & 0.399 & 0.328 \\ \hline
        PLS Regression - 4 Components & Multimodal  & 0.361 & 0.702 & 0.614 & -0.048 & 0.407 & 0.336 \\ \hline
        Random Forest Regressor & Best Unimodal & 0.433 & 0.760 & 0.682 & -0.115 & 0.440 & 0.395 \\ \hline
        Random Forest Regressor & Multimodal  & 0.408 & 0.719 & 0.695 & -0.086 & 0.434 & 0.374 \\ \hline
        XGBoost Regressor & Best Unimodal & 0.478 & 0.669 & 0.703 & -0.093 & 0.439 & 0.368 \\ \hline
        XGBoost Regressor & Multimodal  & 0.591 & 0.707 & 0.708 & -0.197 & 0.452 & 0.436 \\ \hline
        TabNet Regressor & Best Unimodal & 0.135 & 0.469 & 0.403 & -0.127 & 0.220 & 0.273 \\ \hline
        \textbf{TabNet Regressor} & \textbf{Multimodal} & \textbf{0.402} & \textbf{0.692} & \textbf{0.679} & \textbf{0.194} & \textbf{0.492} & \textbf{0.240} \\ \hline
    \end{tabular}

    \label{tab:prediction_loco_main}
\end{table*}

\normalsize

\begin{figure*}[!ht]
    \centering
    \includegraphics[width=1    \textwidth]{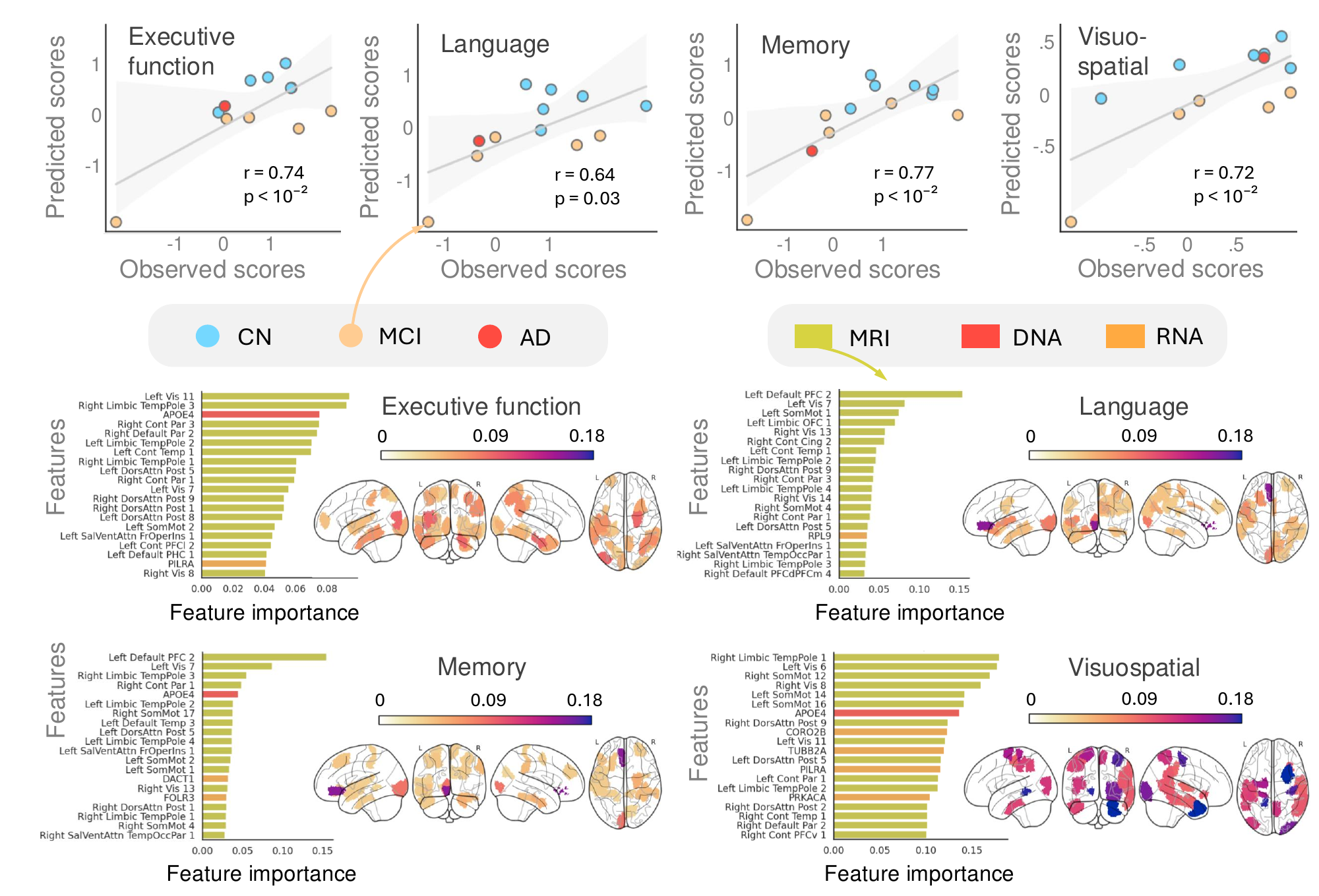}
    \caption{\textbf{Interpret multimodal biomarkers predictive of multivariate outcomes via explainable AI (XAI).} Top Panel. The scatter plots from left to right show predicted scores (Y-axis) against observed scores (X-axis) for executive function, language, memory, and visuospatial score, respectively. Blue~=CN; yellow~=MCI; red~=AD. Each line assesses the goodness-of-fit, and the shaded band indicates the 95\% confidence interval. Bottom panel. Each bar chart indicates the top 20 features with highest feature importance scores, derived from a permutation test (averaged over 10 iterations of feature shuffling), for predicting executive function, language, memory, and visualization function, respectively. Bars were colored according to their modality (MRI - cortical thickness: yellow; DNA - APOE genotype: red, RNA - blood transcriptomics: orange). The top neuroimaging features were further projected to the Schaefer Atlas (200 regions across 7 networks) in four views. 
    }
    \label{fig:tabnet_target}
\end{figure*}

\subsection{Interpreting Biomarkers via XAI} \label{subsec:xai}

While the selected multimodal biomarkers predict multivariate AD outcomes, do they make biological sense? To evaluate this, we need a dual approach: on the one hand, one needs to quantify the importance of the selected biomarkers; on the other hand, for biomarkers that have a spatial correspondence, such as neuroimaging biomarkers, we need to place them in their anatomical and functional locations to evaluate their biological relevance.

To do so, OPTIMUS quantifies feature importance using the permutation importance $I_j$ (also see Supporting Information for alternative scores); in parallel, it interprets neuroimaging biomarkers by projecting them back to the brain space. The results from OPTIMUS provide us with several insights. First, our analysis suggests neuroimaging features are the most predictive biomarkers for cognitive performance prediction. Second, besides confirming that biomarkers in temporal and parietal regions are broadly associated with global cognitive decline~\citep{piniBrainAtrophyAlzheimers2016a, chenResidualPartialLeast2024b}, OPTIMUS identifies brain regions differentially predictive of four AD-related scores (Fig.~\ref{fig:tabnet_target}). 
For predicting executive function, the identified features span both hemispheres, including the prefrontal cortex (around BA10 and BA46, which are related to executive function), the superior and inferior temporal gyri, and the prestriate cortex.
For predicting language scores, important features are found around Broca's area, which is related to speech production and possibly comprehension, as well as the superior temporal gyrus next to it, and a small part of the superior frontal gyrus (left hemisphere).
For predicting memory scores, the key features bear some similarity with those predictive of the language score; but the features have pronounced presentation in the dorsolateral prefrontal cortex (responsible for working memory), especially on the left hemisphere.
Finally, for predicting visuospatial function, key contributions are made from the parietal cortex (responsible for spatial sense and spatial awareness) and the visual cortex.
Third, genetically, several genes were identified as key predictors across the different cognitive domains, with some overlapping features observed between domains (Fig.~\ref{fig:tabnet_target}). The APOE$\epsilon$4 genotype emerged as a top predictor across all cognitive domains, reinforcing its role as a major genetic risk factor for AD~\citep{bellenguezGeneticsAlzheimersDisease2020a}. In parallel, several blood-transcriptomics biomarkers demonstrated predictive importance. Two genes  --  DACT1 and FOLR3  --  stood out as top features for predicting memory score. DACT1 is a key regulator of the Wnt signaling pathway, which controls various aspects of neuronal differentiation and neuro-development. 
Higher FOLR3 expression is linked to lower plasma homocysteine levels, a risk factor for AD~\citep{yoshitomiPlasmaHomocysteineConcentration2020a}, with a possible protective effect against neuro-degeneration.
An immune-related gene, PILRA, previously suggested as a potential protective gene for AD~\citep{lopatkolindmanPILRAPolymorphismModifies2022a}, was identified to be an important predictive biomarker for both executive function and visuospatial scores. 
CORO2B, TUBB2A, and PRKACA genes were highlighted as key features for predicting visuospatial scores. CORO2B is involved in actin cytoskeleton organization, likely impacting neuronal actin structure and has also been identified as a differentially methylated site in AD patients~\citep{renIdentificationMethylatedGene2020a}.  
TUBB2A is a major component of micro-tubules, 
and was found to be differentially expressed in AD patients' brains, particularly in postsynaptic compartments~\citep{zolochevskaPostsynapticProteomeNonDemented2018a}.
Finally, RPL9, previously reported as down-regulated in the brains of AD patients~\citep{shigemizuIdentificationPotentialBlood2020a}, was identified as an important biomarker for predicting the language score.


Together, the biological interpretation of the biomarkers assisted by XAI not only confirms these genetic and neuroimaging biomarkers' relevance to AD in existing literature, but also sheds new light on how they together may jointly, but differentially, contribute to the prediction of multivariate AD-related scores. 


%% file: Sections/4-Discussion.tex
In this study, we proposed OPTIMUS, a comprehensive ML framework, to predict multivariate disease outcomes using multimodal data with missing values. Employing OPTIMUS, we identified multimodal neuroimaging, genetic, and blood transcriptomics data jointly yet differentially predictive of multivariate cognitive scores related to executive function, language, memory, and visuospatial function. We also showed that the multivariate outcome prediction accuracy is higher using multimodal data than using even the most predictive single modality data (Table~\ref{tab:prediction_loco_main}).

Although previous studies indicate LightGBM and XGBoost as the best-performing models in various scenarios~\citep{shwartz-zivTabularDataDeep2022}, particularly for AD diagnosis using ADNI data, neuropsychological scores~\citep{chakrabortyANALYZEADComparativeAnalysis2024a}, brain imaging~\citep{
leandrouCrosssectionalStudyExplainable2023a}, and multimodal features~\citep{hernandezExplainableAIUnderstanding2022a}, our results suggest that deep learning is competitive, especially when dealing with many-to-many problems.

This study has a few limitations. First, the initial step of feature selection based on differential gene expression analysis yielded no significant genes, due likely to the high noise levels in ADNI’s RNA-seq 
data~\citep{leePredictionAlzheimersDisease2020d} and the relatively small sample size~\citep{chen2023roles}. Nevertheless, blood biomarkers remain an area of growing interest in AD research thanks to their non-invasive nature and potential clinical utility. Future work may apply OPTIMUS to RNA-seq data obtained from more advanced methods with a larger sample size.

Second, data imputation introduced biases. While we demonstrated that modality-specific imputations recovered complex underlying data distributions and facilitated outcome prediction and interpretation, future work should compare our approach with alternative imputation techniques, such as denoising autoencoders~\citep{haridasAutoencoderImputationMissing2024b}, MUSE~\citep{wuMultimodalPatientRepresentation2023a}, and DrFuse~\citep{yaoDrFuseLearningDisentangled2024a}.

Third, although multivariate outcome prediction provides a more detailed characterization of AD by inquiring into domain-specific cognitive functions, many within-modality features are highly correlated and some cognitive scores interrelated. This makes the separation of domain-specific features less clear. Future studies may use OPTIMUS to predict finer-grained cognitive measures, such as raw sub-item scores from neuropsychological tests or 
scores that better capture within-domain and cross-subject variations~\citep{cacciamaniDifferentialPatternsDomainSpecific2022a}.


Fourth, feature importance analysis is essential for understanding the biological and pathological relevance of biomarkers and the interpretability of the model. Traditional XAI methods, such as permutation importance and Shapley values, may produce misleading results when features are highly correlated. Owen values, an extension of Shapley values that partitions features into correlated groups before computing importance scores, mitigate this issue by reducing distortions due to feature correlation and providing more realistic estimates~\citep{owenValuesGamesPriori1977a}. In our exploratory work, however, Owen values tended to overemphasize uncorrelated features, leading to nearly zero importance for many imaging features. Future work should explore alternative importance metrics, more efficient Owen value computation algorithms, or develop methods that better balance the consideration 
of correlated and uncorrelated features.

%% file: Sections/5-Conclusion.tex
This study demonstrates the potential of using machine learning to predict multivariate cognitive outcomes in AD using multimodal data with missing values. Leveraging large-scale neuroimaging, genetic, and proteomic data, OPTIMUS identifies explainable biomarkers predictive of cognitive and behavioral outcomes related to executive function, laguage, memory, and visuospatial function.

OPTIMUS provides a valuable platform to integrate multimodal data for uncovering novel pathways related to AD progression. Future work can, therefore, employ OPTIMUS to explore alternative mechanisms contributing to cognitive decline in AD, such as immune system dysregulation, metabolic disturbances, and gut-brain interactions. Additionally, improvements in each aspect of the OPTIMUS framework -- missing data handling, many-to-many predictive modeling, and explainability -- may further enhance its applicability.

To summarize, the ever-growing and -improving AI methods advance the analysis of multimodal data for biomarker discovery and AD prediction. In return, the expansion to include more diverse and high-quality heterogeneous datasets will facilitate the development of AI methods that are interpretable, clinically relevant, and potentially personalized. United, these continuous efforts are contributing to more accurate, biologically informed predictions and decision-making in AD diagnosis and prognosis, and, perhaps one day, better treatment. 

%% file: Sections/6-Supplementary.tex
\section{Methods}

\subsection{Further information about ADNI data}
\label{sec:supp_adni}

ADNI is funded by the National Institute on Aging, the National Institute of Biomedical Imaging and Bioengineering, and through generous contributions from the following: AbbVie, Alzheimer’s Association; Alzheimer’s Drug Discovery Foundation; Araclon Biotech; BioClinica, Inc.; Biogen;
Bristol-Myers Squibb Company; CereSpir, Inc.; Cogstate; Eisai Inc.; Elan Pharmaceuticals, Inc.; Eli Lilly and Company; EuroImmun; F. Hoffmann-La Roche Ltd and its affiliated company Genentech, Inc.; Fujirebio; GE Healthcare; IXICO Ltd.; Janssen Alzheimer Immunotherapy Research \& Development, LLC.; Johnson \& Johnson Pharmaceutical Research \& Development LLC.; Lumosity; Lundbeck; Merck \& Co., Inc.; Meso Scale Diagnostics, LLC.; NeuroRx Research; Neurotrack Technologies; Novartis Pharmaceuticals
Corporation; Pfizer Inc.; Piramal Imaging; Servier; Takeda Pharmaceutical Company; and Transition Therapeutics. The Canadian Institutes of Health Research is providing funds to support ADNI clinical sites in Canada. Private sector contributions are facilitated by the Foundation for the National Institutes of Health (www.fnih.org). The grantee organization is the Northern California Institute for Research and Education, and the study is coordinated by the Alzheimer’s Therapeutic Research Institute at the University of Southern California. ADNI data are disseminated by the Laboratory for Neuro Imaging at the University of Southern \textit{California}.

\subsection{RNA-seq Data Processing and Differential Expression Analysis}
\label{sec:supp_rnaseq_processing}

We employed blood bulk RNA-seq data obtained from 744 participants. RNA sequencing data were generated using Affymetrix expression microarray technology and normalized using the Robust Multi-Array Average (RMA) method~\citep{saykinGeneticStudiesQuantitative2015b}.

The original microarray data contained 49,386 transcript count measurements, which were pre-processed by merging transcripts mapping to the same gene, retaining the maximum count value for each gene across all matching probes.

To remove low-expression genes commonly considered as background noise, we applied a bimodal Gaussian mixture model to the distribution of median gene expression counts across all samples. Two Gaussians were fitted with the following parameters: the low expression Gaussian had a mean of 2.694 and a standard deviation of 0.475, while the high expression Gaussian had a mean of 6.781 and a standard deviation of 1.978887. The mixture proportions were 0.386 and 0.613, respectively. The filtering threshold was set at the midpoint between the means of the two Gaussians, specifically \((2.694 + 6.781)/2\). All genes with median expression levels below this threshold were removed, resulting in a filtered gene expression matrix containing 11,866 genes across 744 participants.

\begin{figure*}[t!]
    \centering
    \begin{subfigure}[b]{0.48\textwidth}
        \centering
        \includegraphics[width=\linewidth]{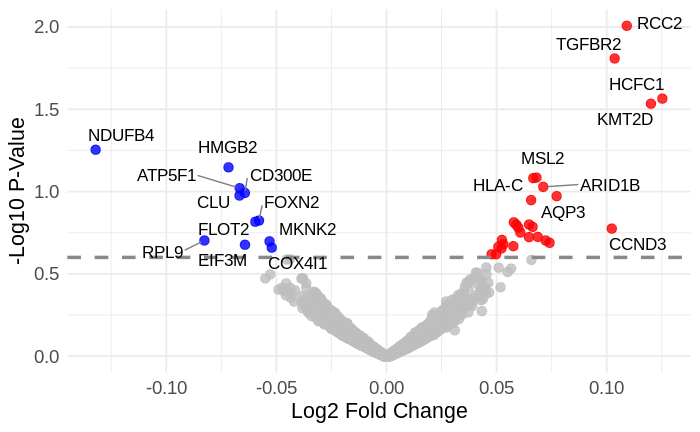}
        \caption{CN vs MCI.}
        \label{fig:volcano_CN_vs_MCI}
    \end{subfigure}
    \vspace{0.5cm}
    \begin{subfigure}[b]{0.48\textwidth}
        \centering
        \includegraphics[width=\linewidth]{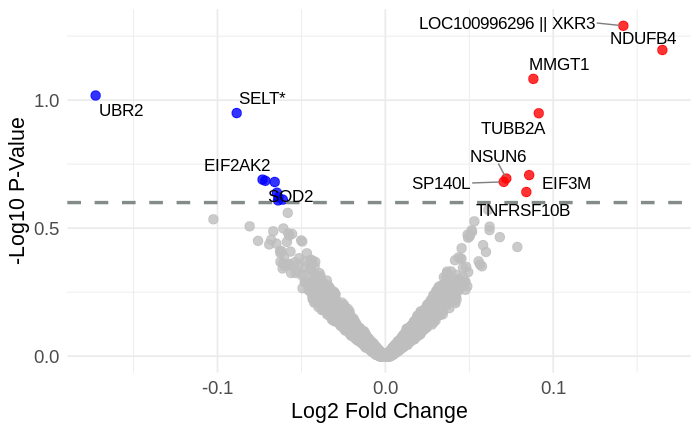}
        \caption{MCI vs AD.}
        \label{fig:volcano_MCI_vs_AD}
    \end{subfigure}
    \vspace{0.5cm}
    \begin{subfigure}[b]{0.48\textwidth}
        \centering
        \includegraphics[width=\linewidth]{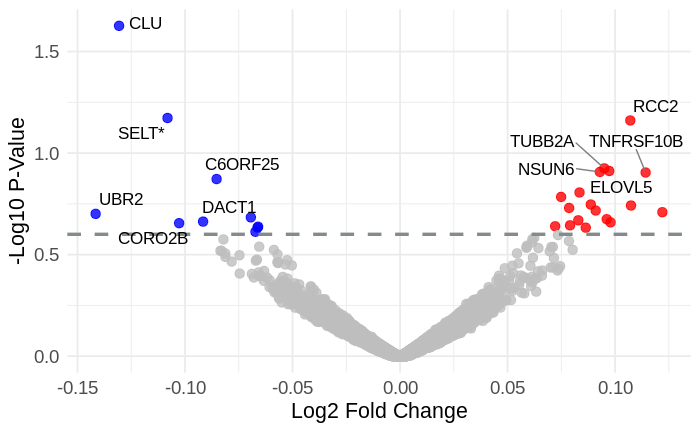}
        \caption{CN vs AD.}
        \label{fig:volcano_CN_vs_AD}
    \end{subfigure}

    \vspace{0.5cm}
    \caption[Volcano plots of RNA-seq analyses across cognitive groups]{
        \textbf{Volcano plots of RNA-seq analyses.} Volcano plots depicting differential gene expression analyses between cognitive groups done with edgeR. Each point represents a gene, with the x-axis displaying the $\log_2$ Fold-Change and the y-axis displaying the $- \log_{10}(p\text{-value})$ across three pairwise tests: (a) CN vs MCI, (b) MCI vs AD, and (c) CN vs AD. An empirical threshold of $- \log_{10}(p\text{-value}) = 0.6$ was used to highlight genes of interest (dashed horizontal line), focusing on those with the highest $\log_2$ Fold-Change (blue: downregulation, red : upregulation, gray: non-retained genes). While no genes met the commonly used false discovery rate significance thresholds (0.05 o 0.01), highlighted genes and the associated gene counts were used as transcriptomic features in downstream prediction tasks.  
    }
    \label{fig:rnaseq_volcanoes}
\end{figure*}

Differential gene expression analysis was then conducted using the \texttt{edgeR} package, comparing gene expression profiles between diagnostic groups (CN, MCI, AD). Genes were ranked by $log_2$ Fold-Change, and those with the largest absolute fold changes were selected as candidate features for downstream analysis. The corresponding volcano plots highlighting these selected genes are shown in Figure~\ref{fig:rnaseq_volcanoes}.

Only a single RNA-seq measurement was available per participant, typically collected during the first year of their enrollment in ADNI. For participants with repeated observations, we propagated their baseline gene expression measurements across all subsequent timepoints, effectively treating the transcriptomic data as static baseline profiles.

\begin{figure*}[t!]
    \centering
    \centerline{\includegraphics[width=0.9 \linewidth]{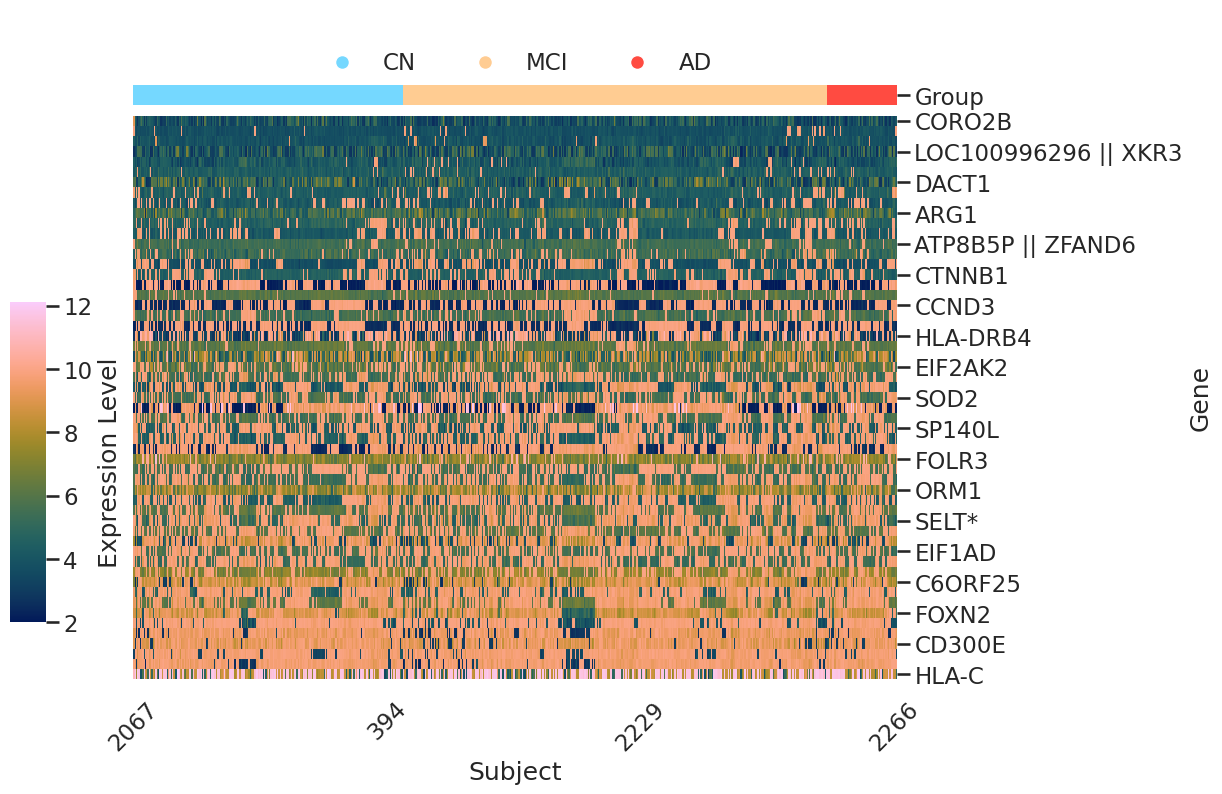}}
    \caption[Heatmap of Expression Profiles of Retained Genes. ]{\textbf{Heatmap of Expression Profiles of Retained Genes. } 
    This heatmap shows the expression levels of the 57 genes with the largest $\log_2$ Fold-Change based on blood bulk RNA-seq data from 744 participants.

    The original RNA micro-array data contained 49386 transcripts counts measurements that then were merged by retaining the maximum count value if they mapped to the same gene. Then to filter low-expression genes commonly considered as background expression, two gaussian with $m1=2.69; sd1=0.48; m2=6.78; sd2=1.98; p1=0.39; p2=0.614$ fitted to the bimodal gaussian distribution of median gene counts and chose as final threshold the crossing point of these two curves so at $threshold = (2.69 + 6.781) / 2$. So after removing all gene counts with values lower than this threshold, we ended up with filtered gene counts matrix of 11866 genes for our 744 subjects to be processed for differential gene expression in \texttt{edgeR}. The y-axis lists genes sorted by their mean expression values, and the x-axis represents subjects, identified by their IDs and sorted by cognitive groups (blue : CN, yellow : MCI, and red : AD). Genes were selected based on an empirical threshold of $- \log_{10}(p\text{-value}) = 0.6$ to select genes with high enough $\log_2$ Fold-Change after differential gene expression analysis using \texttt{edgeR}. Although no genes passed the FDR=0.05 significance threshold  for differential expression after false discovery rate correction, these genes were retained for downstream prediction tasks. }
    \label{fig:data_heatmap_transcriptomics}
\end{figure*}

The final set of selected genes, retained after filtering and differential expression ranking, are shown in the heatmap in Figure~\ref{fig:data_heatmap_transcriptomics}, where gene expression patterns are compared across the CN, MCI, and AD groups.

\begin{table*}[ht]
  \fontsize{8pt}{10pt}\selectfont
  \centering
  \caption[Participant Enrollment and Demographic Characteristics]{\textbf{Participant Enrollment and Demographic Characteristics.} This table summarizes the demographic characteristics of the participants included in the study. A total of N = 1205 participants were enrolled, with the following group distributions: Cognitive Normal (CN, N = 346), Mild Cognitive Impairment (MCI, N = 608), and Alzheimer's Disease (AD, N = 251). The table shows the number of participants in each group based on biological sex, age category, years of education, and APOE4 genotype. Sex, age, and educational attainment were regressed out of the training and testing data post-imputation, with APOE genotype treated as an ordinal (categorical) feature. }
\begin{tabular}{|p{5cm}||p{3cm}|p{3cm}|p{3cm}|}
 \hline
 \textbf{Total} & \textbf{CN} & \textbf{MCI} & \textbf{AD} \\ 
 N = 1205 & N = 346 & N = 608 & N = 251 \\ 
 \hline
 \rowcolor{lightgray}
 Sex & & & \\
 Female & 177 & 249 & 113 \\ 
 Male & 169 & 359 & 138 \\
 \rowcolor{lightgray}
 Age & & & \\
 $<$60 & 4 & 29 & 13 \\
 60-70 & 67 & 180 & 52  \\
 70-80 & 216 & 283 & 129 \\
 80-90 & 59 & 114 & 54 \\
  $>$ 90 & 0 & 2 & 3 \\
 \rowcolor{lightgray}
 Education (years) &  & & \\
 $<$ 12 & 7 & 25 & 9 \\
 12-16 & 100 & 187 &  96 \\
  $>$ 16  & 239 & 396 & 146\\
 \rowcolor{lightgray}
 APOE4 status & & & \\ 
 0 allele & 250 & 313 & 76\\
 1 allele & 89 & 227 & 118\\
 2 alleles & 7 & 68 & 57\\
 \hline
\end{tabular}
\label{tab:pipeline}
 \label{tab:demographics}
\end{table*}

\subsection{Mathematical Formulation}
\label{sec:supp_math}

Here we will detail some of the processes and algorithms used in the OPTIMUS pipeline. 

Consider the many-to-many predictive problem where we aim to identify modality-specific biomarkers from CSF, APOE genotype, and brain cortical thickness data to predict four scores related to memory, executive function, language, and vision.

Let us begin by defining the dataset $\bm{X} = [\bm{X}^{(1)} \vert \ldots \vert   \bm{X}^{(M)}] \in \mathbb{R}^{N \times \sum p_m}$, where $M=4$ is the number of data sources or modalities $\sum_{m=1}^M p_m$ is the total number of features and $N$ the number of samples, therefore $\bm{X}^{(m)} \in \mathbb{R}^{N \times p_m} $ for $m \in \{1, 2, 3\} $ input features from three continuous-valued modalities: structural MRI, CSF protein quantification and blood transcriptomics. For $m=M=4$, the input data $\bm{X}^{(m)} \in \{0, 1, 2\}^{N \times p_4}$ consists of categorical values representing APOE genotyping. Therefore $x^{(m)}_{ij}$ is data point of the $j$-th feature of modality $m$ for observation $i$, with $i \in \{1, 2, ..., N\}$, $j \in \{1, 2, ..., p_m\} $ for each modality $m \in \{1,2,3,4\}$. Also we define $\bm{x}_i = [\bm{X}^{(1)}_{ij} \vert \ldots \vert   \bm{X}^{(M)}_{ij} ] \in  \mathbb{R}^{\sum p_m}$ as the vectorized data for observation $i$, with all modalities considered.

Due to the clinical protocol, not all modalities can be recorded for the same measurement or observation, due to time or technical constraints, like availability of certain equipment. Therefore, full modalities can be missing for one observation, and as such data points can be separated in two sets: missing indices - defined as: $ \mathcal{M} = \{(i, j,m) \, | \, x^{(m)}_{ij} \text{ is missing}\}$ and observed indices - defined as $\mathcal{O} = \{(i, j,m) \, | \, x^{(m)}_{ij} \text{is observed}\}$. The goal of imputation is to estimate data values at \((i, j,m) \in \mathcal{M}\) using the observed data \((i, j,m) \in \mathcal{O}\).

There exists a variety of imputation algorithms performing such a task, we will write the general imputation process as the output of an imputer $\bm{\hat{X}} = g_\mathcal{I}(\bm{X}, \mathcal{O})$, with $g_\mathcal{I}$ an imputer and $\bm{{X}}$ the observed data containing missing values:
\begin{equation*}
    \hat{x}_{ij}^{(m)} = 
    \begin{cases}
    x_{ij}^{(m)},  & \text{if } (i, j,m) \in \mathcal{O} 
    \\
    g_{ij}^{(m)}(\bm{X}, \mathcal{O}), & \text{if}  (i, j,m) \in \mathcal{M}.
    \end{cases}
\end{equation*}

The final selected imputer, was chosen as the one that yielded the smallest the KL divergence between the observed feature distribution and the imputed feature distribution, for a subset of features. We estimated the probability distributions of the observed values and the imputed values using histograms with a shared set of bins. Let \( \{b_k\}_{k=1}^{K+1} \) be the bin edges defining \( K \) bins. The empirical probabilities are 
$P^{(m)}_{j,k} = \frac{1}{|i:(i,j,m)\in \mathcal{O}|} \sum_{i: (i,j,m)\in \mathcal{O} }^N \bm{1}(b_k \leq x_{ij}^{(m)} < b_{k+1})$, and 
$Q^{(m)}_{j,k} = \frac{1}{|i:(i,j,m)\in \mathcal{M}|} \sum_{i: (i,j,m)\in \mathcal{M} }^N \bm{1}(b_k \leq \hat{x}_{ij}^{(m)} < b_{k+1})$.
The KL divergence is then computed as:
\begin{equation*}
D_{\text{KL}}(P \parallel Q) =\frac{1}{M} \sum_{m=1}^M \sum_{j=1}^{p_m} \sum_{k=1}^{K} P_{j,k}^{(m)} \log \frac{P_{j,k}^{(m)}}{Q_{j,k}^{(m)}}.
\end{equation*}
To avoid numerical issues, we apply a small correction term \( \epsilon>0 \), $P_{j,k}^{(m)} \gets P_{j,k}^{(m)} + \epsilon, \quad Q_{j,k}^{(m)} \gets Q_{j,k}^{(m)} + \epsilon.$

\paragraph{KNN imputation.} KNN imputation replaces missing values using the information from similar samples. Specifically, the distance between a sample \(\bm{x}_{i}\) with missing values and another sample \(\bm{x}_{t} \) is computed using the Euclidean distance:
\begin{equation*}
    d(\bm{x}_{i}, \bm{x}_{t}) = \sqrt{\sum_{j: (i,j,m), (t,j,m) \in \mathcal{O}} \left( {x}^{(m)}_{ij} - {x}^{(m)}_{tj} \right)^2},
\end{equation*}
where \(\mathcal{O}\) is the set of features observed for both \(\bm{x}_i\) and \(\bm{x}_t\).

From the set of observed samples, the \(k\)-nearest neighbors \(\bm{x}_{(1)}, \bm{x}_{(2)}, \dots, \bm{x}_{(k)}\) with the smallest distances to \(\bm{x}_{i}\) are selected. 

The missing value $x^{(m)}_{ij}, (i,j,m) \in \mathcal{M}$ is then estimated as:
\begin{equation*}
    \hat{x}^{(m)}_{ij} = \frac{1}{k} \sum_{t=1}^{k} x^{(m)}_{(t)j}.
\end{equation*}

In our final pipeline, $k = 1$, meaning that $x^{(m)}_{ij}$ is directly replaced by the value of its closest neighbor.

\paragraph{Confounder Effect Correction} Age, gender, and years of education are well-known confounders that may influence both the input features and cognitive outcomes. Failing to account for these confounders can bias feature selection and downstream pathway analyses. Here, to remove the confounding effect, we perform linear model residualization~\citep{chenResidualPartialLeast2024b}.

Given the input features with imputed missing values using the optimal imputer $\hat{\bm{X}^*} \in \mathbb{R}^{N \times \sum{p_m}}$, the target memory, executive function, visuospatial and language cognitive scores $\bm{Y} \in \mathbb{R}^{N \times 4}$, and the confounders $\bm{Z} \in \mathbb{R}^{N \times 3}$, including age, gender, and years of education, the residualization process on the input data can be defined as:
\begin{equation*}
    \epsilon_{\hat{\bm{X}}^*\vert \bm{Z}} = \bm{\hat{X}^*} - \bm{Z} \hat{\bm{\gamma}},
    \\ 
\end{equation*}

with  $\hat{\bm{\gamma}}$, the coefficients obtained via ordinary least squares (OLS) regression of $\bm{\hat{X}^*}$ on $\bm{Z}$. The residualized inputs $\epsilon_{\hat{\bm{X}}^*\vert \bm{Z}}$ are free of the linear effects of the confounders $\bm{Z}$. 

The predicted outcomes from the input residuals is:
\begin{equation*}
    \epsilon_{\hat{\bm{Y}}\vert \bm{Z}} = f^* \left ( \bm{\epsilon}_{\hat{\bm{X}}^* \mid \bm{Z}} \right),
\end{equation*}

where $f^*$ represents the predictive model trained on the residualized inputs and predicting residualized outcomes. 
During training of $f^*$ predicted $\epsilon_{\hat{\bm{Y}}\vert \bm{Z}}$ are compared against the residualized true targets  
$\bm{Y} - \bm{Z} \bm{\hat{\beta}}$. Coefficients $\bm{\hat{\gamma}}$ and $\bm{\hat{\beta}}$ are computed using the training data only to prevent data leakage. During evaluation, the predicted outcomes for the original targets $\hat{\bm{Y}}$ are then reconstructed by reintroducing the confounder effects:
\begin{equation*}
\hat{\bm{Y}} = f^* \left ( \bm{\epsilon}_{\hat{\bm{X}}^* \mid \bm{Z}}\right) + \bm{Z}\hat{\bm{\beta}}.
\end{equation*}

with $\hat{\bm{\beta}}$ the OLS estimators for $\epsilon_{\bm{Y}\vert \bm{Z}} = \bm{Y} - \bm{Z} \hat{\bm{\beta}}$.  This reconstruction ensures that predictions are comparable to the original outcomes, allowing performance metrics such as MAE or Pearson correlation to be evaluated against $\bm{Y}$. 

\paragraph{TabNet} TabNet is a transformer-based neural network architecture designed for classification and regression tasks on tabular data ~\citep{arikTabNetAttentiveInterpretable2020}. While decision tree-based architectures have generally shown superior performance on tabular inputs, the authors leveraged attention the mechanism to achieve competitive results across various tasks. This success is attributed to the attention mechanism's ability to effectively select relevant features. In this section, we will outline the some of the algorithmic details of TabNet as applied to our multi-task regression problem.

\paragraph{Attentive transformer} The attentive transformer is composed of a fully connected layer, batch normalization, and a sparsemax layer. In TabNet, attention mechanisms perform feature selection at each decision step by learning feature masks $\bm{M[i]} \in \mathbb{R}^{B \times D}$, where $\bm{i}$ denotes the decision step, $B$ is the batch size, and $D$ is the input dimension (number of features). The attentive transformer is defined as:
\begin{equation*}
    \bm{a[i] := M[i]} = \text{sparsemax}(\bm{P[i-1]})\cdot\text{h}_i (\bm{a[\bm{i-1}])}
\end{equation*}

where $\bm{P[i-1]}$ is the prior scale term, which tracks how much each feature has been used in previous steps. This prior scale is defined as $\bm{P[i] = \prod^i_{j=1} (\gamma - \bm{M[j]})}$ where $\gamma$ is a relaxation parameter. If $\gamma = 1$, a feature can only be used at one decision step. Increasing $\gamma$ allows features to contribute across multiple decision steps.

To ensure that the attention masks can be interpreted as probability distributions over features, the attention weights for each sample sum to one $\sum^D_{j=1} \bm{M[i]}_{\bm{b,j}} =1$ where $\bm{b}$ indexes the sample in the batch and $\bm{j}$ indexes the feature dimension.

\begin{figure*}[htbp]
    \begin{subfigure}[b]{0.29\textwidth}
        \centering
        \includegraphics[width=\textwidth]{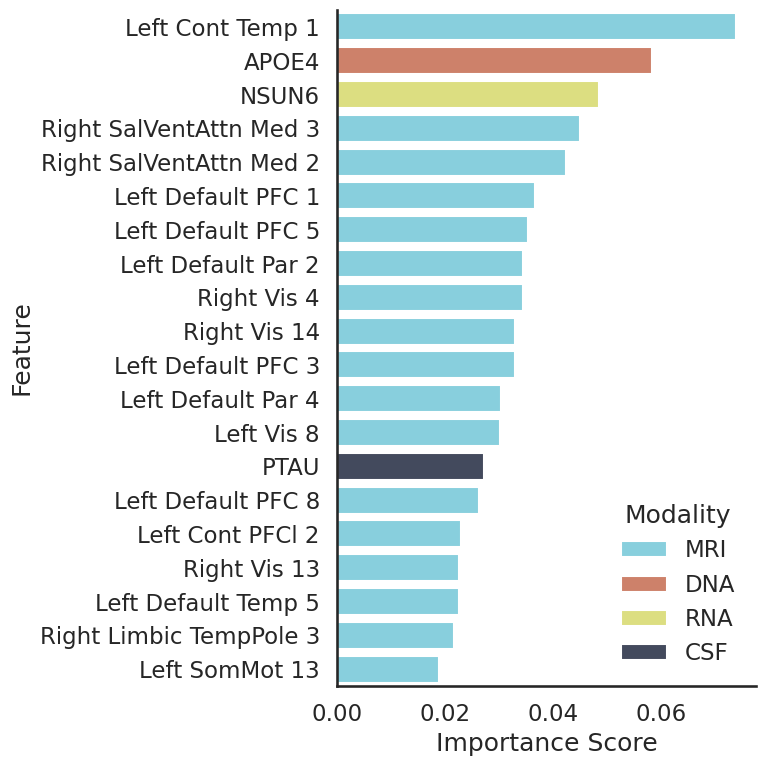}
        \caption{TabNet Local feature importance.}
        \label{fig:tabnet_adni_mem_shap_importance}
    \end{subfigure}
    \begin{subfigure}[b]{0.59\textwidth}
        \centering
        \includegraphics[width=\textwidth]{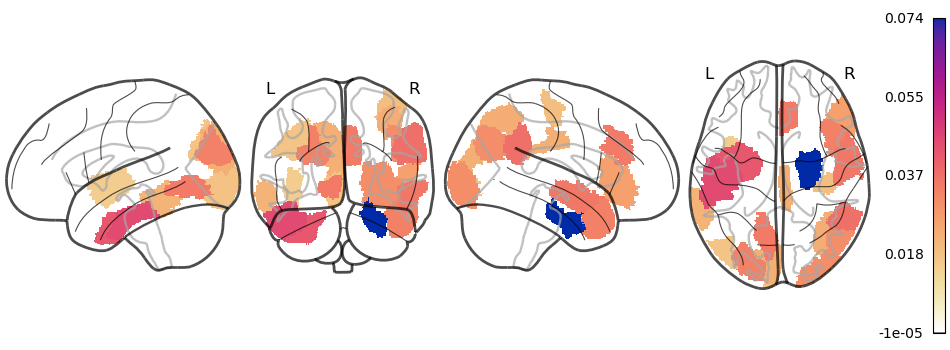}
        \caption{Local feature importance Brain Map.}
        \label{fig:tabnet_adni_mem_schaefer}
    \end{subfigure}
    \caption[TabNet local feature importance - Bar plots and Brain Map.]{\textbf{TabNet local feature importance - Bar plots and Brain Map.}
    TabNet's local feature importance maps are computed by aggreating the attention masks after training TabNet.}
    \label{fig:tabnet_local}
\end{figure*}

The attention masks from all decision steps are combined to compute the aggregated feature importance for each feature. This is computed as:
\begin{equation*}
    \bm{M}_{\text{agg-b}, j} = \frac{\sum^{N_{steps}}_{i=1} \eta_{\bm{b}}[\bm{i}] \bm{M}_{b,j}[\bm{i}]}{\sum^D_{j=1} \sum^{N_{steps}}_{i=1} \eta_{\bm{b}}[\bm{i}] \bm{M}_{b,j}[\bm{i}]^2}
\end{equation*}

where $\eta_{\bm{b}}\bm{[i]}$ is a learned scaling factor at decision step $\bm{i}$, which adjusts the contribution of each step to the final importance score. The resulting local feature importances can be seen in Fig.~\ref{fig:tabnet_local}. 

\paragraph{Permutation importance} Permutation importance measures the contribution of each feature to a fitted model's statistical performance, it involves randomly shuffling values of a single feature and observing the resulting drop in performance to determine how much the model relies on such particular feature.  In this instance we observe the effect of permuting features on the correlation between the feature and the target consequently on the model statistical performance. 

With $\bm{\hat{X}^*} \in \mathbb{R}^{N \times \sum_{m}p_m}$ the input data, $\bm{Y}$ the observed targets, $\text{r}$ is Pearson's correlation coefficient between the observed and the predicted targets $\bm{\hat{Y}(.)}$ from input $(.)$, $\pi$ is a random permutation and $\hat{\bm{X}}^*_{\pi(j)}$ the dataset where feature $j$ is shuffled across rows, while the others remain intact. The permutation importance is computed formally defined as follows : 
\begin{equation*}
\begin{split}
    I_j & = \mathbb{E}_{\pi} 
    \left \{ 
    r \left (\bm{Y}, \hat{\bm{Y}} (\hat{\bm{X}}^*) \right) - r \left(\bm{Y}, \hat{\bm{Y}} (\hat{\bm{X}}^*_{\pi(j)}) \right) 
    \right \} \\
    & \approx \frac{1}{K} \sum^{K}_{k=1} [r \left (\bm{Y}, \hat{\bm{Y}} (\hat{\bm{X}}^*) \right) - r \left(\bm{Y}, \hat{\bm{Y}} (\hat{\bm{X}}^*_{\pi(j)}) \right)])
    \end{split}
\end{equation*}

Due to the random nature of $\pi$, one feature $j$ is shuffled $K$ times and importance is averaged. 

\paragraph{Shapley values}

\begin{figure*}[htbp]
    \begin{subfigure}[b]{0.39\textwidth}
        \centering
        \includegraphics[width=\textwidth]{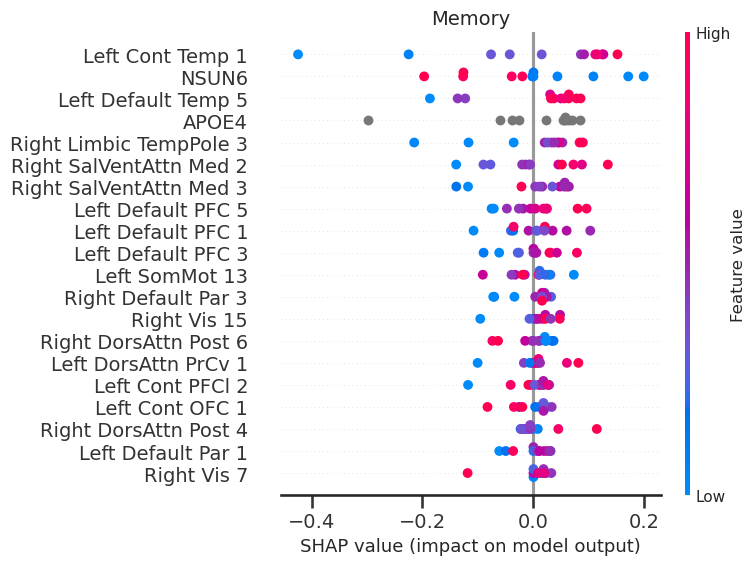}
        \caption{Memory - SHAP Beeswarm.}
        \label{fig:tabnet_adni_mem_shap_importance}
    \end{subfigure}
    \begin{subfigure}[b]{0.59\textwidth}
        \centering
        \includegraphics[width=\textwidth]{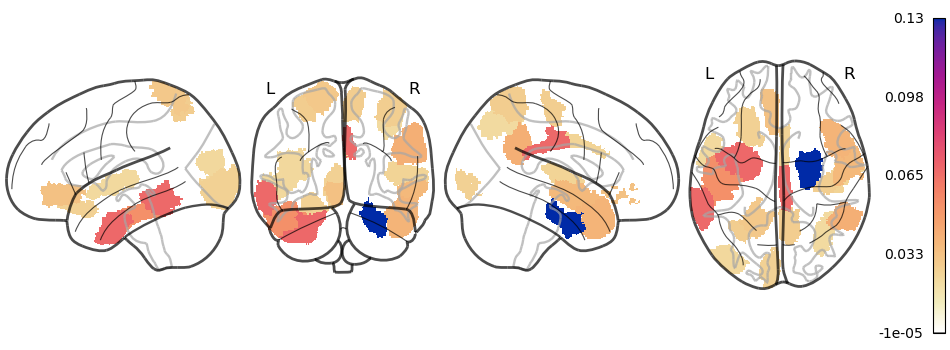}
        \caption{Memory - SHAP Brain Map.}
        \label{fig:tabnet_adni_mem_schaefer}
    \end{subfigure}
    \begin{subfigure}[b]{0.39\textwidth}
        \centering
        \includegraphics[width=\textwidth]{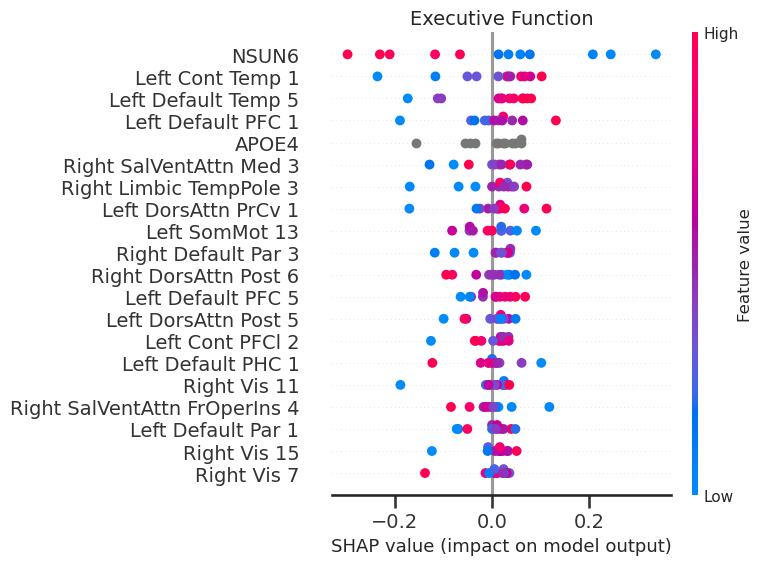}
        \caption{Executive Function - SHAP Beeswarm.}
        \label{fig:tabnet_adni_ef_shap_importance}
    \end{subfigure}
    \begin{subfigure}[b]{0.59\textwidth}
        \centering
        \includegraphics[width=\textwidth]{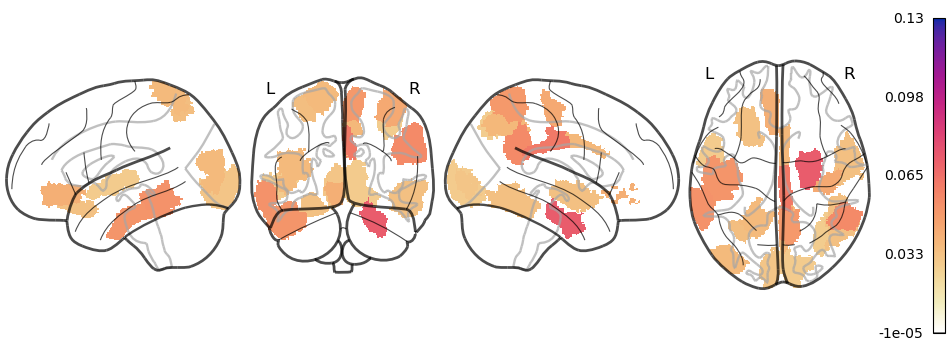}
        \caption{Executive Function - SHAP Brain Map.}
        \label{fig:tabnet_adni_ef_schaefer}
    \end{subfigure}
    \begin{subfigure}[b]{0.39\textwidth}
        \centering
        \includegraphics[width=\textwidth]{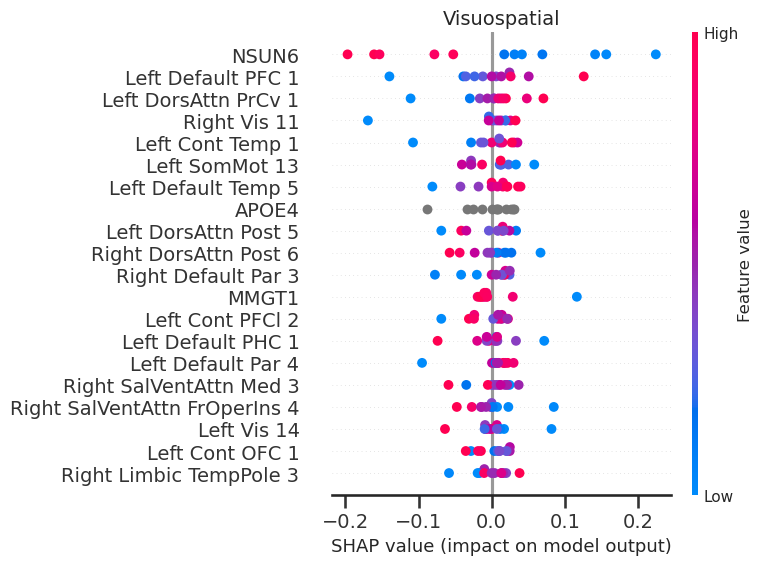}
        \caption{Visuospatial - SHAP Beeswarm.}
        \label{fig:tabnet_adni_vs_shap_importance}
    \end{subfigure}
    \begin{subfigure}[b]{0.59\textwidth}
        \centering
        \includegraphics[width=\textwidth]{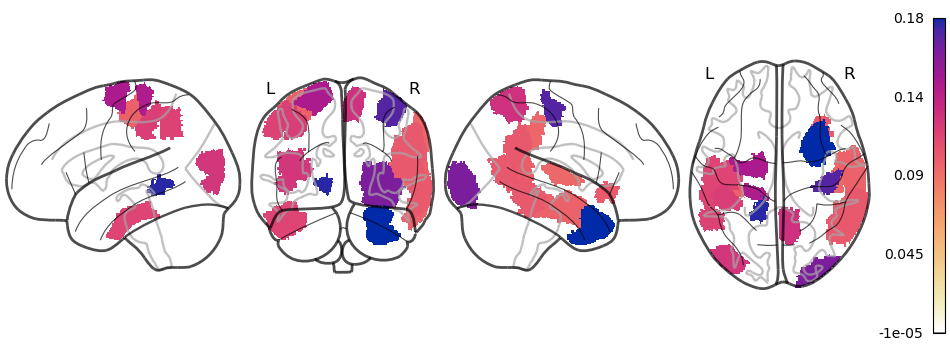}
        \caption{Visuospatial - SHAP Brain Map.}
        \label{fig:tabnet_adni_vs_schaefer}
    \end{subfigure}
    \begin{subfigure}[b]{0.39\textwidth}
        \centering
        \includegraphics[width=\textwidth]{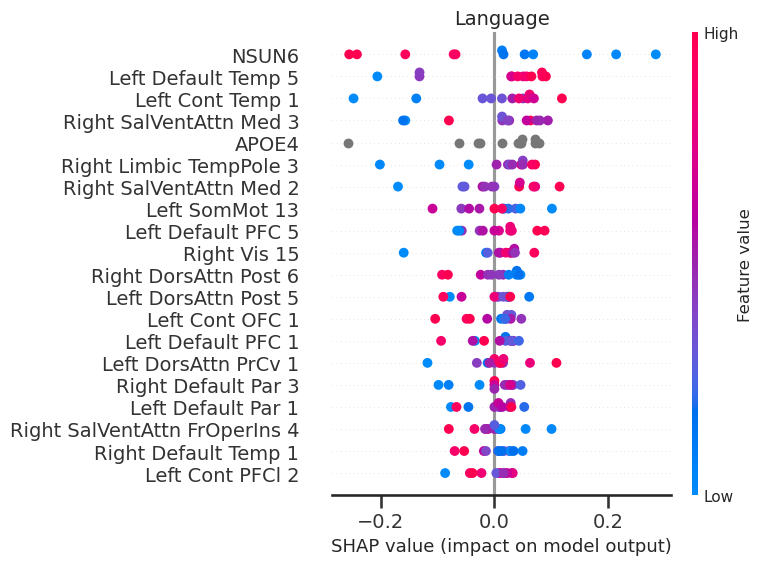}
        \caption{Language - SHAP Beeswarm.}
        \label{fig:tabnet_adni_lan_shap_importance}
    \end{subfigure}
    \begin{subfigure}[b]{0.59\textwidth}
        \centering
        \includegraphics[width=\textwidth]{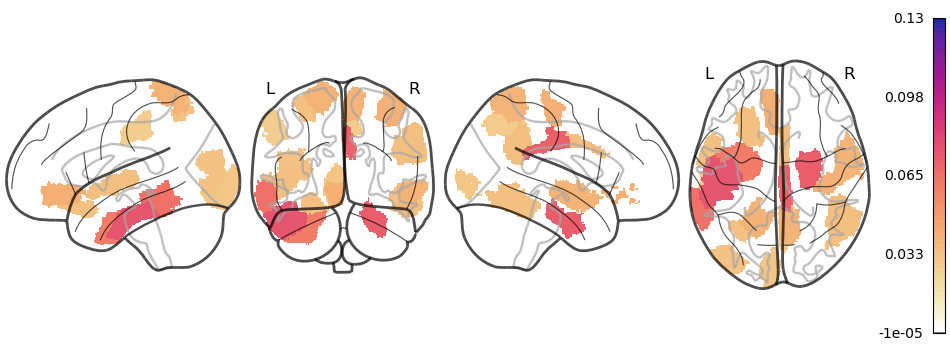}
        \caption{Language - SHAP Brain Map.}
        \label{fig:tabnet_adni_lan_schaefer}
    \end{subfigure}
    \caption[SHAP Feature Importances - Beeswarm and Brain Map.]{\textbf{SHAP Feature Importances - Beeswarm and Brain Map.}
    Each row corresponds to a cognitive domain (Memory, Executive Function, Visuospatial, and Language).
    The left column (a), (c), (e), (g) features SHAP beeswarm plots, showing the top 20 features with the highest mean SHAP values. The right column (b), (d), (f), (h) presents the corresponding brain maps, displaying SHAP mean feature importances mapped to the Schaefer Atlas (200 regions, 7 networks) in four views (top: lateral, bottom: midsagittal, left: left hemisphere, right: right hemisphere).}
    \label{fig:tabnet_shap}
\end{figure*}

Shapley values are derived from game theory and attributes to each feature a "contribution" to model's prediction. For each feature $j$, the Shapley values $\phi_j$ represents the average marginal contribution of $x_j$ to the model's prediction over all possible subsets $S \subseteq \{1, 2, ..., q\}$ of features and is computed as such : 
\begin{equation*}
    \phi_j= \sum_{S \subseteq \{1, 2, ..., q\} \backslash \{j\}} \frac{|S|!(q - |S|-1)!}{q!} [(f(S \cup \{j\}) - f(S)]],
\end{equation*}
where $S$ is a subset of features that does not include $j$, the current feature of interest. Shapley values were computed to assess feature importance and Shapley values computed for MRI features were plotted to the corresponding region of interest on Schaefer's brain \citep{schaeferLocalGlobalParcellationHuman2018a} in Fig.~\ref{fig:tabnet_shap}.  

\begin{figure*}[h]
    \centering
\includegraphics[width=0.8\textwidth]{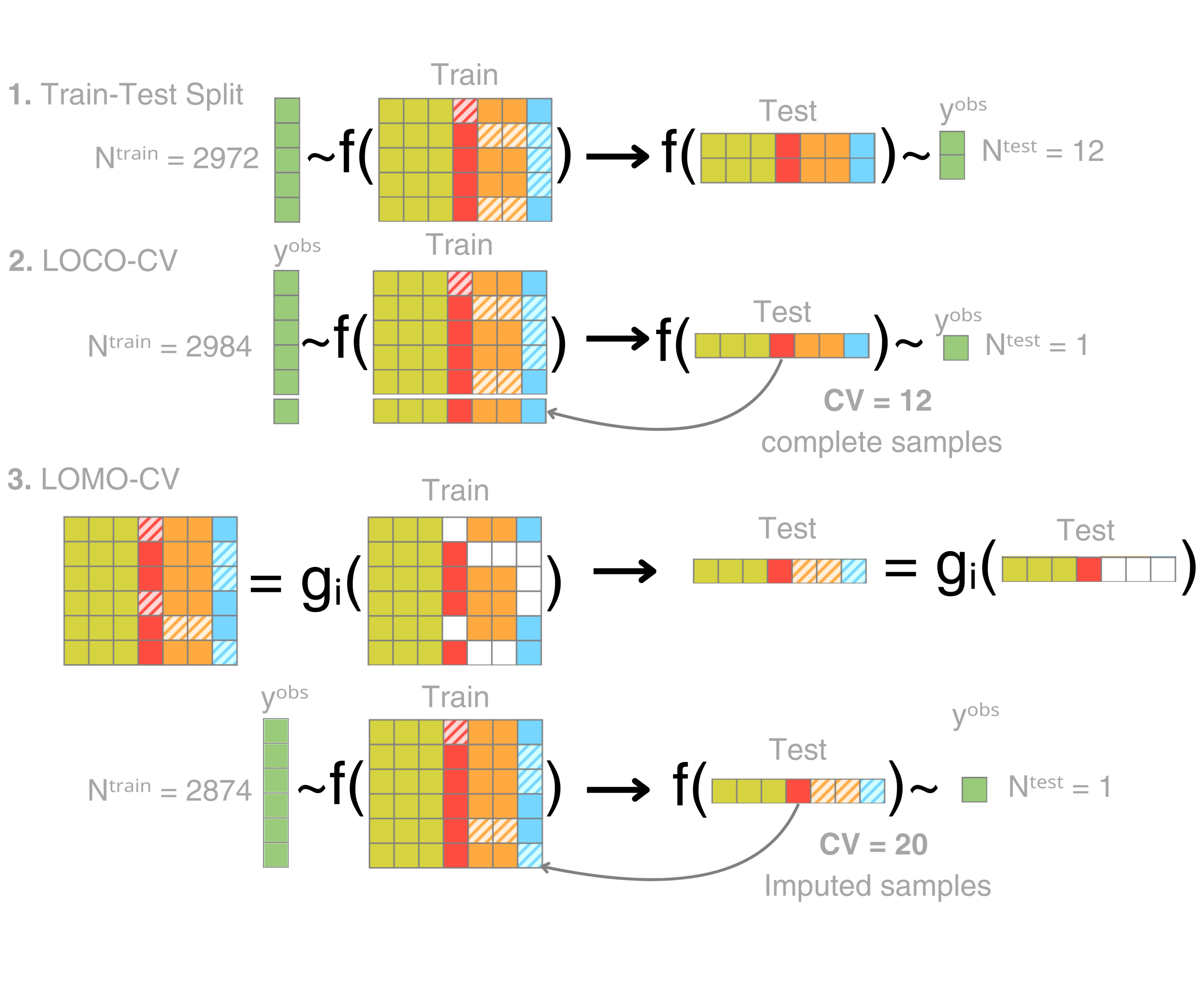}
    \caption{\textbf{Overview of the evaluation pipeline. } From top to bottom to right. (1) Train-Test Split: Performance was first assessed using a fixed train-test split, where the test set comprised 12 unique, complete measurements as described in the Section ~\ref{subsec:model_comparison_predictive}. (2) Leave-One-Complete-Out Cross-Validation (LOCO-CV): To obtain robust performance estimates, each of the 12 complete samples was sequentially treated as a test set while the remaining samples were used for training. This ensured that model performance was evaluated on entirely unseen data in each fold. (3) Leave-One-Missing-Out Cross-Validation (LOMO-CV): To simulate a realistic scenario with missing data, we randomly selected 20 participant measurements with incomplete features and applied a leave-one-missing-out cross-validation scheme. To prevent data leakage, we ensured that participants with missing features in the test set did not appear in the training set.
    }
    \label{fig:crossvalidation}
\end{figure*}

\section{Results}

\begin{figure*}[t!]
    \centering
    \begin{subfigure}[b]{0.4\textwidth}
        \centering
        \includegraphics[width=\linewidth]{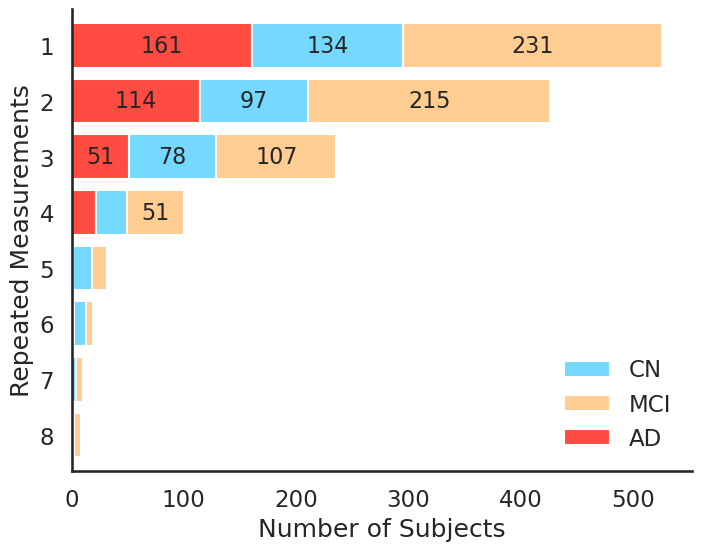}
        \caption{Participants with repeated measurements.}
        \label{fig:data_longitudinal}
    \end{subfigure}
    \hfill
    \begin{subfigure}[b]{0.36\textwidth}
        \centering
        \includegraphics[width=\linewidth]{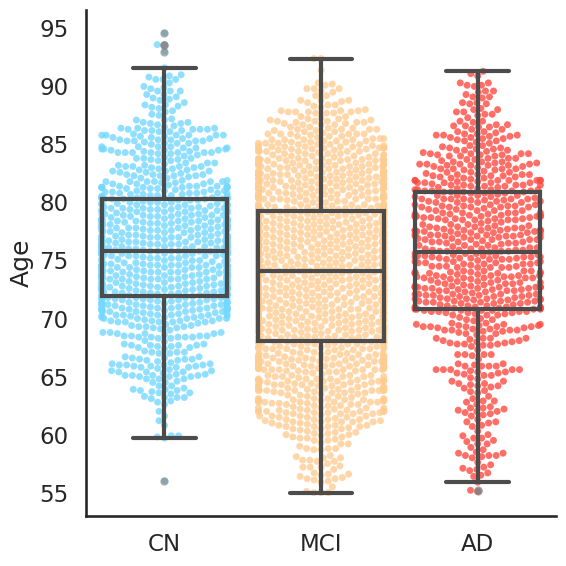}
        \caption{Age at enrollment distribution.}
        \label{fig:data_age_enrolment}
    \end{subfigure}

    \vspace{0.2cm}
    \begin{subfigure}[b]{0.4\textwidth}
        \centering
        \includegraphics[width=\linewidth]{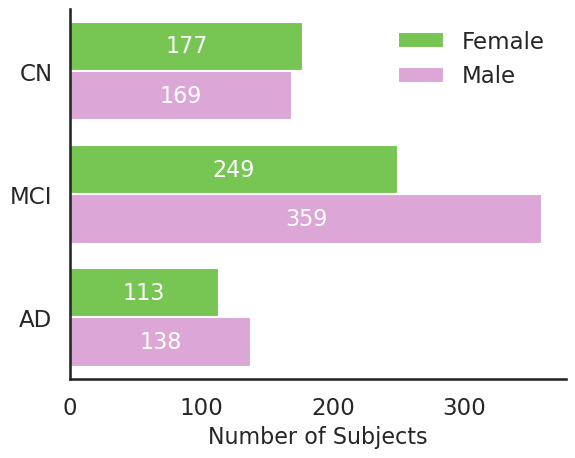}
        \caption{Sex counts.}
        \label{fig:data_gender}
    \end{subfigure}
    \hfill
    \begin{subfigure}[b]{0.4\textwidth}
        \centering
        \includegraphics[width=\linewidth]{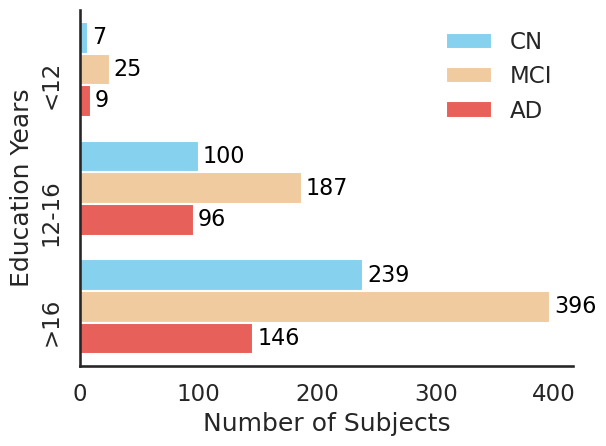}
        \caption{Education years.}
        \label{fig:data_education}
    \end{subfigure}
    
    \vspace{0.2cm}
    \begin{subfigure}[b]{\textwidth}
        \centering
        \includegraphics[width=\linewidth]{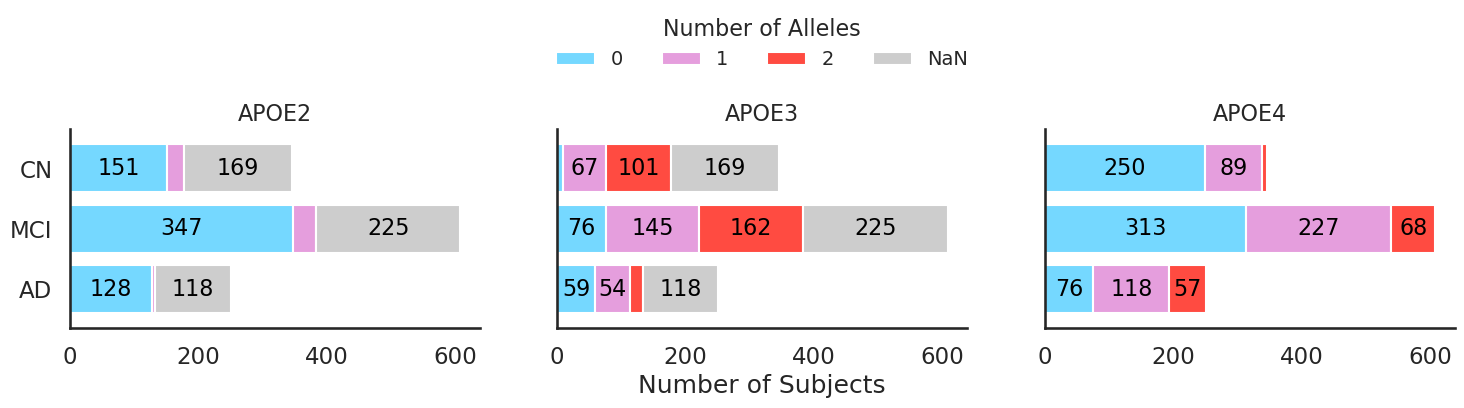}
        \caption{APOE genotypes.}
        \label{fig:data_genotype}
    \end{subfigure}
    \caption[Summary of available data and modalities.]{\textbf{Summary of available data and demographic characteristics of subjects.} The summary of the values in Table \ref{tab:demographics} and are presented as barplots or boxplots per cognitive group (blue : CN, yellow : MCI, and red : AD). 
    (a) Barplot illustrating the distribution of repeated measurements, highlighting the longitudinal design.
    (b) Age at enrollment distribution by cognitive group, shown with boxplots (median and interquartile range) and overlaid swarmplots with individual data points .
    (c) Barplot depicting the number of females (green) and males (pink) within each cognitive group.
    (d) Barplot of education level categorized as \textless12 years, 12–16 years, and \textgreater16 years across cognitive groups.
    (e) Stacked barplot of APOE alleles counts ($\epsilon$2, $\epsilon$3, $\epsilon$4) across cognitive groups, with allele counts indicated by color (blue: 0, purple: 1, red: 2, grey: missing). The APOE genotype is associated with cognitive decline and AD progression, with the APOE4 isoform being the one that predisposes to AD, whereas APOE2 is suspected to have a protective effect against AD and APOE3 being the neutral variant. }
    \label{fig:exploratory_analysis}
\end{figure*}

\begin{figure*}[t!]
    \centering
    \begin{subfigure}[b]{\textwidth}
        \centerline{\includegraphics[width=\linewidth]{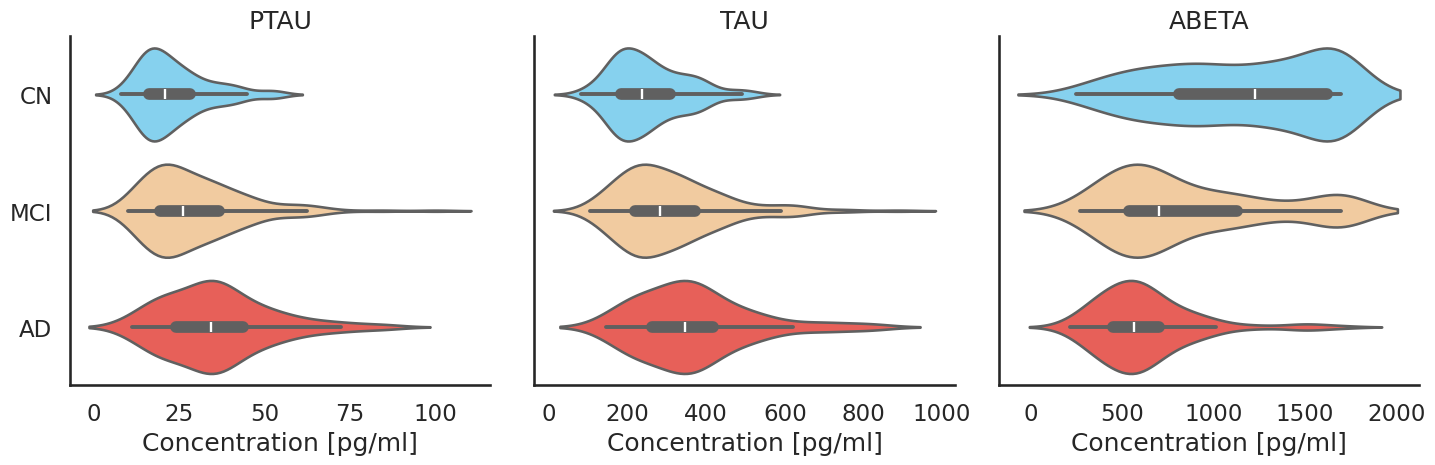}}
        \caption{Violin plot of CSF biomarker concentrations.}
        \label{fig:data_csf_violin}
    \end{subfigure}
    
    \vspace{0.5cm}
    \begin{subfigure}[b]{\textwidth}
        \centerline{\includegraphics[width=\linewidth]{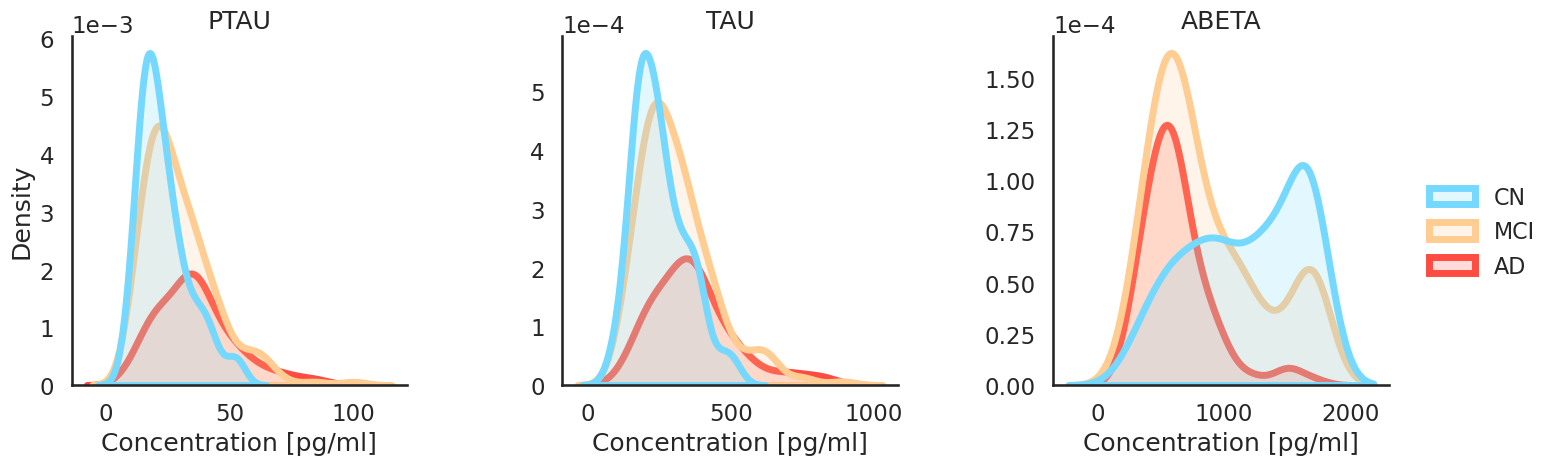}}
        \caption{KDE plot of CSF biomarker distributions.}
        \label{fig:data_csf_distplot}
    \end{subfigure}

    \vspace{0.5cm}

    \caption[Distribution of cerebrospinal fluid (CSF) biomarkers.]{\textbf{Distribution of cerebrospinal fluid (CSF) biomarkers.}
    
    Distribution plots show phosphorylated tau (PTAU), total tau (TAU), and amyloid beta (ABETA) concentrations in pg/ml across cognitive groups (blue : CN, yellow : MCI, and red : AD).
    (a) Violin plots show biomarker concentrations with a symmetric Kernel Density Estimation (KDE) colored by group around the quartiles / whiskers of a box plot in black with mean value in white. 
    (b) Kernel Density Estimation (KDE) plots to better visualize the distribution of CSF proteins concentrations, curves are colored per group.}
    \label{fig:data_csf_combined}
\end{figure*}

\begin{figure*}[htbp]
    \centering
    \includegraphics[width=0.8\linewidth]{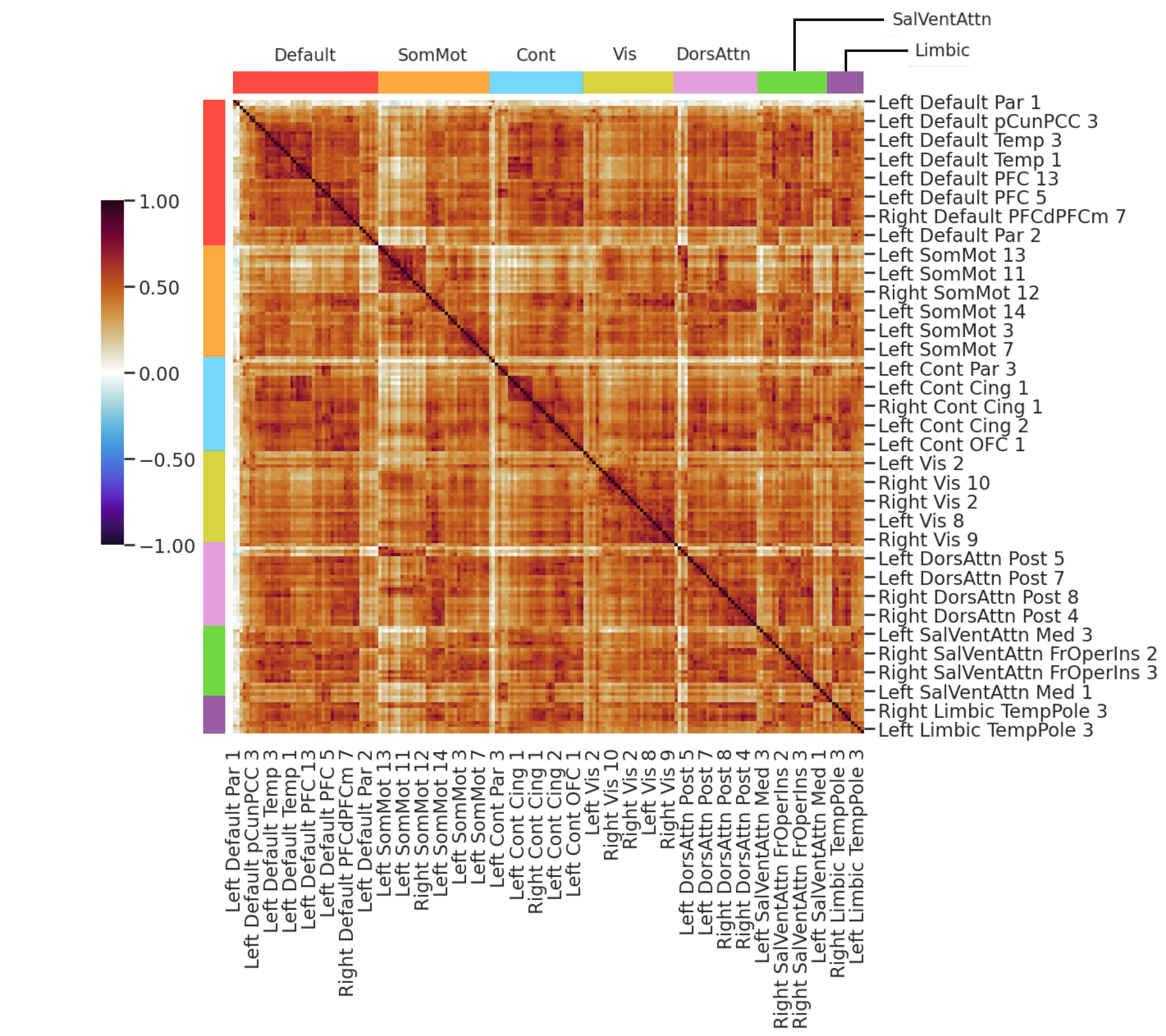}
    \caption[ Pearson correlation heatmap of cortical features with hierarchical clustering within brain networks. ]{\textbf{Pearson correlation heatmap of cortical features with hierarchical clustering within brain networks.} This heatmap shows the reordered correlation matrix for 200 cortical regions of interest (ROIs), grouped into seven brain networks based on the Schaefer atlas - red : Default, orange : Somato-motor (SomMot), blue : Control (Cont), yellow : Visual (Vis), pink : Dorsal Attention (DorsAttn), green : Salience and Ventral Attention (SalVentAttn), purple : Limbic. 
    Cortical thickness measures were derived from structural MRI data for each ROI. Pairwise Pearson correlations were computed across the entire dataset (N = 2894) of cortical thickness measures (K = 200) for each ROI.
    Features within each network were hierarchically clustered using the average linkage method, with distances derived from pairwise correlations. The rows and columns represent individual cortical regions, reordered by hierarchical clustering within their respective networks. The color map encodes Pearson correlation coefficients, with blue indicating negative correlation (-1) and red indicating positive correlations (1).
    Highly correlated cortical features can distort the interpretation of individual feature importance in perturbation-based feature analysis (permutation importance or Shapley values). }
    \label{fig:mri_correlation}
\end{figure*}

\begin{figure*}[t!]
    \centering
\centerline{\includegraphics[width=0.6\linewidth]{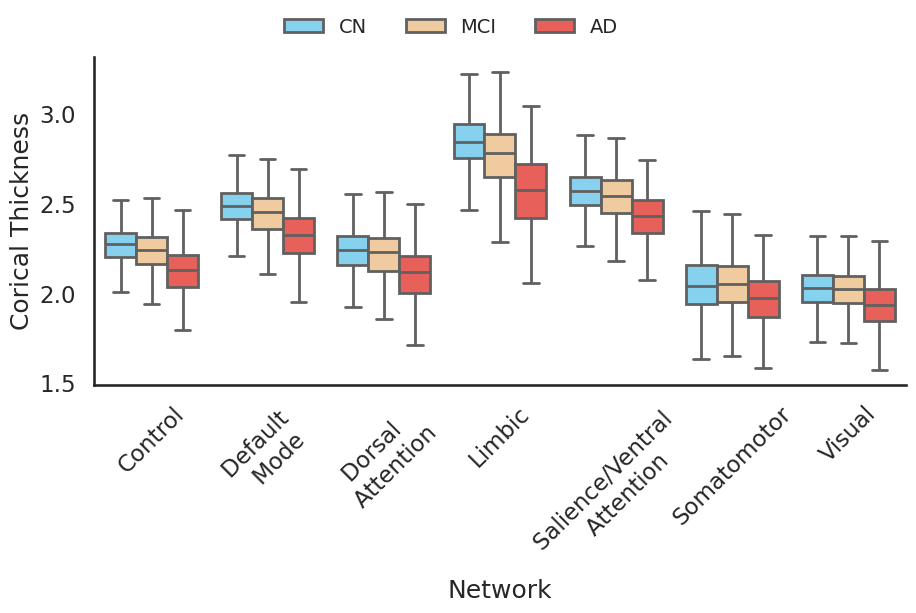}} 
    \caption[Boxplots of Cortical Thickness per Network.]{\textbf{Boxplots of Cortical Thickness per Network.}  These boxplots display average cortical thickness per functional network as defined in Schaefer atlas with 200 regions of interest (ROI) and 7 networks for the different cognitive subject groups (blue : CN, yellow : MCI, and red : AD). }
    \label{fig:data_combined_network_mri}
\end{figure*}

\begin{figure*}[t!]
    \centering
    \begin{subfigure}[b]{0.45\textwidth}
        \centering
        \includegraphics[width=\linewidth]{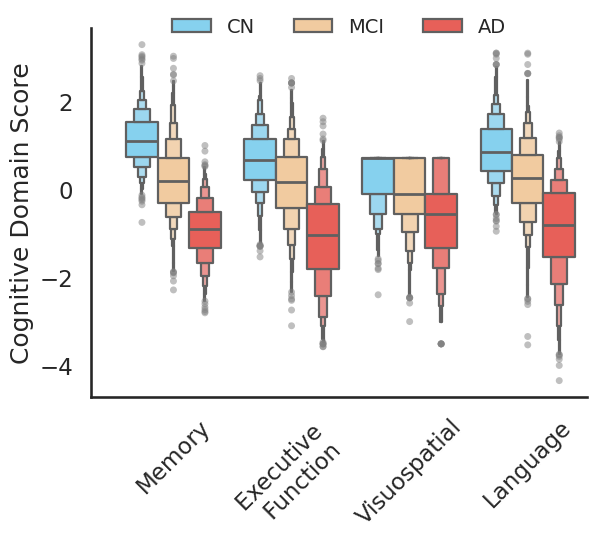}
        \caption{Boxen plot of target domain scores.}
        \label{fig:data_targets_boxenplot}
    \end{subfigure}
    \hfill
    \begin{subfigure}[b]{0.48\textwidth}
        \centering
        \includegraphics[width=\linewidth]{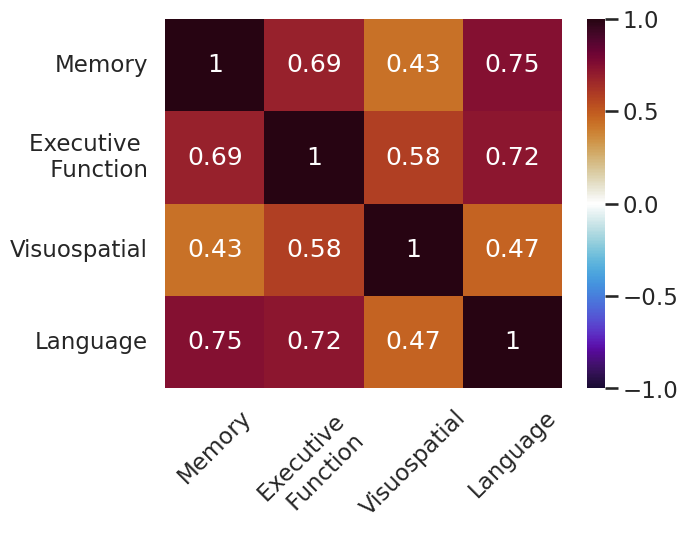}
        \caption{Heatmap of Pearson correlation coefficients.}
        \label{fig:data_targets_heatmap}
    \end{subfigure}
    \vspace{1cm} 
    \caption[Distribution of Cognitive Domain Scores.]{\textbf{Distribution of Cognitive Domain Scores.} 
    (a) Boxen plot shows target cognitive domain scores, Memory, Executive Function, Visuospatial, Language in x-axis colored per cognitive groups (blue : CN, yellow : MCI, and red : AD). In every domain, the scores are lower for MCI and worsen in AD subjects. 
    (b) Heatmap displays Pearson correlation coefficients between cognitive domain scores, highlighting that they are highly correlated. Values}
    \label{fig:exploratory_targets}
\end{figure*}

\begin{table*}[!ht]
  \fontsize{8pt}{10pt}\selectfont
  \centering
    \caption[Enriched KEGG pathways for genes of interest in the functional interaction network.]{\textbf{Enriched KEGG pathways for genes of interest in the functional interaction network.} This table lists the significantly enriched KEGG pathways derived from STRING-db functional enrichment analysis. 
    Columns include the pathway name in Description with corresponding color in Figure \ref{fig:results_dge}, strength, false discovery rate (FDR), and the corresponding number genes of interest annotated with the pathway. The strength metric (log10(observed/expected)) describes the enrichment effect size, meaning for one annotation, the number of genes observed in the network that have this annotation divided by the number of proteins expected to be annotated with this term in a random network of the same size. The signal is the weighted harmonic mean between the observed/expected ratio and -log(FDR), aims to offer a balance between strength and FDR metrics for enriched terms ordering. FDR values represent significance after correction for multiple testing using the Benjamini–Hochberg procedure. Pathway colors correspond to those shown in the network figure for easy cross-reference. Rows were ordered by ascending FDR values.}
  \begin{tabular}{|p{3cm}||p{2cm}|p{1cm}p{1cm}p{1cm}p{1cm}p{3cm}p{2cm}|}
    \hline
        Description & Chart Color & Strength & Signal & FDR & P-Value & \# Background Genes & \# Genes \\
        \hline
        Thermogenesis & \cellcolor[HTML]{FBFF32}~ & 0.99 & 0.53 & 0.0133 & 3.97E-5 & 226 & 6 \\ 
        Viral Carcinogenesis & \cellcolor[HTML]{974EA3}~ & 1.0 & 0.46 & 0.027 & 1.6E-4 & 183 & 5 \\ 
        Alzheimer Disease & \cellcolor[HTML]{FF7F00}~ & 0.89 & 0.38 & 0.0486 & 4.3E-4 & 354 & 6 \\ 
        Parkinson Disease & \cellcolor[HTML]{4DAF49}~ & 0.79 & 0.37 & 0.0486 & 5.0E-4 & 236 & 5 \\ \hline
    \end{tabular}
    \label{tab:results_dge}
\end{table*}

\begin{figure*}
\centering
\includegraphics[width=0.8\textwidth, trim=0 3cm 0 0, clip]{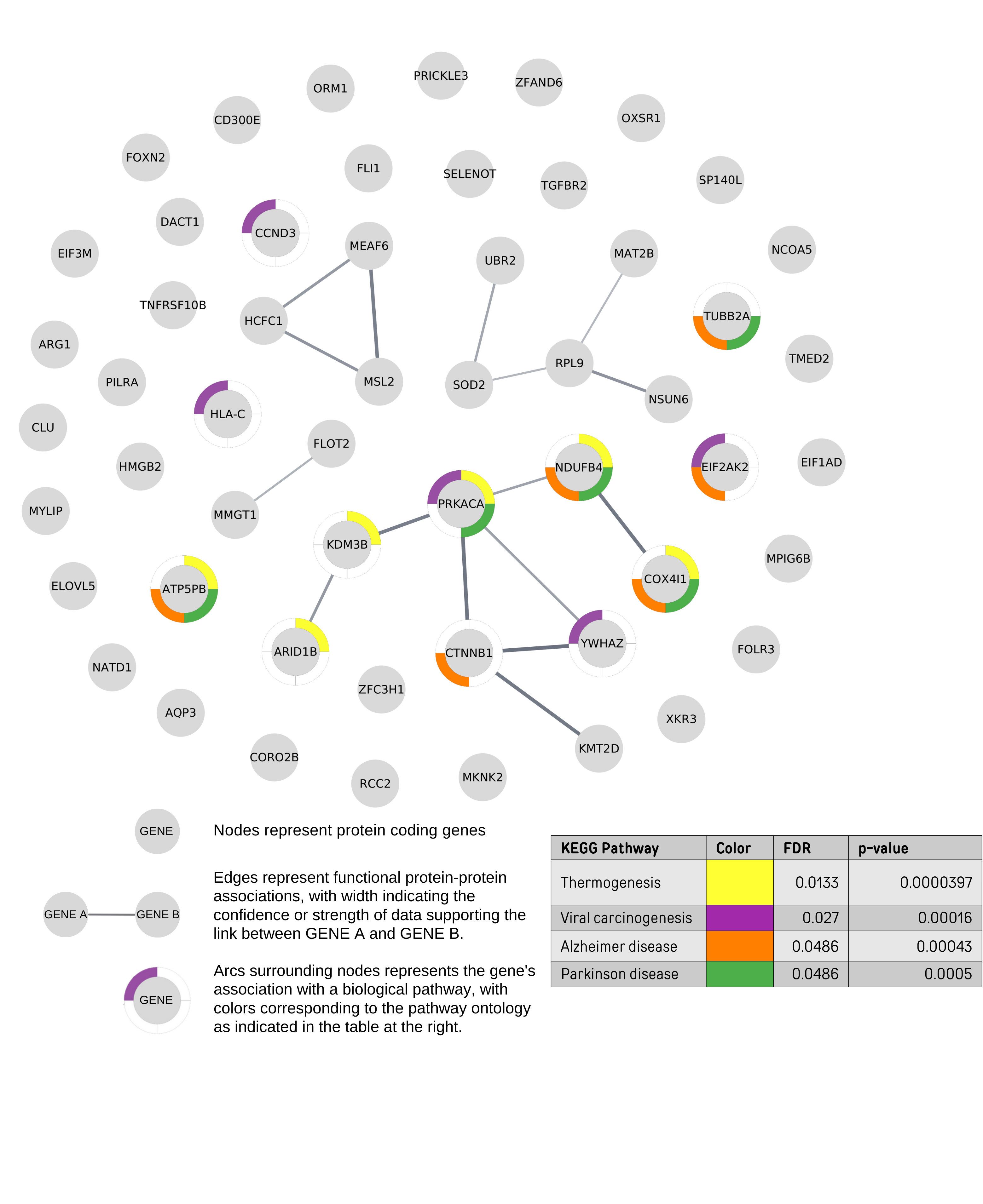}
\caption[Functional interaction network of selected blood-expressed genes]{\textbf{Functional interaction network of selected blood-expressed genes.}
This network displays functional interactions (not physical interactions) among the selected genes of interest, as identified by STRING-db. Transcripts and pseudogenes were excluded due to database constraints, resulting in some genes (e.g.,  ATP8B5P) not being included. The network was constructed using evidence from curated databases and experimental data, with a confidence threshold of 0.4. The width of the edges reflects the confidence value of the interactions. Nodes represent individual genes, and colors around the nodes indicate enriched KEGG pathways. Enrichment results are based on the network and computed using STRING-db's methods, with details provided in the supplementary Table \ref{tab:results_dge}. The enriched KEGG pathways and their associated colors are -  yellow : Thermogenesis, purple : Viral Carcinogenesis , orange : Alzheimer Disease and green : Parkinson disease}
\label{fig:results_dge}
\end{figure*}

\begin{figure*}[t!]
    \centering
    \begin{subfigure}[b]{.65\textwidth}
        \centering
        \includegraphics[width=\linewidth]{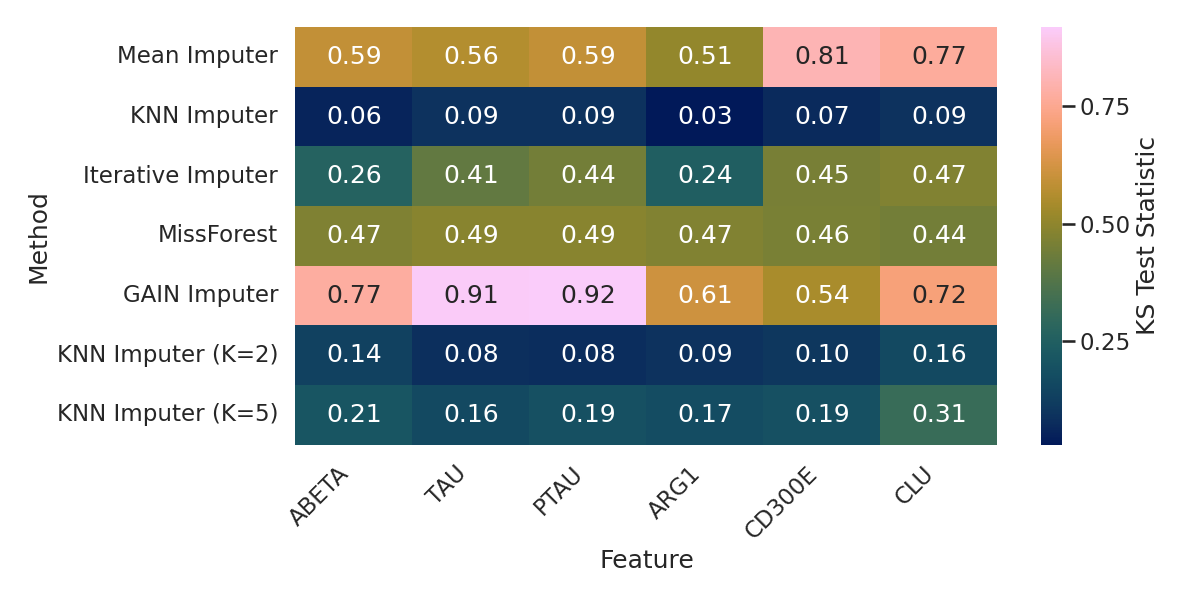}
        \caption{KS statistics heatmap.}
        \label{fig:imputation_ks_statistic}
    \end{subfigure}
    \begin{subfigure}[b]{.65\textwidth}
        \centering
        \includegraphics[width=\linewidth]{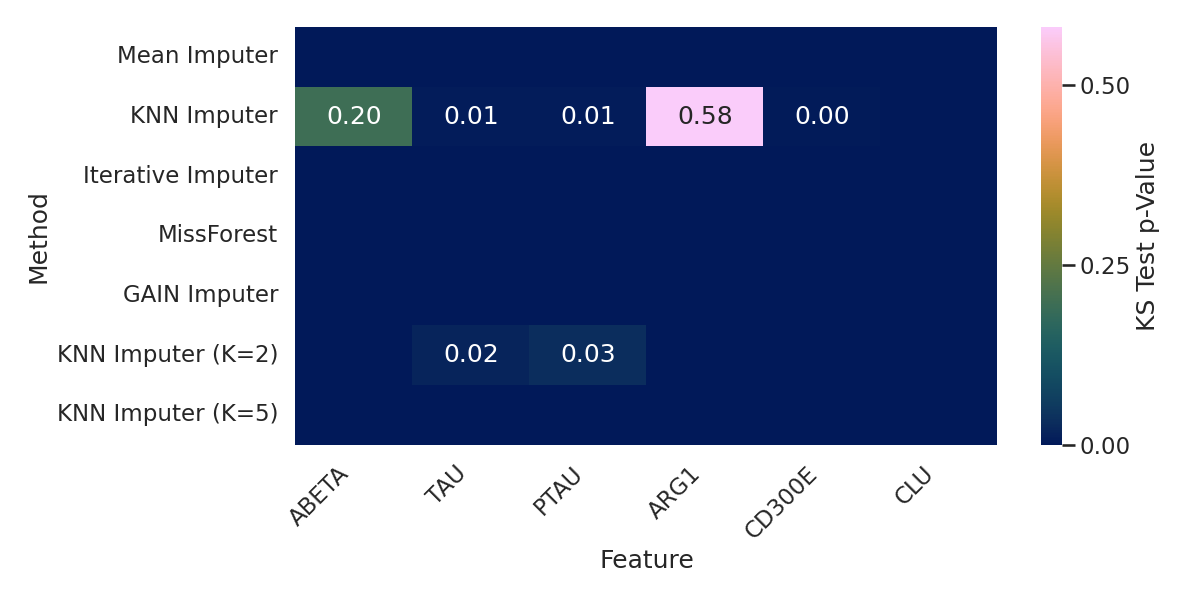}
        \caption{KS p-values heatmap.}
        \label{fig:imputation_ks_pvalues}
    \end{subfigure}
    \begin{subfigure}[b]{.65\textwidth}
        \centering
        \includegraphics[width=\linewidth]{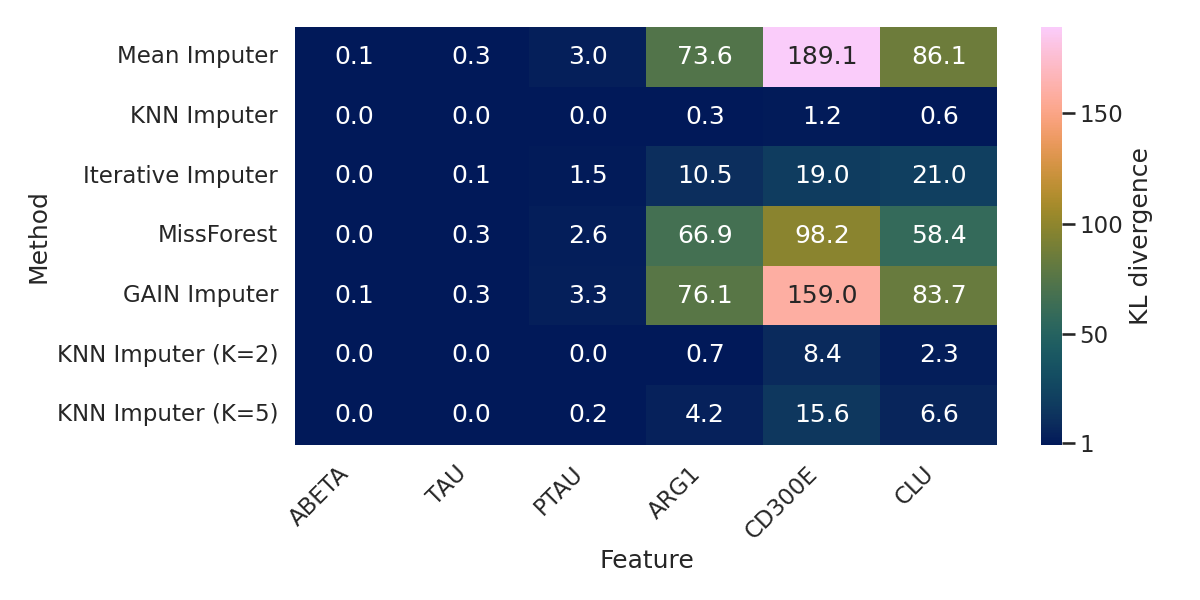}
        \caption{KL divergence heatmap.}
        \label{fig:imputation_kl_divergence}
    \end{subfigure}
    \caption[Heatmaps of KS statistics, KS p-values, and KL divergence for imputed missing values]{ \textbf{Heatmaps of KS statistics, KS p-values, and KL divergence for imputed missing values. }
        The heatmaps compares the imputed missing values to the original data distributions across various imputation methods.
        (a) Kolmogorov–Smirnov test statistics measure the maximum difference between cumulative distributions of original and imputed data.
        (b) Kolmogorov–Smirnov test p-values evaluate the statistical significance of the observed differences.
        (c) Kullback–Leibler divergence quantifies information loss between original and imputed distributions.}
    \label{fig:imputation_heatmaps}
\end{figure*}

\begin{table*}[ht]
  \fontsize{8pt}{10pt}\selectfont
  \centering
  \caption[Summary and parameters of some imputation algorithms.]{\textbf{Summary and parameters of some imputation algorithms.} This table summarizes the imputation algorithms tested, categorized into continuous and ordinal value imputation methods with the continuous values imputation used to impute CSF and RNA-seq features and the ordinal for APOE genotype imputation. Each algorithm's description and key parameters are provided. For methods where parameters are unspecified, default values were applied. }
\begin{tabular}{|p{4cm}||p{9cm}|p{3cm}|}
 \hline
 \textbf{Algorithm} & \textbf{Description} & \textbf{Parameters} \\
 \hline
  \rowcolor{lightgray}
  Continuous values imputation & & \\
 Mean & Replace missing values by the average of features. & None. \\
 K-Nearest Neighbors & K-Nearest Neighbors with an euclidean distance metric to find the nearest neighbors and each missing value is imputed using the values from the nearest neighbors that have a value for the feature. Then the feature of neighbors are averaged uniformly. &  Number of neighbors = 1, 2, 5 \\
 MICE & Iterative algorithm that will fill in missing values by fitting a linear model on the available data one feature at a time .&  Default parameters. \\
 Missforest & Same as MICE but instead of a linear regression, it uses a random forest model to predict missing values. &  Default parameters.\\
GAIN  & Algorithm for denoising high-dimensional data most commonly applied to single-cell RNA sequencing data. MAGIC learns the manifold data, using the resultant graph to smooth the features and restore the structure of the data. &  Default parameters.\\
  \rowcolor{lightgray}
  Ordinal values imputation & & \\
Most Frequent & Replace by the most frequent value & None. \\
K-Nearest Neighbors & K-Nearest Neighbors with an euclidean distance metric. &  Number of neighbors = 1. \\
Missing Indicator & Replace by a constant value to indicate that it is missing & Value= -1. \\ \hline
\end{tabular}
 \label{tab:imputation}
\end{table*}

\begin{table*}[!ht]
  \fontsize{8pt}{10pt}\selectfont
  \centering
  \caption[Pearson correlation scores for imputation algorithm comparison]{\textbf{Pearson correlation scores for imputation algorithm comparison. } Missing value imputation performance on cognitive scores (Memory,Executive Function, Language Visuospatial) prediction task using a linear regression as predictor, values indicated are Pearson's correlation coefficients between true and predicted cognitive scores. The number of samples in the testing set was 12. Mean and std denote the average and standard deviation of the row-wise correlation scores (across target cognitive domain scoring) in order to get an overview of the best performing imputation combination. True target residuals were adjusted to remove the effects of confounders before comparison with the predicted target values. Values were sorted in descending mean correlation score. }
    \begin{tabular}{|p{3.66cm}p{3.4cm}||p{1.3cm}p{1.3cm}p{1.3cm}p{1.3cm}||p{0.6cm}p{0.6cm}|}
     \hline
        \textbf{ Ordinal  Imputer  } & \textbf & \textbf{Executive Function} & \textbf{ Language} & \textbf{Memory} & \textbf{Visuo-spatial} & \textbf{Mean} & \textbf{Std} \\ \hline
        Most Frequent & KNNImputer & 0.831 & 0.228 & 0.400 & 0.211 & 0.418 & 0.289 \\ 
        KNNImputer & KNNImputer & 0.831 & 0.222 & 0.383 & 0.219 & 0.414 & 0.289 \\ 
       Missing Indicator & KNNImputer & 0.831 & 0.212 & 0.383 & 0.219 & 0.411 & 0.291 \\ 
        KNNImputer & SimpleImputer  Mean & 0.844 & 0.180 & 0.304 & 0.195 & 0.381 & 0.314 \\ 
        Most Frequent & SimpleImputer  Mean & 0.843 & 0.181 & 0.317 & 0.180 & 0.380 & 0.315 \\ 
       Missing Indicator & SimpleImputer  Mean & 0.843 & 0.170 & 0.307 & 0.198 & 0.379 & 0.315 \\ 
       Missing Indicator & GAINImputer & 0.794 & 0.190 & 0.314 & 0.220 & 0.379 & 0.281 \\ 
        Most Frequent & IterativeImputer & 0.765 & 0.252 & 0.212 & 0.249 & 0.370 & 0.264 \\ 
       Missing Indicator & IterativeImputer & 0.765 & 0.244 & 0.209 & 0.257 & 0.369 & 0.265 \\ 
        KNNImputer & IterativeImputer & 0.764 & 0.249 & 0.201 & 0.256 & 0.368 & 0.266 \\ 
        Most Frequent & GAINImputer & 0.783 & 0.151 & 0.292 & 0.180 & 0.351 & 0.294 \\ 
        KNNImputer & GAINImputer & 0.781 & 0.152 & 0.270 & 0.192 & 0.349 & 0.293 \\ 
        KNNImputer & MissForest & 0.838 & 0.141 & 0.257 & 0.106 & 0.335 & 0.341 \\ 
        Most Frequent & MissForest & 0.837 & 0.142 & 0.272 & 0.089 & 0.335 & 0.344 \\ 
       Missing Indicator & MissForest & 0.837 & 0.132 & 0.261 & 0.107 & 0.334 & 0.342 \\ \hline
    \end{tabular}
    \label{tab:imputation_corr}
\end{table*}

\begin{table*}[!ht] \fontsize{8pt}{10pt}\selectfont
\centering
\caption[Mean absolute error scores for imputation algorithm comparison]{\textbf{Mean absolute error scores for imputation algorithm comparison.}
This table compares the performance of missing value imputation algorithms on predicting cognitive scores (Memory, Executive Function, Language, Visuospatial) using linear regression as the predictor. The reported metric is the mean absolute error (MAE) between true and predicted cognitive scores. The testing set contained 12 samples. Mean and Std columns represent the average and standard deviation of row-wise MAE scores across the target cognitive domains, providing an overview of the best-performing imputation method. True target residuals were adjusted for confounders before comparison with predicted values. Results are sorted by ascending MAE scores.}
    \begin{tabular}{|p{3.66cm}p{3.3cm}||p{1.3cm}p{1.3cm}p{1.3cm}p{1.3cm}||p{0.6cm}p{0.6cm}|}
     \hline
        \textbf{ Ordinal  Imputer  } & \textbf{ Continuous  Imputer  } & \textbf{Executive Function} & \textbf{ Language} & \textbf{Memory} & \textbf{Visuo-spatial} & \textbf{Mean} & \textbf{Std} \\ \hline
        SimpleImputer  Most Frequent & KNNImputer & 0.527 & 0.799 & 0.767 & 0.650 & 0.686 & 0.124 \\ 
        KNNImputer & KNNImputer & 0.527 & 0.802 & 0.772 & 0.649 & 0.687 & 0.126 \\ 
       Missing Indicator & KNNImputer & 0.528 & 0.808 & 0.776 & 0.648 & 0.690 & 0.128 \\ 
        Most Frequent & SimpleImputer  Mean & 0.462 & 0.884 & 0.836 & 0.670 & 0.713 & 0.191 \\ 
        KNNImputer & SimpleImputer  Mean & 0.462 & 0.885 & 0.841 & 0.667 & 0.714 & 0.192 \\ 
       Missing Indicator & SimpleImputer  Mean & 0.463 & 0.891 & 0.844 & 0.665 & 0.716 & 0.195 \\ 
        Most Frequent & MissForest & 0.445 & 0.880 & 0.901 & 0.678 & 0.726 & 0.213 \\ 
        KNNImputer & MissForest & 0.444 & 0.880 & 0.906 & 0.676 & 0.727 & 0.215 \\ 
       Missing Indicator & GAINImputer & 0.505 & 0.883 & 0.858 & 0.660 & 0.727 & 0.178 \\ 
       Missing Indicator & MissForest & 0.446 & 0.886 & 0.908 & 0.674 & 0.729 & 0.216 \\ 
        Most Frequent & IterativeImputer & 0.550 & 0.805 & 0.927 & 0.679 & 0.740 & 0.163 \\ 
        KNNImputer & IterativeImputer & 0.550 & 0.806 & 0.928 & 0.678 & 0.740 & 0.163 \\ 
       Missing Indicator & IterativeImputer & 0.549 & 0.812 & 0.923 & 0.678 & 0.741 & 0.162 \\ 
        Most Frequent & GAINImputer & 0.523 & 0.906 & 0.870 & 0.677 & 0.744 & 0.178 \\ 
        KNNImputer & GAINImputer & 0.526 & 0.903 & 0.881 & 0.671 & 0.745 & 0.180 \\ \hline
    \end{tabular}
    \label{tab:imputation_mae}
\end{table*}

\begin{figure*}[t!]
    \centerline{\includegraphics[width=0.8\linewidth]{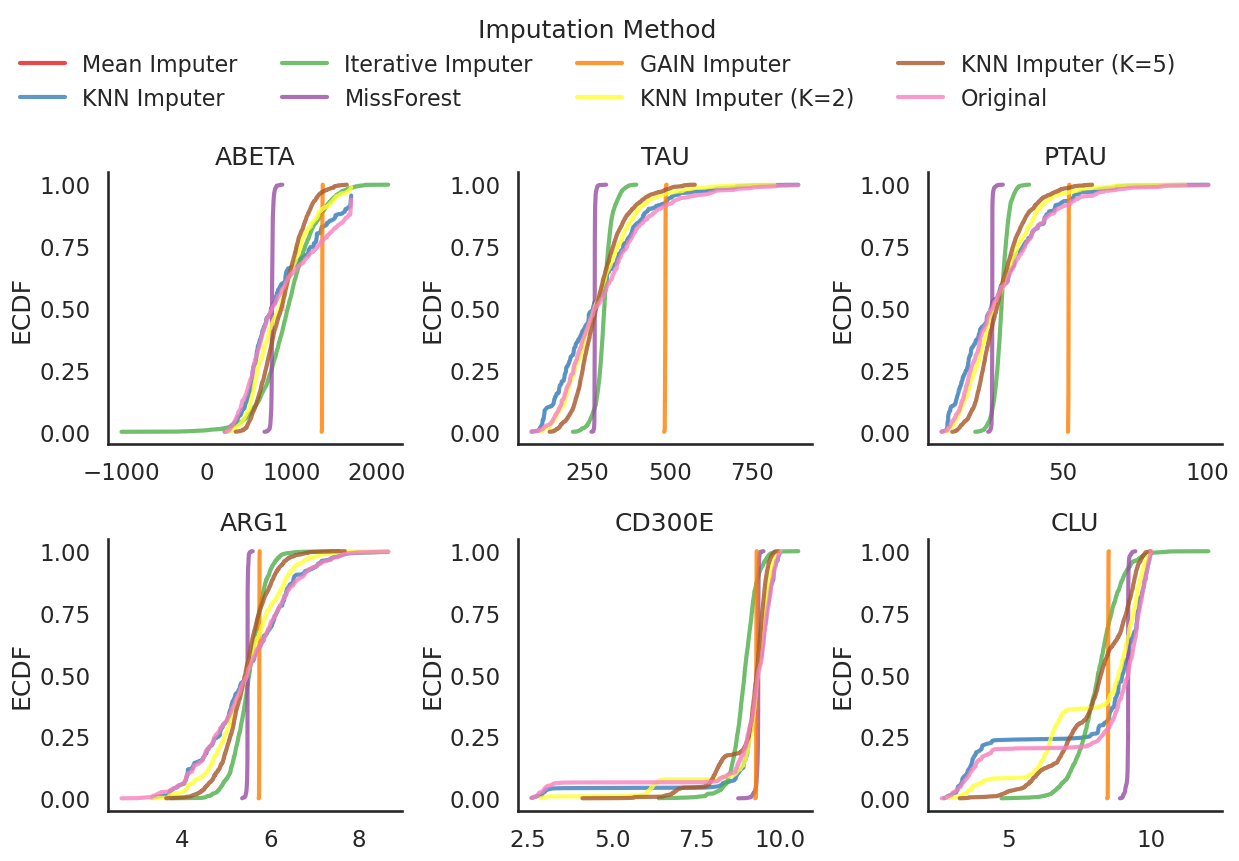}}
\caption[Empirical Cumulative Distribution Function (ECDF) Plots of Missing Values Imputation]{\textbf{Empirical Cumulative Distribution Function (ECDF) Plots of Missing Values Imputation.} Line plots representing the empirical cumulative distribution function (ECDF) for the features: ABETA (A\textbeta), TAU, PTAU (phosphorylated tau) from cerebrospinal fluid (CSF), and AGR1, CD300E, CLU, ORM1, rCC2, TGFBR2, and TUBB2A from blood RNA-seq expression data.
Each subplot corresponds to one feature, showing the ECDF of the original dataset and each imputation method used. The x-axis represents the feature values, while the y-axis represents the cumulative proportion of data points.
Imputation methods detailed in Table \ref{tab:imputation} are indicated by different colors, with the original data shown in red as a reference. This visualization highlights the distribution similarities between the original and imputed values.}
\label{fig:imputation_edf}
\end{figure*}

\begin{table*}[!ht]
\fontsize{8pt}{10pt}\selectfont
\centering
\caption[Pearson correlation scores for regression model comparison in train-test split]{\textbf{Pearson correlation scores for regression model comparison in train-test split.}
This table evaluates the performance of regression models in predicting cognitive scores (Memory, Executive Function, Language, Visuospatial) using 1-Nearest Neighbor as the continuous value imputer. The reported values are Pearson correlation coefficients between true and predicted cognitive scores. The testing set included 12 samples. Mean and Std columns represent the average and standard deviation of row-wise correlation scores across cognitive domains, providing an overall measure of model performance. True target residuals were adjusted for confounders before comparison with predicted values. Results are sorted by descending mean correlation coefficient to identify the best-performing regression model.}
\begin{tabular}{|p{1.8cm}p{2.cm}p{4.cm}||p{1.28cm}p{1.08cm}p{1.08cm}p{1.08cm}||p{0.6cm}p{0.6cm}|}
        \hline
        \textbf{Modality} & \textbf{Ordinal Imputer} & \textbf{Model} & \textbf{Executive Function} & \textbf{Language} & \textbf{Memory} & \textbf{Visuo-spatial} & \textbf{Mean} & \textbf{Std} \\ \hline
        Multiple & KNNImputer & TabNetRegressor default & 0.737 & 0.637 & 0.766 & 0.723 & 0.716 & 0.056 \\ 
        Multiple & Missing Indicator & XGBoostRegressor & 0.792 & 0.622 & 0.634 & 0.598 & 0.662 & 0.088 \\ 
        Multiple & Missing Indicator & TabNetRegressor custom & 0.750 & 0.700 & 0.679 & 0.515 & 0.661 & 0.102 \\ 
        Multiple & Most Frequent & XGBoostRegressor & 0.783 & 0.622 & 0.597 & 0.598 & 0.650 & 0.090 \\ 
        Multiple & KNNImputer & XGBoostRegressor & 0.785 & 0.622 & 0.553 & 0.598 & 0.640 & 0.101 \\ 
        Multiple & Most Frequent & RandomForestRegressor & 0.721 & 0.716 & 0.642 & 0.419 & 0.625 & 0.142 \\ 
        Multiple & Missing Indicator & RandomForestRegressor & 0.717 & 0.737 & 0.638 & 0.407 & 0.625 & 0.151 \\ 
        Multiple & Missing Indicator & PLSRegression 6 components & 0.782 & 0.623 & 0.634 & 0.346 & 0.596 & 0.182 \\ 
        Multiple & Most Frequent & PLSRegression 6 components & 0.770 & 0.631 & 0.636 & 0.322 & 0.590 & 0.190 \\ 
        Multiple & KNNImputer & PLSRegression 6 components & 0.761 & 0.626 & 0.630 & 0.299 & 0.579 & 0.197 \\ 
        Multiple & KNNImputer & RandomForestRegressor & 0.712 & 0.653 & 0.539 & 0.357 & 0.565 & 0.156 \\ 
        Multiple & Missing Indicator & TabNetRegressor default & 0.680 & 0.550 & 0.546 & 0.442 & 0.555 & 0.097 \\ 
        Multiple & Most Frequent & TabNetRegressor default & 0.579 & 0.713 & 0.712 & 0.207 & 0.553 & 0.239 \\ 
        MRI & - & TabNetRegressor default & 0.615 & 0.629 & 0.561 & 0.381 & 0.546 & 0.114 \\ 
        Multiple &  Most Frequent & PLSRegression 4 components & 0.721 & 0.526 & 0.529 & 0.303 & 0.520 & 0.171 \\ 
        Multiple & KNNImputer & PLSRegression 4 components & 0.719 & 0.520 & 0.524 & 0.311 & 0.518 & 0.167 \\ 
        MRI & - & TabNetRegressor custom & 0.687 & 0.543 & 0.457 & 0.343 & 0.508 & 0.145 \\ 
        Multiple & Missing Indicator & PLSRegression 4 components & 0.686 & 0.515 & 0.516 & 0.274 & 0.498 & 0.169 \\ 
        Multiple & Most Frequent & MultiTaskLasso & 0.755 & 0.409 & 0.512 & 0.303 & 0.495 & 0.193 \\ 
        Multiple & KNNImputer & MultiTaskLasso & 0.755 & 0.409 & 0.512 & 0.303 & 0.495 & 0.193 \\ 
        Multiple & Missing Indicator & MultiTaskLasso & 0.755 & 0.409 & 0.512 & 0.303 & 0.495 & 0.193 \\ 
        Multiple & Most Frequent & MultiTaskElasticNet & 0.753 & 0.406 & 0.509 & 0.299 & 0.491 & 0.194 \\ 
        Multiple & Missing Indicator & MultiTaskElasticNet & 0.753 & 0.406 & 0.509 & 0.299 & 0.491 & 0.194 \\ 
        Multiple & KNNImputer & MultiTaskElasticNet & 0.753 & 0.406 & 0.509 & 0.299 & 0.491 & 0.194 \\ 
        Multiple & KNNImputer & PLSRegression 10 components & 0.755 & 0.454 & 0.572 & 0.115 & 0.474 & 0.269 \\ 
        Multiple & Most Frequent & PLSRegression 10 components & 0.754 & 0.455 & 0.577 & 0.110 & 0.474 & 0.272 \\ 
        MRI & - & RandomForestRegressor & 0.582 & 0.645 & 0.540 & 0.104 & 0.468 & 0.246 \\ 
        Multiple & Missing Indicator & PLSRegression 10 components & 0.715 & 0.335 & 0.399 & 0.388 & 0.459 & 0.173 \\ 
        Multiple & Most Frequent & LinearRegression & 0.831 & 0.228 & 0.400 & 0.211 & 0.418 & 0.289 \\ 
        Multiple & KNNImputer & LinearRegression & 0.831 & 0.222 & 0.383 & 0.219 & 0.414 & 0.289 \\ 
        Multiple & Missing Indicator & LinearRegression & 0.831 & 0.212 & 0.383 & 0.219 & 0.411 & 0.291 \\ 
        MRI & - & XGBoostRegressor & 0.589 & 0.414 & 0.335 & 0.275 & 0.403 & 0.136 \\ 
        Multiple & KNNImputer & PLSRegression 2 components & 0.429 & 0.647 & 0.621 & -0.103 & 0.399 & 0.348 \\ 
        Multiple & Most Frequent & PLSRegression 2 components & 0.422 & 0.646 & 0.618 & -0.109 & 0.394 & 0.350 \\ 
        MRI & - & PLSRegression 10 components & 0.657 & 0.420 & 0.304 & 0.197 & 0.394 & 0.197 \\ 
        Multiple & Missing Indicator & PLSRegression 2 components & 0.414 & 0.646 & 0.615 & -0.116 & 0.390 & 0.353 \\ 
        Multiple & KNNImputer & TabNetRegressor custom & 0.317 & 0.696 & 0.624 & -0.080 & 0.389 & 0.353 \\ 
        MRI & - & LinearRegression & 0.716 & 0.371 & 0.296 & 0.084 & 0.367 & 0.263 \\ 
        MRI & - & PLSRegression 2 components & 0.360 & 0.612 & 0.567 & -0.148 & 0.348 & 0.348 \\ 
        MRI & - & PLSRegression 4 components & 0.515 & 0.454 & 0.414 & 0.005 & 0.347 & 0.232 \\ 
        Multiple & Most Frequent & TabNetRegressor custom & 0.345 & 0.211 & 0.443 & 0.028 & 0.257 & 0.180 \\  \hline
    \end{tabular}
    \label{tab:prediction_test_train_corr}
\end{table*}

\begin{table*}[!ht]
\fontsize{8pt}{10pt}\selectfont
\centering
\caption[Mean absolute error scores for regression model comparison in train-test split]{\textbf{Mean absolute error scores for regression model comparison in train-test split.}
This table evaluates the performance of regression models in predicting cognitive scores (Memory, Executive Function, Language, Visuospatial) using 1-Nearest Neighbor as the continuous value imputer. The reported values are the mean absolute error (MAE) between true and predicted cognitive scores. The testing set included 12 samples. Mean and Std denote the average and standard deviation of row-wise MAE scores across cognitive domains, providing an overall measure of model performance. True target residuals were adjusted for confounders before comparison with predicted values. Results are sorted by ascending mean MAE to identify the best-performing regression model.}
\begin{tabular}{|p{1.8cm}p{2.cm}p{4.cm}||p{1.28cm}p{1.08cm}p{1.08cm}p{1.08cm}||p{0.6cm}p{0.6cm}|}
        \hline
        \textbf{Modality} & \textbf{Ordinal Imputer} & \textbf{Model} & \textbf{Executive Function} & \textbf{Language} & \textbf{Memory} & \textbf{Visuo-spatial} & \textbf{Mean} & \textbf{Std} \\ \hline
        Multiple & KNNImputer & TabNetRegressor default & 0.487 & 0.685 & 0.615 & 0.383 & 0.543 & 0.134 \\ 
        Multiple & Missing Indicator & TabNetRegressor custom & 0.636 & 0.583 & 0.689 & 0.499 & 0.602 & 0.081 \\ 
        Multiple & Missing Indicator & XGBoostRegressor & 0.564 & 0.632 & 0.673 & 0.540 & 0.602 & 0.061 \\ 
        Multiple & Most Frequent & XGBoostRegressor & 0.576 & 0.632 & 0.695 & 0.540 & 0.611 & 0.068 \\ 
        Multiple & KNNImputer & XGBoostRegressor & 0.574 & 0.632 & 0.698 & 0.540 & 0.611 & 0.069 \\ 
        MRI & - & TabNetRegressor default & 0.603 & 0.637 & 0.702 & 0.554 & 0.624 & 0.062 \\ 
        Multiple &Missing Indicator & PLSRegression 6 components & 0.586 & 0.654 & 0.665 & 0.632 & 0.634 & 0.035 \\ 
        Multiple &Most Frequent & RandomForestRegressor & 0.661 & 0.622 & 0.674 & 0.598 & 0.639 & 0.035 \\ 
        Multiple &Missing Indicator & RandomForestRegressor & 0.664 & 0.611 & 0.680 & 0.602 & 0.639 & 0.039 \\ 
        Multiple &Most Frequent & PLSRegression 6 components & 0.600 & 0.652 & 0.675 & 0.642 & 0.642 & 0.032 \\ 
        Multiple &KNNImputer & PLSRegression 6 components & 0.610 & 0.651 & 0.678 & 0.653 & 0.648 & 0.028 \\ 
        Multiple &Most Frequent & TabNetRegressor default & 0.782 & 0.536 & 0.629 & 0.660 & 0.652 & 0.102 \\ 
        Multiple &Missing Indicator & TabNetRegressor default & 0.668 & 0.673 & 0.741 & 0.545 & 0.657 & 0.082 \\ 
        Multiple &KNNImputer & RandomForestRegressor & 0.661 & 0.648 & 0.732 & 0.617 & 0.664 & 0.049 \\ 
        Multiple &Most Frequent & PLSRegression 10 components & 0.602 & 0.694 & 0.703 & 0.695 & 0.674 & 0.048 \\ 
        Multiple &KNNImputer & PLSRegression 10 components & 0.610 & 0.692 & 0.700 & 0.696 & 0.674 & 0.043 \\ 
        Multiple &Most Frequent & PLSRegression 4 components & 0.650 & 0.695 & 0.724 & 0.646 & 0.679 & 0.037 \\ 
        Multiple &KNNImputer & PLSRegression 4 components & 0.652 & 0.699 & 0.724 & 0.645 & 0.680 & 0.038 \\ 
        Multiple &Missing Indicator & PLSRegression 4 components & 0.672 & 0.689 & 0.713 & 0.655 & 0.682 & 0.025 \\ 
        MRI & - & TabNetRegressor custom & 0.611 & 0.698 & 0.793 & 0.631 & 0.683 & 0.082 \\ 
        Multiple & Most Frequent & LinearRegression & 0.527 & 0.799 & 0.767 & 0.650 & 0.686 & 0.124 \\ 
        Multiple &KNNImputer & LinearRegression & 0.527 & 0.802 & 0.772 & 0.649 & 0.687 & 0.126 \\ 
        Multiple &Missing Indicator & LinearRegression & 0.528 & 0.808 & 0.776 & 0.648 & 0.690 & 0.128 \\ 
        MRI & - & RandomForestRegressor & 0.714 & 0.637 & 0.751 & 0.678 & 0.695 & 0.049 \\ 
        Multiple &Missing Indicator & PLSRegression 10 components & 0.638 & 0.774 & 0.808 & 0.623 & 0.711 & 0.094 \\ 
        MRI & - & LinearRegression & 0.648 & 0.725 & 0.793 & 0.687 & 0.713 & 0.062 \\ 
        MRI & - & PLSRegression 10 components & 0.724 & 0.700 & 0.813 & 0.676 & 0.728 & 0.060 \\ 
        MRI & - & XGBoostRegressor & 0.715 & 0.769 & 0.812 & 0.642 & 0.735 & 0.074 \\ 
        Multiple &KNNImputer & PLSRegression 2 components & 0.815 & 0.695 & 0.728 & 0.709 & 0.737 & 0.054 \\ 
        Multiple &Missing Indicator & PLSRegression 2 components & 0.819 & 0.690 & 0.729 & 0.710 & 0.737 & 0.057 \\ 
        Multiple &Most Frequent & PLSRegression 2 components & 0.819 & 0.697 & 0.733 & 0.710 & 0.740 & 0.055 \\ 
        Multiple &Missing Indicator & MultiTaskLasso & 0.661 & 0.856 & 0.854 & 0.616 & 0.747 & 0.126 \\ 
        Multiple &KNNImputer & MultiTaskLasso & 0.661 & 0.856 & 0.854 & 0.616 & 0.747 & 0.126 \\ 
        Multiple &Most Frequent & MultiTaskLasso & 0.661 & 0.856 & 0.854 & 0.616 & 0.747 & 0.126 \\ 
        Multiple &KNNImputer & MultiTaskElasticNet & 0.661 & 0.856 & 0.854 & 0.617 & 0.747 & 0.126 \\ 
        Multiple &Most Frequent & MultiTaskElasticNet & 0.661 & 0.856 & 0.854 & 0.617 & 0.747 & 0.126 \\ 
        Multiple &Missing Indicator & MultiTaskElasticNet & 0.661 & 0.856 & 0.854 & 0.617 & 0.747 & 0.126 \\ 
        MRI & - & PLSRegression 2 components & 0.848 & 0.716 & 0.780 & 0.718 & 0.765 & 0.063 \\ 
        MRI & - & PLSRegression 4 components & 0.793 & 0.753 & 0.809 & 0.711 & 0.766 & 0.044 \\ 
        Multiple &Most Frequent & TabNetRegressor custom & 0.825 & 0.835 & 0.828 & 0.642 & 0.782 & 0.094 \\ 
        Multiple &KNNImputer & TabNetRegressor custom & 0.890 & 0.738 & 0.769 & 0.799 & 0.799 & 0.066 \\ 
        MRI & - & MultiTaskElasticNet & 0.837 & 0.926 & 0.974 & 0.625 & 0.840 & 0.154 \\ 
        MRI & - & MultiTaskLasso & 0.837 & 0.926 & 0.974 & 0.625 & 0.840 & 0.154 \\ \hline
\end{tabular}
\label{tab:prediction_test_train_mae}
\end{table*}

\begin{table*}[!ht]
\fontsize{8pt}{10pt}\selectfont
\centering
\caption[Pearson correlation scores for regression model comparison in leave-one-complete-out cross-validation]{\textbf{Pearson correlation scores for regression model comparison in leave-one-complete-out cross-validation.}
This table presents the performance of regression models in predicting cognitive scores (Memory, Executive Function, Language, Visuospatial) using 1-Nearest Neighbor for continuous value imputation. Values represent the Pearson correlation coefficients between the observed and predicted cognitive scores. Leave-one-complete-out cross-validation was performed on 12 unique test samples, each containing complete cognitive domain data. The mean and standard deviation (std) of the correlation coefficients were calculated row-wise across cognitive domains to provide an overall assessment of model performance. True target residuals were adjusted to account for the effects of confounders before comparison with predicted values. Results are sorted in descending order of the mean correlation coefficient, highlighting the top-performing regression model.}
\begin{tabular}{|p{1.8cm}p{2cm}p{4.cm}||p{1.28cm}p{1.08cm}p{1.08cm}p{1.08cm}||p{0.6cm}p{0.6cm}|}
        \hline
        \textbf{Modality} & \textbf{Ordinal Imputer} & \textbf{Model} & \textbf{Executive Function} & \textbf{Language} & \textbf{Memory} & \textbf{Visuo-spatial} & \textbf{Mean} & \textbf{Std} \\ \hline
        Multiple &Missing Indicator & TabNetRegressor default & 0.797 & 0.802 & 0.901 & 0.476 & 0.744 & 0.185 \\ 
        Multiple &KNNImputer & TabNetRegressor default & 0.709 & 0.739 & 0.763 & 0.585 & 0.699 & 0.079 \\ 
        Multiple &Most Frequent & TabNetRegressor custom & 0.872 & 0.775 & 0.672 & 0.465 & 0.696 & 0.174 \\ 
        Multiple &Most Frequent & RandomForestRegressor & 0.746 & 0.760 & 0.723 & 0.445 & 0.669 & 0.150 \\ 
        Multiple &KNNImputer & RandomForestRegressor & 0.699 & 0.725 & 0.736 & 0.383 & 0.636 & 0.169 \\ 
        Multiple &Missing Indicator & RandomForestRegressor & 0.705 & 0.657 & 0.679 & 0.421 & 0.616 & 0.131 \\ 
        Multiple &Missing Indicator & PLSRegression 6 components & 0.788 & 0.615 & 0.629 & 0.360 & 0.598 & 0.177 \\ 
        Multiple &Most Frequent & PLSRegression 6 components & 0.775 & 0.632 & 0.638 & 0.329 & 0.593 & 0.188 \\ 
        Multiple &KNNImputer & TabNetRegressor custom & 0.714 & 0.786 & 0.717 & 0.153 & 0.592 & 0.295 \\ 
        Multiple &KNNImputer & PLSRegression 6 components & 0.764 & 0.626 & 0.632 & 0.303 & 0.581 & 0.196 \\ 
        Multiple &Most Frequent & TabNetRegressor default & 0.543 & 0.688 & 0.655 & 0.360 & 0.562 & 0.148 \\ 
        MRI & - & RandomForestRegressor & 0.679 & 0.730 & 0.584 & 0.252 & 0.561 & 0.215 \\ 
        Multiple &Missing Indicator & XGBoostRegressor & 0.827 & 0.440 & 0.582 & 0.363 & 0.553 & 0.204 \\ 
        Multiple &KNNImputer & XGBoostRegressor & 0.779 & 0.478 & 0.471 & 0.366 & 0.523 & 0.178 \\ 
        Multiple &Most Frequent & PLSRegression 4 components & 0.724 & 0.523 & 0.528 & 0.309 & 0.521 & 0.169 \\ 
        Multiple &Most Frequent & XGBoostRegressor & 0.776 & 0.431 & 0.523 & 0.350 & 0.520 & 0.185 \\ 
        Multiple &KNNImputer & PLSRegression 4 components & 0.722 & 0.518 & 0.523 & 0.317 & 0.520 & 0.165 \\ 
        MRI & - & XGBoostRegressor & 0.722 & 0.525 & 0.431 & 0.384 & 0.515 & 0.150 \\ 
        Multiple &Missing Indicator & PLSRegression 4 components & 0.688 & 0.513 & 0.515 & 0.280 & 0.499 & 0.168 \\ 
        Multiple &Most Frequent & MultiTaskLasso & 0.750 & 0.407 & 0.503 & 0.314 & 0.494 & 0.188 \\ 
        Multiple &Missing Indicator & MultiTaskLasso & 0.750 & 0.407 & 0.503 & 0.314 & 0.494 & 0.188 \\ 
        Multiple &KNNImputer & MultiTaskLasso & 0.750 & 0.407 & 0.503 & 0.314 & 0.494 & 0.188 \\ 
        Multiple &Most Frequent & MultiTaskElasticNet & 0.749 & 0.404 & 0.500 & 0.309 & 0.490 & 0.189 \\ 
        Multiple &KNNImputer & MultiTaskElasticNet & 0.749 & 0.404 & 0.500 & 0.309 & 0.490 & 0.189 \\ 
        Multiple &Missing Indicator & MultiTaskElasticNet & 0.749 & 0.404 & 0.500 & 0.309 & 0.490 & 0.189 \\ 
        Multiple &Most Frequent & PLSRegression 10 components & 0.761 & 0.454 & 0.577 & 0.114 & 0.477 & 0.273 \\ 
        Multiple &KNNImputer & PLSRegression 10 components & 0.761 & 0.453 & 0.577 & 0.112 & 0.476 & 0.273 \\ 
        Multiple &Missing Indicator & PLSRegression 10 components & 0.719 & 0.333 & 0.399 & 0.393 & 0.461 & 0.175 \\ 
        Multiple &Missing Indicator & TabNetRegressor custom & 0.537 & 0.585 & 0.707 & -0.005 & 0.456 & 0.315 \\ 
        MRI & - & TabNetRegressor default & 0.599 & 0.501 & 0.402 & 0.198 & 0.425 & 0.171 \\ 
        Multiple &Most Frequent & LinearRegression & 0.831 & 0.221 & 0.388 & 0.219 & 0.414 & 0.289 \\ 
        Multiple &KNNImputer & LinearRegression & 0.831 & 0.213 & 0.370 & 0.227 & 0.410 & 0.289 \\ 
        Multiple &Missing Indicator & LinearRegression & 0.830 & 0.202 & 0.369 & 0.226 & 0.407 & 0.292 \\ 
        MRI & - & PLSRegression 10 components & 0.660 & 0.427 & 0.308 & 0.199 & 0.398 & 0.198 \\ 
        Multiple &KNNImputer & PLSRegression 2 components & 0.429 & 0.645 & 0.619 & -0.103 & 0.398 & 0.347 \\ 
        Multiple &Most Frequent & PLSRegression 2 components & 0.422 & 0.644 & 0.616 & -0.109 & 0.393 & 0.349 \\ 
        Multiple &Missing Indicator & PLSRegression 2 components & 0.414 & 0.644 & 0.613 & -0.117 & 0.389 & 0.352 \\ 
        MRI & - & LinearRegression & 0.719 & 0.376 & 0.302 & 0.087 & 0.371 & 0.262 \\ 
        MRI & - & PLSRegression 4 components & 0.516 & 0.455 & 0.415 & 0.006 & 0.348 & 0.232 \\ 
        MRI & - & PLSRegression 2 components & 0.359 & 0.611 & 0.566 & -0.149 & 0.347 & 0.348 \\ 
        MRI & - & TabNetRegressor custom & 0.459 & 0.364 & 0.439 & -0.106 & 0.289 & 0.266 \\ 
        MRI & - & MultiTaskLasso & -0.229 & -0.329 & 0.293 & 0.384 & 0.030 & 0.361 \\ 
        MRI & - & MultiTaskElasticNet & -0.229 & -0.329 & 0.293 & 0.384 & 0.030 & 0.361 \\  \hline
    \end{tabular}
    \label{tab:prediction_loonona_corr}
\end{table*}

\begin{table*}[!ht]
\fontsize{8pt}{10pt}\selectfont
\centering
\caption[Mean absolute error scores for regression model comparison in leave-one-complete-out cross-validation]{\textbf{Mean absolute error scores for regression model comparison in leave-one-complete-out cross-validation.}
This table presents the performance of regression models in predicting cognitive scores (Memory, Executive Function, Language, Visuospatial) using 1-Nearest Neighbor for continuous value imputation. Values represent the mean absolute error (MAE) between the observed and predicted cognitive scores. Leave-one-complete-out cross-validation was performed on 12 unique test samples, each containing complete cognitive domain data. The mean and standard deviation (std) of the MAE values were calculated row-wise across cognitive domains to provide an overall assessment of model performance. True target residuals were adjusted to account for the effects of confounders before comparison with predicted values. Results are sorted in ascending order of the average MAE, highlighting the top-performing regression model.}
\begin{tabular}{|p{1.8cm}p{2cm}p{4.cm}||p{1.28cm}p{1.08cm}p{1.08cm}p{1.08cm}||p{0.6cm}p{0.6cm}|}
        \hline
        \textbf{Modality} & \textbf{Ordinal Imputer} & \textbf{Model} & \textbf{Executive Function} & \textbf{Language} & \textbf{Memory} & \textbf{Visuo-spatial} & \textbf{Mean} & \textbf{Std} \\ \hline
        Multiple &Missing Indicator & TabNetRegressor default & 0.490 & 0.411 & 0.377 & 0.504 & 0.445 & 0.062 \\ 
        Multiple &Most Frequent & TabNetRegressor custom & 0.445 & 0.441 & 0.622 & 0.519 & 0.507 & 0.085 \\ 
        Multiple &KNNImputer & TabNetRegressor default & 0.604 & 0.613 & 0.619 & 0.528 & 0.591 & 0.043 \\ 
        Multiple &Most Frequent & RandomForestRegressor & 0.615 & 0.571 & 0.620 & 0.573 & 0.595 & 0.026 \\ 
        Multiple &KNNImputer & TabNetRegressor custom & 0.540 & 0.605 & 0.639 & 0.687 & 0.618 & 0.062 \\ 
        Multiple &Most Frequent & TabNetRegressor default & 0.732 & 0.521 & 0.663 & 0.574 & 0.622 & 0.093 \\ 
        Multiple &Missing Indicator & RandomForestRegressor & 0.626 & 0.638 & 0.644 & 0.589 & 0.624 & 0.025 \\ 
        Multiple &KNNImputer & RandomForestRegressor & 0.671 & 0.603 & 0.624 & 0.611 & 0.627 & 0.030 \\ 
        Multiple &Missing Indicator & PLSRegression 6 components & 0.576 & 0.662 & 0.675 & 0.625 & 0.635 & 0.044 \\ 
        Multiple &Most Frequent & PLSRegression 6 components & 0.593 & 0.653 & 0.674 & 0.639 & 0.640 & 0.035 \\ 
        Multiple &KNNImputer & PLSRegression 6 components & 0.605 & 0.651 & 0.677 & 0.651 & 0.646 & 0.030 \\ 
        MRI & - & RandomForestRegressor & 0.687 & 0.581 & 0.701 & 0.637 & 0.651 & 0.054 \\ 
        Multiple &Missing Indicator & XGBoostRegressor & 0.496 & 0.785 & 0.735 & 0.608 & 0.656 & 0.130 \\ 
        Multiple &KNNImputer & XGBoostRegressor & 0.554 & 0.696 & 0.802 & 0.610 & 0.665 & 0.108 \\ 
        Multiple &Most Frequent & XGBoostRegressor & 0.555 & 0.733 & 0.773 & 0.614 & 0.669 & 0.102 \\ 
        Multiple &Most Frequent & PLSRegression 10 components & 0.592 & 0.696 & 0.709 & 0.694 & 0.673 & 0.054 \\ 
        Multiple &KNNImputer & PLSRegression 10 components & 0.601 & 0.695 & 0.704 & 0.698 & 0.674 & 0.049 \\ 
        Multiple &Most Frequent & PLSRegression 4 components & 0.644 & 0.695 & 0.726 & 0.645 & 0.677 & 0.040 \\ 
        Multiple &KNNImputer & PLSRegression 4 components & 0.646 & 0.698 & 0.727 & 0.644 & 0.679 & 0.041 \\ 
        Multiple &Missing Indicator & PLSRegression 4 components & 0.666 & 0.689 & 0.715 & 0.654 & 0.681 & 0.027 \\ 
        Multiple &Most Frequent & LinearRegression & 0.520 & 0.800 & 0.774 & 0.647 & 0.685 & 0.129 \\ 
        Multiple &KNNImputer & LinearRegression & 0.520 & 0.803 & 0.779 & 0.645 & 0.687 & 0.131 \\ 
        Multiple &Missing Indicator & LinearRegression & 0.521 & 0.810 & 0.784 & 0.645 & 0.690 & 0.134 \\ 
        Multiple &Missing Indicator & TabNetRegressor custom & 0.733 & 0.721 & 0.672 & 0.667 & 0.698 & 0.034 \\ 
        MRI & - & LinearRegression & 0.645 & 0.713 & 0.793 & 0.685 & 0.709 & 0.063 \\ 
        Multiple &Missing Indicator & PLSRegression 10 components & 0.632 & 0.777 & 0.812 & 0.621 & 0.711 & 0.098 \\ 
        MRI & - & TabNetRegressor custom & 0.757 & 0.706 & 0.787 & 0.601 & 0.712 & 0.081 \\ 
        MRI & - & XGBoostRegressor & 0.673 & 0.700 & 0.843 & 0.639 & 0.714 & 0.090 \\ 
        MRI & - & TabNetRegressor default & 0.728 & 0.686 & 0.827 & 0.650 & 0.723 & 0.077 \\ 
        MRI & - & PLSRegression 10 components & 0.723 & 0.694 & 0.813 & 0.675 & 0.726 & 0.061 \\ 
        MRI & KNNImputer & PLSRegression 2 components & 0.815 & 0.693 & 0.726 & 0.709 & 0.736 & 0.054 \\ 
        Multiple &Missing Indicator & PLSRegression 2 components & 0.818 & 0.688 & 0.726 & 0.711 & 0.736 & 0.057 \\ 
        Multiple &Most Frequent & PLSRegression 2 components & 0.818 & 0.695 & 0.731 & 0.711 & 0.739 & 0.055 \\ 
        Multiple &Missing Indicator & MultiTaskLasso & 0.660 & 0.854 & 0.856 & 0.615 & 0.746 & 0.127 \\ 
        Multiple &KNNImputer & MultiTaskLasso & 0.660 & 0.854 & 0.856 & 0.615 & 0.746 & 0.127 \\ 
        Multiple &Most Frequent & MultiTaskLasso & 0.660 & 0.854 & 0.856 & 0.615 & 0.746 & 0.127 \\ 
        Multiple &KNNImputer & MultiTaskElasticNet & 0.660 & 0.854 & 0.856 & 0.615 & 0.747 & 0.127 \\ 
        Multiple &Missing Indicator & MultiTaskElasticNet & 0.660 & 0.854 & 0.856 & 0.615 & 0.747 & 0.127 \\ 
       Multiple & Most Frequent & MultiTaskElasticNet & 0.660 & 0.854 & 0.856 & 0.615 & 0.747 & 0.127 \\ 
        MRI & - & PLSRegression 4 components & 0.790 & 0.749 & 0.806 & 0.711 & 0.764 & 0.043 \\ 
        MRI & - & PLSRegression 2 components & 0.848 & 0.714 & 0.778 & 0.718 & 0.765 & 0.063 \\ 
        MRI & - & MultiTaskElasticNet & 0.838 & 0.924 & 0.973 & 0.625 & 0.840 & 0.154 \\ 
        MRI & - & MultiTaskLasso & 0.838 & 0.924 & 0.973 & 0.625 & 0.840 & 0.154 \\  \hline
    \end{tabular}
    \label{tab:prediction_loonona_mae}
\end{table*}

\begin{table*}[!ht]
\fontsize{8pt}{10pt}\selectfont
\centering
\caption[Pearson correlation scores for regression model comparison in leave-one-missing-out cross-validation]{\textbf{Pearson correlation scores for regression model comparison in leave-one-missing-out cross-validation.}
This table presents the performance of regression models in predicting cognitive scores (Memory, Executive Function, Language, Visuospatial) using 1-Nearest Neighbor for continuous value imputation. Values represent the Pearson correlation coefficients between the observed and predicted cognitive scores. Leave-one-missing-out cross-validation was conducted on 20 randomly selected incomplete test samples. The mean and standard deviation (std) of the Pearson correlation coefficients were calculated row-wise across cognitive domains to provide an overall assessment of model performance. True target residuals were adjusted to account for the effects of confounders before comparison with predicted values. Results are sorted in descending order of the mean correlation coefficient to highlight the best-performing regression model.}
    \begin{tabular}{|p{1.8cm}p{2.cm}p{4.cm}||p{1.28cm}p{1.08cm}p{1.08cm}p{1.08cm}||p{0.6cm}p{0.6cm}|}
        \hline
        \textbf{Modality} & \textbf{Ordinal Imputer} & \textbf{Model} & \textbf{Executive Function} & \textbf{Language} & \textbf{Memory} & \textbf{Visuo-spatial} & \textbf{Mean} & \textbf{Std} \\ \hline
        Multiple &Missing Indicator & TabNetRegressor default & 0.402 & 0.692 & 0.679 & 0.194 & 0.492 & 0.240 \\ 
        Multiple &Most Frequent & TabNetRegressor custom & 0.529 & 0.828 & 0.493 & -0.013 & 0.459 & 0.349 \\ 
        Multiple &KNNImputer & RandomForestRegressor & 0.378 & 0.701 & 0.717 & 0.033 & 0.457 & 0.323 \\ 
        Multiple &Most Frequent & RandomForestRegressor & 0.392 & 0.727 & 0.754 & -0.054 & 0.455 & 0.377 \\ 
        Multiple &Missing Indicator & XGBoostRegressor & 0.591 & 0.707 & 0.708 & -0.197 & 0.452 & 0.436 \\ 
        Multiple &KNNImputer & TabNetRegressor default & 0.487 & 0.668 & 0.620 & -0.008 & 0.441 & 0.310 \\ 
        MRI & - & RandomForestRegressor & 0.433 & 0.760 & 0.682 & -0.115 & 0.440 & 0.395 \\ 
        MRI & - & XGBoostRegressor & 0.478 & 0.669 & 0.703 & -0.093 & 0.439 & 0.368 \\ 
        Multiple &Missing Indicator & RandomForestRegressor & 0.408 & 0.719 & 0.695 & -0.086 & 0.434 & 0.374 \\ 
        Multiple &Most Frequent & PLSRegression 2 components & 0.380 & 0.671 & 0.628 & -0.022 & 0.414 & 0.318 \\ 
        Multiple &Most Frequent & PLSRegression 4 components & 0.363 & 0.701 & 0.634 & -0.042 & 0.414 & 0.337 \\ 
        Multiple &KNNImputer & PLSRegression 2 components & 0.382 & 0.667 & 0.622 & -0.026 & 0.411 & 0.317 \\ 
        Multiple &Missing Indicator & PLSRegression 2 components & 0.379 & 0.671 & 0.622 & -0.030 & 0.411 & 0.321 \\ 
        MRI & - & PLSRegression 2 components & 0.380 & 0.669 & 0.643 & -0.059 & 0.408 & 0.338 \\ 
        Multiple &Missing Indicator & PLSRegression 4 components & 0.361 & 0.702 & 0.614 & -0.048 & 0.407 & 0.336 \\ 
        Multiple &Missing Indicator & TabNetRegressor custom & 0.507 & 0.529 & 0.568 & 0.024 & 0.407 & 0.257 \\ 
        Multiple &KNNImputer & PLSRegression 4 components & 0.358 & 0.690 & 0.622 & -0.050 & 0.405 & 0.335 \\ 
        MRI & - & PLSRegression 4 components & 0.342 & 0.686 & 0.609 & -0.041 & 0.399 & 0.328 \\ 
        Multiple &KNNImputer & XGBoostRegressor & 0.561 & 0.547 & 0.586 & -0.191 & 0.376 & 0.378 \\ 
        Multiple &Most Frequent & XGBoostRegressor & 0.508 & 0.545 & 0.599 & -0.191 & 0.365 & 0.373 \\ 
        Multiple &KNNImputer & TabNetRegressor custom & 0.299 & 0.689 & 0.494 & -0.059 & 0.356 & 0.319 \\ 
        Multiple &Most Frequent & PLSRegression 6 components & 0.287 & 0.659 & 0.624 & -0.162 & 0.352 & 0.382 \\ 
        Multiple & Missing Indicator & PLSRegression 6 components & 0.315 & 0.680 & 0.597 & -0.208 & 0.346 & 0.401 \\ 
        Multiple &KNNImputer & PLSRegression 6 components & 0.277 & 0.638 & 0.616 & -0.174 & 0.339 & 0.380 \\ 
        MRI & - & TabNetRegressor custom & 0.124 & 0.597 & 0.487 & -0.006 & 0.301 & 0.287 \\ 
        Multiple & Most Frequent & PLSRegression 10 components & 0.241 & 0.590 & 0.672 & -0.312 & 0.298 & 0.448 \\ 
        Multiple &KNNImputer & PLSRegression 10 components & 0.246 & 0.589 & 0.670 & -0.343 & 0.291 & 0.461 \\ 
        Multiple &Most Frequent & TabNetRegressor default & 0.115 & 0.479 & 0.490 & -0.004 & 0.270 & 0.253 \\ 
        Multiple & Missing Indicator & PLSRegression 10 components & 0.216 & 0.562 & 0.616 & -0.318 & 0.269 & 0.429 \\ 
        Multiple &KNNImputer & LinearRegression & 0.243 & 0.440 & 0.646 & -0.299 & 0.258 & 0.406 \\ 
        Multiple &Most Frequent & LinearRegression & 0.242 & 0.444 & 0.657 & -0.316 & 0.257 & 0.418 \\ 
        Multiple &Missing Indicator & LinearRegression & 0.241 & 0.433 & 0.630 & -0.308 & 0.249 & 0.404 \\ 
        MRI & - & PLSRegression 10 components & 0.106 & 0.572 & 0.587 & -0.270 & 0.249 & 0.412 \\ 
        MRI & - & TabNetRegressor default & 0.135 & 0.469 & 0.403 & -0.127 & 0.220 & 0.273 \\ 
        MRI & - & LinearRegression & 0.138 & 0.346 & 0.604 & -0.304 & 0.196 & 0.384 \\ 
        Multiple & Missing Indicator & MultiTaskElasticNet & -0.017 & 0.307 & 0.143 & 0.134 & 0.142 & 0.132 \\ 
        Multiple &Most Frequent & MultiTaskElasticNet & -0.017 & 0.307 & 0.143 & 0.134 & 0.142 & 0.132 \\ 
        Multiple &KNNImputer & MultiTaskElasticNet & -0.017 & 0.307 & 0.143 & 0.134 & 0.142 & 0.132 \\ 
        Multiple &Missing Indicator & MultiTaskLasso & -0.017 & 0.307 & 0.143 & 0.133 & 0.142 & 0.132 \\ 
        Multiple &Most Frequent & MultiTaskLasso & -0.017 & 0.307 & 0.143 & 0.133 & 0.142 & 0.132 \\ 
        Multiple &KNNImputer & MultiTaskLasso & -0.017 & 0.307 & 0.143 & 0.133 & 0.142 & 0.132 \\ 
        MRI & - & MultiTaskLasso & -0.486 & -0.190 & 0.462 & 0.488 & 0.068 & 0.485 \\ 
        MRI & - & MultiTaskElasticNet & -0.486 & -0.190 & 0.462 & 0.488 & 0.068 & 0.485 \\  \hline
    \end{tabular}
    \label{tab:prediction_loona_corr}
\end{table*}

\begin{table*}[!ht]
\fontsize{8pt}{10pt}\selectfont
\centering
\caption[Mean absolute error scores of regression models comparison in leave-one-missing-out cross-validation]{\textbf{Mean absolute error scores of regression models comparison in leave-one-missing-out cross-validation.}
Machine learning model performance in predicting multivariate cognitive scores (Memory, Executive Function, Language, Visuospatial) using 1-Nearest Neighbor as the continuous value imputer. Values represent the MAE between the true and predicted cognitive scores. Leave-one-out cross-validation was carried out on 20 randomly selected incomplete samples test samples. Mean and standard deviation (std) were calculated row-wise across cognitive domains to provide an overall assessment of regression model performance. True target residuals were adjusted to remove the effects of confounders before comparison with the predicted target values. Results are sorted in descending order of the average MAE to highlight the best-performing regression model.}
    \begin{tabular}{|p{1.8cm}p{2cm}p{4.cm}||p{1.28cm}p{1.08cm}p{1.08cm}p{1.08cm}||p{0.6cm}p{0.6cm}|}
        \hline
        \textbf{Modality} & \textbf{Ordinal Imputer} & \textbf{Model} & \textbf{Executive Function} & \textbf{Language} & \textbf{Memory} & \textbf{Visuo-spatial} & \textbf{Mean} & \textbf{Std} \\ \hline
        MRI & - & XGBoostRegressor & 0.530 & 0.459 & 0.473 & 0.572 & 0.509 & 0.052 \\ 
        Multiple & Most Frequent & RandomForestRegressor & 0.629 & 0.450 & 0.425 & 0.551 & 0.514 & 0.094 \\ 
        MRI & - & RandomForestRegressor & 0.594 & 0.423 & 0.490 & 0.549 & 0.514 & 0.074 \\ 
        Multiple &KNNImputer & RandomForestRegressor & 0.644 & 0.458 & 0.452 & 0.524 & 0.520 & 0.089 \\ 
        Multiple &Missing Indicator & RandomForestRegressor & 0.596 & 0.444 & 0.480 & 0.565 & 0.521 & 0.071 \\ 
        Multiple &Missing Indicator & XGBoostRegressor & 0.578 & 0.437 & 0.507 & 0.566 & 0.522 & 0.065 \\ 
        Multiple &Missing Indicator & PLSRegression 2 components & 0.606 & 0.488 & 0.499 & 0.547 & 0.535 & 0.054 \\ 
        MRI & - & PLSRegression 2 components & 0.621 & 0.480 & 0.497 & 0.545 & 0.536 & 0.063 \\ 
        Multiple &Most Frequent & PLSRegression 2 components & 0.609 & 0.491 & 0.497 & 0.547 & 0.536 & 0.055 \\ 
        Multiple &Missing Indicator & TabNetRegressor default & 0.622 & 0.444 & 0.521 & 0.559 & 0.537 & 0.074 \\ 
        Multiple &KNNImputer & PLSRegression 2 components & 0.607 & 0.494 & 0.503 & 0.548 & 0.538 & 0.052 \\ 
        Multiple &Missing Indicator & PLSRegression 4 components & 0.613 & 0.466 & 0.499 & 0.579 & 0.539 & 0.068 \\ 
        Multiple &Most Frequent & PLSRegression 4 components & 0.619 & 0.467 & 0.485 & 0.587 & 0.540 & 0.075 \\ 
        Multiple &Most Frequent & TabNetRegressor custom & 0.509 & 0.386 & 0.623 & 0.641 & 0.540 & 0.118 \\ 
        Multiple & KNNImputer & PLSRegression 4 components & 0.621 & 0.477 & 0.500 & 0.587 & 0.546 & 0.069 \\ 
        MRI & - & PLSRegression 4 components & 0.653 & 0.465 & 0.511 & 0.569 & 0.550 & 0.081 \\ 
        MRI & KNNImputer & XGBoostRegressor & 0.560 & 0.546 & 0.548 & 0.565 & 0.555 & 0.009 \\ 
        Multiple & Missing Indicator & PLSRegression 6 components & 0.649 & 0.461 & 0.498 & 0.619 & 0.557 & 0.091 \\ 
        Multiple &Most Frequent & XGBoostRegressor & 0.585 & 0.556 & 0.522 & 0.565 & 0.557 & 0.026 \\ 
        Multiple &Most Frequent & PLSRegression 6 components & 0.688 & 0.483 & 0.484 & 0.615 & 0.567 & 0.101 \\ 
        Multiple & KNNImputer & PLSRegression 6 components & 0.693 & 0.496 & 0.491 & 0.614 & 0.574 & 0.098 \\ 
        Multiple & KNNImputer & TabNetRegressor default & 0.604 & 0.544 & 0.572 & 0.586 & 0.576 & 0.025 \\ 
        Multiple & Missing Indicator & TabNetRegressor custom & 0.611 & 0.564 & 0.548 & 0.591 & 0.579 & 0.028 \\ 
        MRI & - & PLSRegression 10 components & 0.705 & 0.522 & 0.501 & 0.589 & 0.579 & 0.092 \\ 
        Multiple &KNNImputer & PLSRegression 10 components & 0.680 & 0.529 & 0.484 & 0.636 & 0.582 & 0.091 \\ 
        Multiple &Most Frequent & PLSRegression 10 components & 0.686 & 0.527 & 0.490 & 0.629 & 0.583 & 0.090 \\ 
        Multiple & KNNImputer & MultiTaskLasso & 0.704 & 0.562 & 0.578 & 0.504 & 0.587 & 0.084 \\ 
        Multiple &Most Frequent & MultiTaskLasso & 0.704 & 0.562 & 0.578 & 0.504 & 0.587 & 0.084 \\ 
        Multiple & Missing Indicator & MultiTaskLasso & 0.704 & 0.562 & 0.578 & 0.504 & 0.587 & 0.084 \\ 
        Multiple & Missing Indicator & MultiTaskElasticNet & 0.704 & 0.562 & 0.578 & 0.505 & 0.587 & 0.084 \\ 
        Multiple & KNNImputer & MultiTaskElasticNet & 0.704 & 0.562 & 0.578 & 0.505 & 0.587 & 0.084 \\ 
        Multiple &Most Frequent & MultiTaskElasticNet & 0.704 & 0.562 & 0.578 & 0.505 & 0.587 & 0.084 \\ 
        MRI & - & TabNetRegressor custom & 0.701 & 0.511 & 0.582 & 0.579 & 0.593 & 0.079 \\ 
        Multiple & Missing Indicator & PLSRegression 10 components & 0.700 & 0.557 & 0.496 & 0.643 & 0.599 & 0.091 \\ 
        Multiple & Most Frequent & TabNetRegressor default & 0.709 & 0.572 & 0.551 & 0.599 & 0.608 & 0.070 \\ 
        MRI  & - & LinearRegression & 0.709 & 0.638 & 0.492 & 0.637 & 0.619 & 0.091 \\ 
        Multiple & Most Frequent & LinearRegression & 0.671 & 0.638 & 0.512 & 0.670 & 0.623 & 0.075 \\ 
        Multiple & KNNImputer & LinearRegression & 0.671 & 0.638 & 0.516 & 0.674 & 0.625 & 0.074 \\ 
        Multiple & Missing Indicator & LinearRegression & 0.671 & 0.641 & 0.527 & 0.671 & 0.628 & 0.068 \\ 
        Multiple & - & TabNetRegressor default & 0.671 & 0.592 & 0.708 & 0.595 & 0.642 & 0.057 \\ 
        Multiple & - & MultiTaskLasso & 0.810 & 0.662 & 0.631 & 0.489 & 0.648 & 0.131 \\ 
        Multiple & - & MultiTaskElasticNet & 0.810 & 0.662 & 0.631 & 0.489 & 0.648 & 0.131 \\ 
        \textbf{Multiple} & \textbf{KNNImputer} &  \textbf{TabNetRegressor custom } & \textbf{0.823} & \textbf{0.619} & \textbf{0.612} & \textbf{0.629} & \textbf{0.671} & \textbf{0.102} \\ \hline
    \end{tabular}
    \label{tab:prediction_loona_mae}
\end{table*}

